\pdfoutput=1

\documentclass[pdflatex,sn-nature]{sn-jnl}%

\usepackage{graphicx}%
\usepackage{float}%
\usepackage{multirow}%
\usepackage{amsmath,amssymb,amsfonts}%
\usepackage{amsthm}%
\usepackage{mathrsfs}%
\usepackage[title]{appendix}%
\usepackage{xcolor}%
\usepackage{textcomp}%
\usepackage{manyfoot}%
\usepackage{booktabs}
\usepackage{eurosym}
\usepackage{booktabs}%
\usepackage{longtable}%
\usepackage{algorithm}%
\usepackage{algorithmicx}%
\usepackage{algpseudocode}%
\usepackage{listings}%
\usepackage{xurl}

\theoremstyle{thmstyleone}%

\theoremstyle{thmstyletwo}%

\theoremstyle{thmstylethree}%

\begin{document}

\title[Intuition-based rhetoric after far-right entry]{German parties shifted towards intuition-based rhetoric after the far right's parliamentary breakthrough}

\author*[1]{\fnm{Peer} \sur{Saleth}}\email{peer.saleth@uni-konstanz.de}

\author[1]{\fnm{Segun T.} \sur{Aroyehun}}%

\author[3]{\fnm{Fabio} \sur{Carrella}}%

\author[4]{\fnm{Christoph M.} \sur{Abels}}%

\author[2, 4]{\fnm{Stephan} \sur{Lewandowsky}}%

\author[1]{\fnm{David} \sur{Garcia}}%

\affil*[1]{\orgdiv{Department of Politics and Public Administration}, \orgname{University of Konstanz}, \orgaddress{\street{Universitätsstraße 10}, \city{Konstanz}, \postcode{78457}, \state{Baden-Württemberg}, \country{Germany}}}

\affil[2]{\orgdiv{School of Psychological Science}, \orgname{Faculty of Life Sciences, University of Bristol}, \orgaddress{\street{12a Priory Road}, \postcode{BS8 1TU}, \city{Bristol}, \country{United Kingdom}}}

\affil[3]{\orgdiv{Artificial Intelligence Lab (Recod.ai), Institute of Computing}, \orgname{University of Campinas}, \orgaddress{\street{R. Saturnino de Brito, 573}, \postcode{13083-852}, \city{Campinas}, \state{São Paulo}, \country{Brazil}}}

\affil[4]{\orgdiv{Department of Psychology}, \orgname{University of Potsdam}, \orgaddress{\street{Karl-Liebknecht-Str.~24--25}, \postcode{14476}, \city{Potsdam}, \state{Brandenburg}, \country{Germany}}}

\abstract{The spread of misinformation is widely perceived as a threat to democratic deliberation, yet how political elites' rhetorical commitments to truth shift alongside the rise of populist actors remains poorly understood. Analysing 4.5 million tweets and 59,170 parliamentary speeches by German political elites between 2015 and 2025, we measure evidence-based and intuition-based rhetoric using a validated distributed dictionary representation. Across both arenas, intuition-based language has become more prominent, and right-leaning actors consistently exhibit the lowest Evidence Minus Intuition (EMI) scores. The parliamentary entry of the extreme-right Alternative for Germany (AfD) in 2017 coincides with sharp downward shifts in EMI across the broader chamber, while a more gradual decline is observed on Twitter. These findings document an association between far-right visibility and a changing approach to truth in elite discourse in a multiparty European democracy.}

\maketitle
\section{Introduction}
Democratic institutions are widely reported to be experiencing a period of erosion and declining public trust, with the spread of misinformation across social media, hyper-partisan outlets, and political polarization posing substantial challenges to democratic deliberation \cite{house2020freedom, repucci2021democracy, lazer-2018, van-bavel-2021}. A shared consensus about what constitutes truth (e.g., public trust in elections and scientific institutions) is a necessary condition for societal coordination and, therefore, democratic functioning. Billions of people rely on social media platforms for their news consumption, which has also exposed them to a new corpus of disinformation and misinformation \cite{lazer-2018, van-bavel-2021, simchon-2022}. Misinformation can take many forms, ranging from conspiracy theories and propaganda to false news and deep fakes \cite{mccright_combatting_2017}, and can undermine trust in scientific and academic institutions. Previous literature identifies many examples, such as the undermining of scientific findings relating to global warming and vaccinations \cite{loomba_measuring_2021, van_der_linden_inoculating_2017}, or stoking of political polarization and cynicism \cite{ribeiro_everything_2017, jones-jang_perceptions_2021}. Misinformation exposure has been linked to shifts in public attitudes and behaviour, including increased electoral support for populist parties in Italy and contributions to ethnic hate crimes in Germany \cite{cantarella2023does,muller-2020,lorenz-spreen-2022} leading to a breakdown in the shared consensus of truth. 

In addition to the factual accuracy of specific claims, there are two critical dimensions of political discourse that are directly tied to democratic deliberation \cite{lewandowsky-2024}.  The first is \emph{honesty}, a personal attribute whereby the speaker sincerely and authentically expresses what the they deem to be true \cite{cooper-2023}. Because sincerity does not entail accuracy, a demonstrably false statement may still be perceived as honest if its speaker is seen to voice a genuinely held belief \cite{lewandowsky-2024}. Lasser et al. \cite{lasser-2023} distinguish fact- from belief-speaking in tweets by members of the US Congress, finding that belief-speaking predicts lower quality of shared news sources among Republican but not Democratic members (extending earlier evidence on partisan asymmetry in elite source sharing \cite{lasser-2022}). Belief-speaking predicts reliance on low-quality information sources, and it propagates: exposure to belief-speaking increases belief-speaking in downstream replies \cite{carrella-2025}.

The second is \emph{truth}, an external attribute concerned with how the state of the world is characterized. The \emph{truth} may be pursued through evidence-based reasoning grounded in facts, data and other elements of external reality, or through intuition-based reasoning grounded in feelings, instincts and personal conviction \cite{garrett-2017, hertwig_willful_2021, lewandowsky-2024}. This latter \emph{epistemic orientation} \cite{aroyehun-2026} echoes the radical constructivist conception of truth that also characterised 1930s fascism, which sometimes rejected the role of evidence altogether; Nazi ideology posited an ``organic truth'' rooted in personal experience and accessible only through inner reflection \cite{varshizky-2012, voegelin-1989}. Such postures are consequential, as beliefs about how truth is to be reached shape susceptibility to misperception and misinformation \cite{garrett-2017, lewandowsky-2017, mccright_combatting_2017}.

One way to understand the role of truth in political discourse is by measuring the use of evidence verses intuition-based language in congressional speeches, by using an Evidence-Minus-Intuition (EMI) score which measures the relative prevalence of evidence-grounded rhetoric to intuition-grounded rhetoric in political text \cite{aroyehun-2025}. Aroyehun et al. \cite{aroyehun-2025} track
evidence versus intuition-based language in congressional speeches, documenting a decline in evidence-based language since the mid-1970s alongside rising polarisation, increased income inequality, and falling
legislative productivity. They then extend EMI to 15 million parliamentary
speech segments across seven countries and 80 years \cite{aroyehun-2026}, finding it positively associated with
deliberative democracy, transparency and net of judicial independence. When discourse relies exclusively on intuition, evidence can no longer adjudicate between competing positions, and the democratic functions of
accountability, informed policy evaluation, and cross-partisan deliberation lose their footing \cite{aroyehun-2025}. It is important to note that evidence enables adjudication while intuition contributes moral and experiential dimensions that are often indispensable in public life. Rather, it is over-reliance on
intuition at the expense of evidence that is of concern for modern democracies \cite{aroyehun-2026}. Most of the available evidence concerning the decline of an evidence based approach to truth in the political context originates from the United States. Whether comparable shifts occur in multiparty European democracies across arenas (Twitter and Parliament), and how they relate to the rise of new far-right actors, remains an open question. 

Far right success can reshape the behaviour of mainstream parties \cite{vanspanje-2010, abouchadi-krause-2020, meguid-2005, gessler-hunger-2022}. Radical right electoral success exerts ``contagion'' effects on other parties' immigration positions \cite{vanspanje-2010}, and an effect that has been causally identified by exploiting electoral-threshold discontinuities in parliamentary representation \cite{abouchadi-krause-2020}. During the 2015 refugee crisis, radical right parties drove mainstream parties' attention to immigration within very short time intervals \cite{gessler-hunger-2022}, and in Germany specifically, far-right actors and their issues became discursively mainstreamed in mass media debates, with a marked shift in 2015, linked to the AfD's growing institutional access \cite{voelker-saldivia-2024}. This literature has, however, focused on \emph{what} parties talk about (issue salience) and \emph{where} they stand (policy positions). Whether far-right success is also associated with shifts in the epistemic style of elite discourse has not been examined.

Since 2015, German politics has been reshaped by rising affective polarisation \cite{renstrom-2023, wagner-2020}, manifesting as positive evaluations of political ingroups and negative evaluations of opposing party supporters, especially surrounding the Alternative für Deutschland (AfD, Alternative for Germany). The AfD party, which pivoted from an anti European Union platform to anti-immigration rhetoric during the 2015 refugee crisis, entered Germany's federal parliament, the Bundestag in 2017 \cite{berning2017alternative, lewandowsky-2016b}. In 2025, Germany's domestic intelligence agency (Bundesamt für Verfassungsschutz) classified the party as a "confirmed right-wing extremist" organization, although this designation remains subject to ongoing legal proceedings \cite{Reuters2025AfD}. In parallel, attacks against members of established parties have risen sharply, an escalation that has been linked to an increasingly polarised political climate online and offline \cite{welle-2025, backes-2013, BKA, muller-2020}.

Germany is an informative test case for examining how far-right parties entering parliament affect the rhetorical commitments of political elites. First, Germany's constitutional setup differs considerably from that of the United States: a multiparty system (in which voters cast one vote for a constituency candidate and a second, decisive vote for a party list) and a post-war anti-extremist consensus that, until 2017, kept the federal Bundestag free of any party to the right of the centre-right CDU/CSU. Second, the AfD's 2017 parliamentary entry provides one of the cleanest temporal anchors available in any contemporary democracy: a single, well-dated event introducing a far-right actor into a chamber that had been extreme-right-free since the Nazi era. Parliamentary representation, rather than vote share alone, has been shown to constitute the theoretically decisive watershed in how established parties respond to radical right challengers, as it confers resources, media attention, and credibility as a durable competitor \cite{abouchadi-krause-2020}. Third, as the European Union's largest democracy, evidence from Germany speaks directly to the resilience of European democratic discourse more broadly: European radical right parties have repeatedly adopted the discursive repertoires of successful counterparts abroad rather than developing them independently, so an erosion of evidence-based justification in German elite discourse is unlikely to stay nationally contained \cite{rydgren2005contagious,vandewardt2024contagion}.

We ask whether the rhetorical balance between evidence- and intuition-based language in German political elite communication has shifted between 2015 and 2025, and whether such shifts coincide with the rising visibility of far-right actors. We analysed 4.5~million tweets from over 2{,}000 German politicians and 59{,}170 plenary speeches from the 18th--21st Bundestag, applying a validated Distributed Dictionary Representation (DDR) of evidence- and intuition-based language and testing for systematic differences across ideological camps and for structural shifts coinciding with the AfD's parliamentary entry. 

Our design exploits two complementary observation arenas with opposite failure modes. One arena is politicians' rhetoric on Twitter (renamed X in 2023; we retain the original name, as it applied throughout most of our observation period), and the other is the parliament (Bundestag) itself.  Twitter provides open and high-velocity communication but is subject to self-selection and algorithmic amplification. In contrast, the Bundestag is procedurally rigid and unaffected by platform-side change, but is constrained by formal speech rules. A shift towards an intuition based approach to truth visible in both arenas is therefore unlikely to be an artifact of the specific attributes of either. Ours is among the first studies to use platform data obtained through an official request under the European Union's Digital Services Act (DSA) Article~40 (see Methods). A regulatory regime that, since 2024, has made systematic longitudinal research on very large online platforms feasible at the scale required for the present question. The result is the first cross-arena, multiparty European evidence on the relationship between far-right visibility and the rhetorical commitments to truth of political elites.

\section{Results}\label{sec2}

\subsection{Twitter, Bundestag speeches and ideological grouping}
Twitter is an important platform for political agenda setting, with elites using it strategically through hashtag campaigns, real-time commentary on high-attention events, amplification of external movements \cite{10.3389/fpos.2021.635822} (e.g., Fridays for Future), and metacommunication about campaign developments \cite{barbera_who_2019, lewandowsky_using_2020}. Its public-facing visibility and retweet features create spillover effects to traditional media \cite{doi:10.1287/isre.2019.0897}, making it particularly informative for studying elite communication. We analyse a Twitter corpus of 4.5 million tweets from over 2{,}000 German politicians across eight parties between 2015 and 2025. We update the dataset originally collected by Lasser and colleagues \cite{lasser-2022} by retrieving 3.5 million additional tweets via the official Twitter handles of German parliamentarians elected in 2022 (GESIS handle list) \cite{ZA7721}, supplemented with the handles of the Lasser dataset. This results in a corpus of all tweets posted by every politician who was active during one or more of the given legislative periods of the German Bundestag.

Parliamentary speeches represent the most institutionalised form of political communication in Germany, governed by formal procedures and normative constraints that distinguish them from elite communication on social media. To train an embedding model fine tuned on parliamentary language, we combine two historical German Parliamentary Corpora (1867--2021) \cite{abrami_german_2022, abrami-etal-2024-german, opendiscourse2020}, comprising 329{,}390 speeches by 3{,}786 politicians across 22 parties. This combined corpus is used solely to pre-train the embedding model on a large body of institutional political text. Our empirical analyses draw on a separate set of publicly available Bundestag speeches collected from the official document server (see Methods), comprising 59{,}170 speeches from 7 parties and 1{,}391 politicians delivered during the 18th--21st (2015--2025) electoral terms.

Political leaning was operationalized based on the actor's party affiliation, following the left--right positioning of the Chapel Hill Expert Survey (CHES) \cite{JOLLY2022102420}. Accounts affiliated with the Alternative for Germany (AfD) were classified as far right, those associated with Die Linke and BSW as left-leaning, and all remaining actors as centrist. This aggregation supports analyses of the role of far-right actors relative to the rest of the political spectrum. We additionally report fine-grained party-level analyses in the results.

\subsection{Temporal dynamics of evidence- and intuition-based rhetoric}
To identify evidence- and intuition-based components in the texts, we apply a Distributed Dictionary Representation (DDR) approach following the pipeline of Aroyehun and colleagues \cite{aroyehun-2025}, using two existing English intuition and evidence dictionaries developed and validated within this framework. The German versions of the dictionaries were translated, reviewed, and validated in a pre-registered online survey ($N = 47$; see Methods). The resulting keywords are shown in Extended Data Table~\ref{tab:keywords}. Following the DDR approach, those keywords were converted into embeddings, and the average across embeddings in each dictionary provided a single vector representation that was used in the analysis.

A gensim word2vec model was pre-trained on the historical Bundestag corpus (1867--2025) and fine-tuned separately on tweets and on contemporary Bundestag speeches to obtain platform-specific embeddings. For each text, we computed the cosine similarity between its averaged word-embedding representation and the averaged representation of each dictionary. The two resulting similarity scores per text were independently $z$-standardised across the corpus, and we then computed the \textit{Evidence Minus Intuition} (EMI) score as the difference between the two: texts with $\text{EMI} < 0$ lean towards intuition-based language and texts with $\text{EMI} > 0$ towards evidence-based language. We validated EMI at the document level through a Prolific survey ($N = 284$; see Methods) in which human raters annotated a sample of 1{,}300 parliamentary passages stratified on both dimensions, across time and party affiliation. Using speeches that a majority of human raters agreed were representative of intuition or evidence, we found satisfactory agreement between the computed intuition-based and evidence-based similarity scores and human ratings with an AUC of 0.72 for EMI (see Methods and Supplementary Note 2.11). Extended Data Tables~\ref{tab:intuition_passages} and~\ref{tab:evidence_passages} show examples of passages with high intuition or high evidence loadings. Robustness checks using generic German fastText embeddings yield a correlation of $r = 0.93$ between EMI scores produced by the two embedding models, and bootstrapping the dictionary composition (1{,}000 iterations) confirms that the document-level scores are not driven by a small subset of influential words (see Supplementary Note 2.7).

\begin{figure*}[h]
    \centering
    \includegraphics[width=\textwidth]{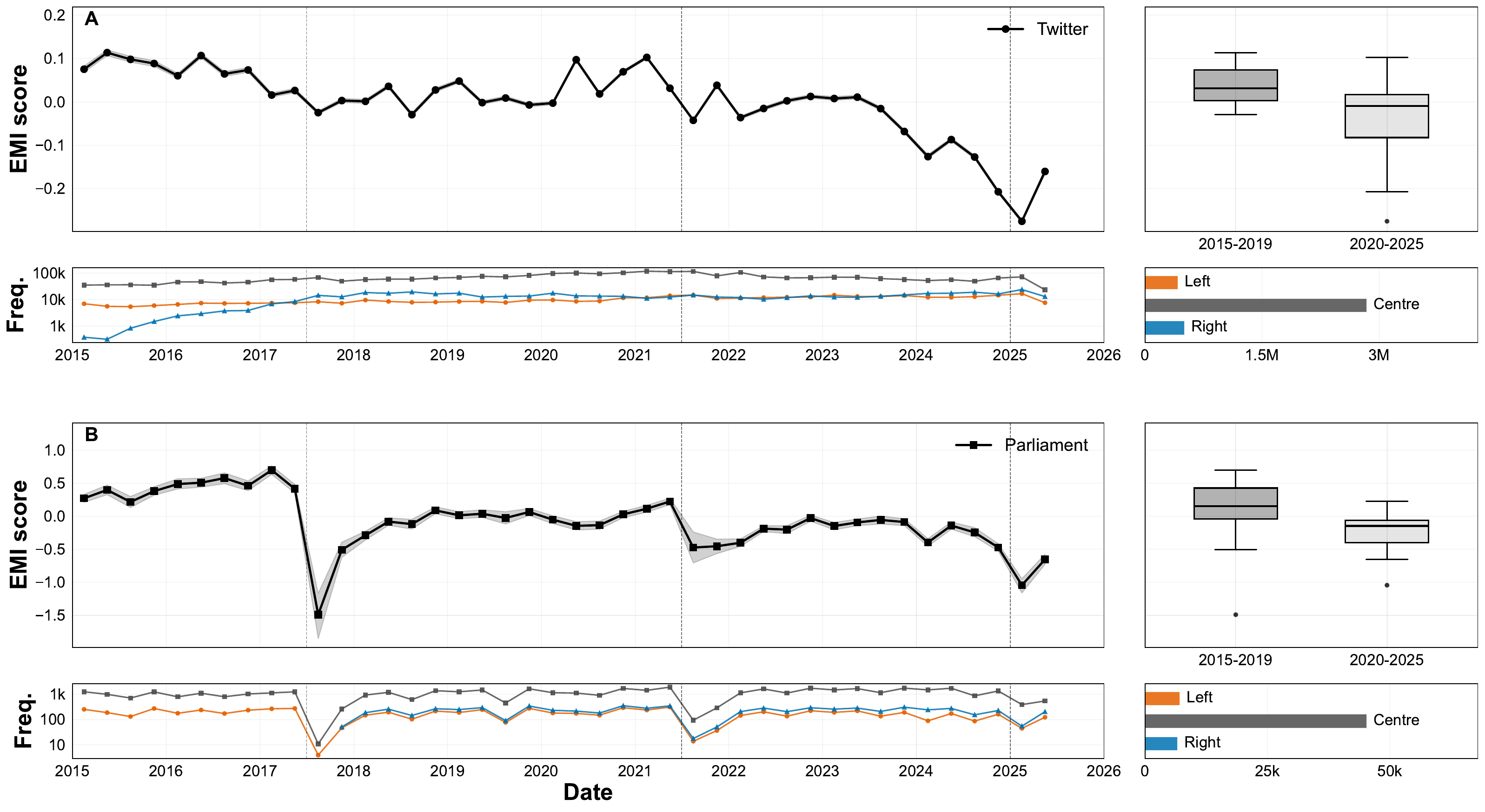}
    \caption{
       \textbf{Temporal trends and density distributions of evidence- and intuition-based language in tweets and parliamentary speeches.} \textbf{A}, Quarterly mean EMI in tweets with log transformed frequency of tweets per leaning, \textbf{B},  Quarterly mean EMI in Parliament speeches with log transformed frequency of speeches per ideology; shaded bands are bootstrapped 95\% confidence intervals (1{,}000 resamples). Dashed vertical lines in \textbf{A} and \textbf{B} mark the election breakpoints (1~July 2017, 1~July 2021, 1~January 2025). \textbf{C},\textbf{D}, Document-level EMI scores on Twitter and in Parliament as Box plots for the periods 2015--2019 and 2020--2025. Box plots show the median (centre line), interquartile range (box bounds, 25th and 75th percentiles) and whiskers extending to the most extreme values within 1.5 $\times$ the interquartile range from the box; points beyond the whiskers are plotted individually as outliers.}
    \label{fig:rq1}
\end{figure*}

We computed the quarterly average similarity of intuition ($\langle D'_i \rangle$), evidence ($\langle D'_e \rangle$), and EMI ($\langle D'_\text{emi} \rangle$) for speech in each corpus using a 60-day rolling window. Figure~\ref{fig:rq1} shows how these distributions shifted between 2015 and 2025. The 2019 cutoff falls after the AfD's initial entry into the Bundestag in late 2017 and roughly coincides with the subsequent emergence of new far-right Twitter accounts. Rhetoric scores shift towards the intuition-based end of the scale over time on both platforms. Time-series trends are shown relative to the four-yearly federal elections, which we treat as epistemic shocks. Across both arenas, EMI differs significantly between the early (2015--2019) and late (2020--2025) periods. On Twitter, mean EMI declined from $0.027$ to $-0.035$ ($t = 78.58$, $P < 0.001$, Cohen's $d = -0.083$, 95\% CI$_\text{early} = [0.026, 0.028]$, 95\% CI$_\text{late} = [-0.036, -0.034]$; Mann--Whitney $U = 1.77 \times 10^{12}$, $P < 0.001$). Parliamentary speeches show a substantially larger decrease, from $0.205$ to $-0.147$ ($t = 33.75$, $P < 0.001$, Cohen's $d = -0.281$, 95\% CI$_\text{early} = [0.190, 0.221]$, 95\% CI$_\text{late} = [-0.160, -0.134]$; Mann--Whitney $U = 4.92 \times 10^{8}$, $P < 0.001$).

\subsection{Partisan patterns in evidence- and intuition-based rhetoric}

\begin{figure}[htp]
    \centering
    \includegraphics[width=\textwidth]{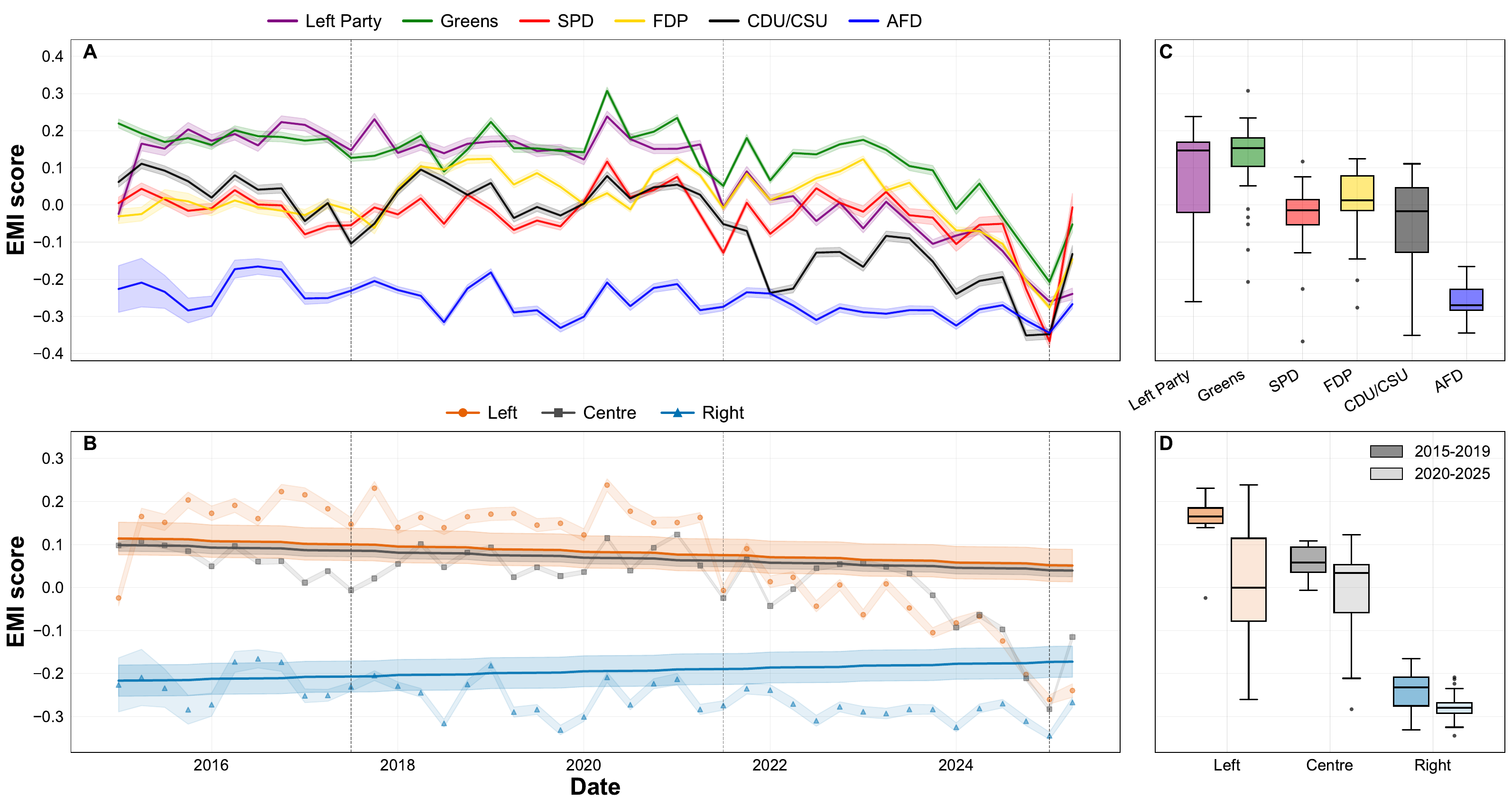}
    \caption{\textbf{Temporal dynamics of Evidence Minus Intuition (EMI) scores by political leaning on Twitter, 2015--2025.}  \textbf{A}, Quarterly mean EMI per party with bootstrapped 95\% confidence intervals (1{,}000 resamples; shaded bands).
    \textbf{B}, Quarterly mean EMI per leaning with bootstrapped 95\% confidence intervals and fitted from the linear mixed-effects model per leaning (Equation \ref{eq:1}); shaded bands are 95\% confidence intervals
    obtained via the delta method.
    Dashed vertical lines in \textbf{A} and \textbf{B} mark the model's
    election breakpoints (1~July 2017, 1~July 2021, 1~January 2025).
    \textbf{C}, Document level EMI as Boxplots per party
    \textbf{D},\,\textbf{E},  Document level EMI as Boxplots per leaning before (\textbf{D}; 2015--2019) and from (\textbf{E}; 2020--2025)
    the sample split.
    Party abbreviations: CDU/CSU, Christian Democratic Union/Christian
    Social Union; SPD, Social Democratic Party of Germany; AfD, Alternative
    for Germany; FDP, Free Democratic Party.
    }
    \label{fig:emi_twitter}
\end{figure}

\begin{figure}[htp]
    \centering
    \includegraphics[width=\textwidth]{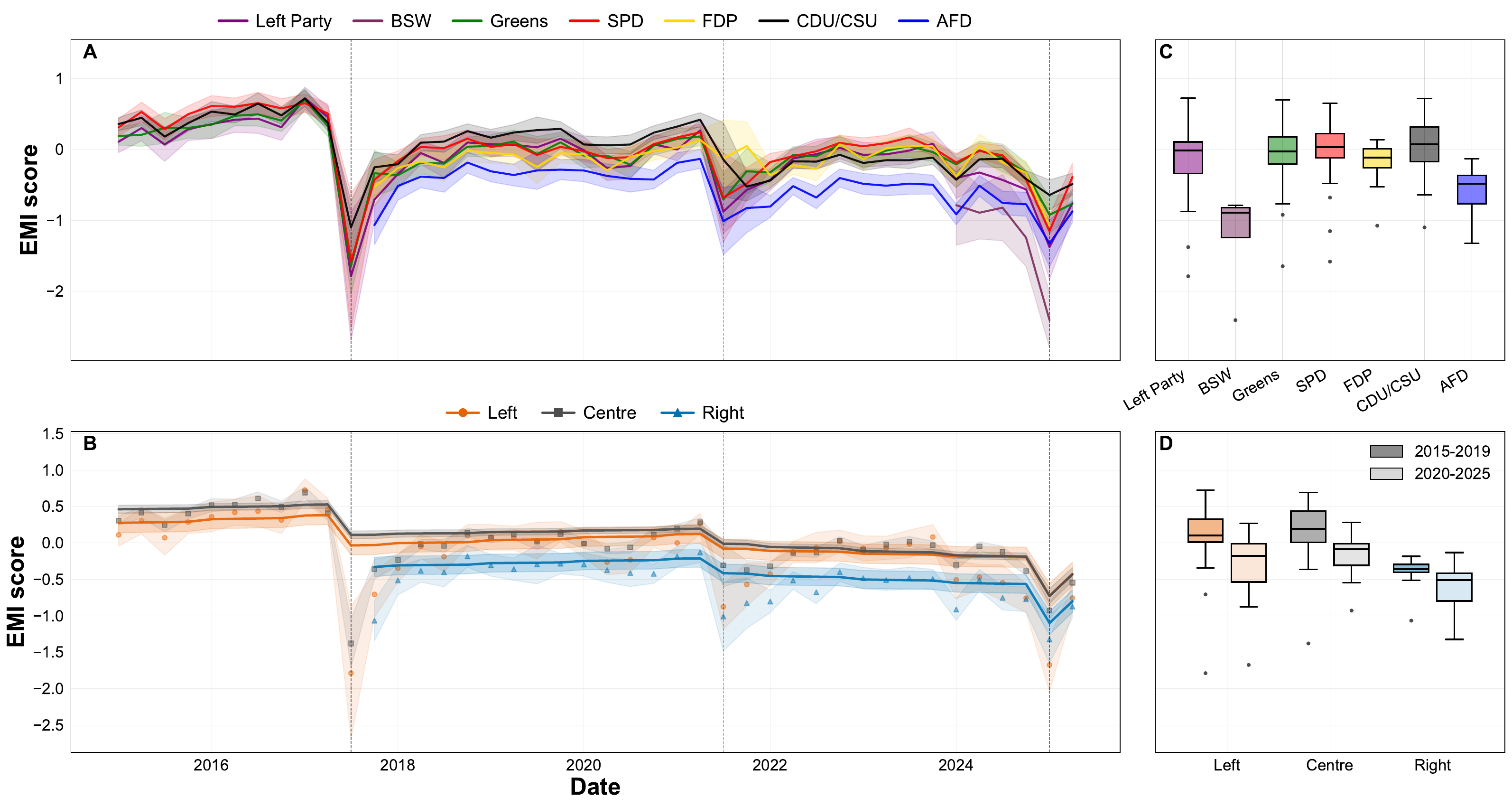}
   \caption{\textbf{Temporal dynamics of Evidence Minus Intuition (EMI) scores by political leaning in the German Bundestag, 2015--2025.}  \textbf{A}, Quarterly mean EMI per party with bootstrapped 95\% confidence intervals (1{,}000
    resamples; shaded bands).
    \textbf{B}, Quarterly mean EMI per leaning with bootstrapped 95\% confidence intervals and fitted trajectories from the piecewise linear mixed-effects
    breakpoint model per leaning (Equation \ref{eq:2}); shaded bands are 95\% confidence intervals
    obtained via the delta method.
    Dashed vertical lines in \textbf{A} and \textbf{B} mark the model's
    election breakpoints (1~July 2017, 1~July 2021, 1~January 2025).
    \textbf{C}, Document level EMI as Boxplots per party
    \textbf{D},\,\textbf{E},  Document level EMI as Boxplots per leaning before (\textbf{D}; 2015--2019) and from (\textbf{E}; 2020--2025)
    the sample split. The Bundestag's summer recess ran into the 24 September 2017 federal election. The old Bundestag held its last sitting week in early September, and the new one only constituted on 24 October, so there were almost no plenary speeches that quarter explaining the gap in the data. Party abbreviations: CDU/CSU, Christian Democratic Union/Christian
    Social Union; SPD, Social Democratic Party of Germany; AfD, Alternative
    for Germany; FDP, Free Democratic Party; BSW, Sahra Wagenknecht Alliance.
    }
    \label{fig:emi_parliament}

\end{figure}

Party-level mean EMI scores are reported in Extended Data Tables~\ref{tab:emi_twitter} and \ref{tab:emi_speeches}; pairwise differences are statistically significant most combinations expect Greens vs. Left Party and CDU/CSU vs. SPD (Tukey's HSD post-hoc test). A consistent ordering emerges across platforms visualized in Figures~\ref{fig:emi_twitter} and \ref{fig:emi_parliament}. On Twitter, mean EMI is highest for Alliance~90/The~Greens ($0.135$, 95\% CI $[0.133, 0.136]$), followed by Die Linke ($0.049$, 95\% CI $[0.047, 0.051]$) and the FDP ($0.011$, 95\% CI $[0.009, 0.012]$); the SPD ($-0.019$, 95\% CI $[-0.020, -0.017]$) and CDU/CSU ($-0.041$, 95\% CI $[-0.042, -0.039]$) show small negative values; the AfD ranks lowest by a substantial margin ($-0.268$, 95\% CI $[-0.270, -0.266]$).

Parliamentary speeches show a similar but partly re-ordered pattern. The SPD shows the highest mean EMI ($0.113$, 95\% CI $[0.092, 0.135]$), followed by CDU/CSU ($0.108$, 95\% CI $[0.089, 0.127]$), Die Linke ($0.051$, 95\% CI $[0.021, 0.080]$), and Alliance~90/The~Greens ($0.039$, 95\% CI $[0.014, 0.065]$); a negative value is observed for the FDP ($-0.110$, 95\% CI $[-0.140, -0.081]$). The AfD again sits near the bottom ($-0.473$, 95\% CI $[-0.502, -0.443]$); B\"undnis Sahra Wagenknecht (BSW), which was founded in January 2024 and thus entered the chamber late in the observation window with a small sample of speeches ($n = 153$), shows the lowest mean overall ($-1.098$, 95\% CI $[-1.311, -0.885]$). At the ideological-camp level (Extended Data Tables~\ref{tab:emi_twitter}--\ref{tab:emi_speeches}), the same hierarchical ordering holds across both platforms: right-leaning actors consistently display the lowest mean EMI. Figures~\ref{fig:emi_twitter} and \ref{fig:emi_parliament}, the quarterly aggregated means (Panel A) and reveal that, over time, the centrist and left-leaning camps shift towards the lower EMI levels of the right-leaning camp. This convergence is gradual and approximately linear on Twitter, whereas in the Bundestag it is characterised by abrupt level shifts that align temporally with changes in the composition of the Bundestag, especially with electoral entries of new parties.

To formally test these dynamics, we fit a sequence of linear mixed-effects models with random intercepts and slopes at the actor level, controlling for within-actor temporal autocorrelation via a one-quarter lagged EMI term (Methods, Equations~\ref{eq:1}--\ref{eq:3}). For parliamentary speeches we additionally introduced piecewise structural breakpoints at each German federal election (2017, 2021, 2025), first constraining the breakpoint effects to be uniform across camps and then allowing them to vary by ideology. To rule out that the estimated time and breakpoint effects reflect a shift in topic mix rather than in epistemic style, each of these specifications was additionally re-fit with document-level topic fixed effects derived from a BERTopic model (See Methods and Supplementary Note 4); the time-trend, leaning and constrained breakpoint coefficients reported below are robust to this control (See Supplementary Note 4) and EMI follows similar trends across topics with some outliers (Extended Data Figures~\ref{fig:parliament_topic} and ~\ref{fig:twitter_topic}).

On Twitter (Extended Data Table~\ref{tab:model1}, Figure~\ref{fig:emi_twitter} Panel B), right-leaning actors exhibit substantially lower baseline EMI ($\beta = -0.111$, $P < 0.001$, 95\% CI $[-0.149, -0.072]$) than the centrist reference category ($\beta = 0.050$, $P < 0.001$, 95\% CI $[0.036, 0.064]$), while left-leaning actors do not differ from baseline ($\beta = 0.001$, $P = 0.945$, 95\% CI $[-0.039, 0.041]$). EMI shows a small but statistically robust overall decline over time ($\beta = -0.001$, $P < 0.001$, 95\% CI $[-0.001, -0.001]$). This decline is partially offset for right-leaning actors, who show a significantly more positive time trend relative to the centre ($\beta = 0.001$, $P < 0.001$, 95\% CI $[0.001, 0.001]$); the interaction with left-leaning actors is not significant ($P = 0.494$). Higher tweet engagement (log like count) is associated with lower EMI ($\beta = -0.040$, $P < 0.001$, 95\% CI $[-0.040, -0.039]$), and the lagged EMI term shows strong temporal persistence ($\beta = 0.495$, $P < 0.001$, 95\% CI $[0.490, 0.500]$).
\begin{figure*}[h]
    \centering
    \includegraphics[width=\textwidth]{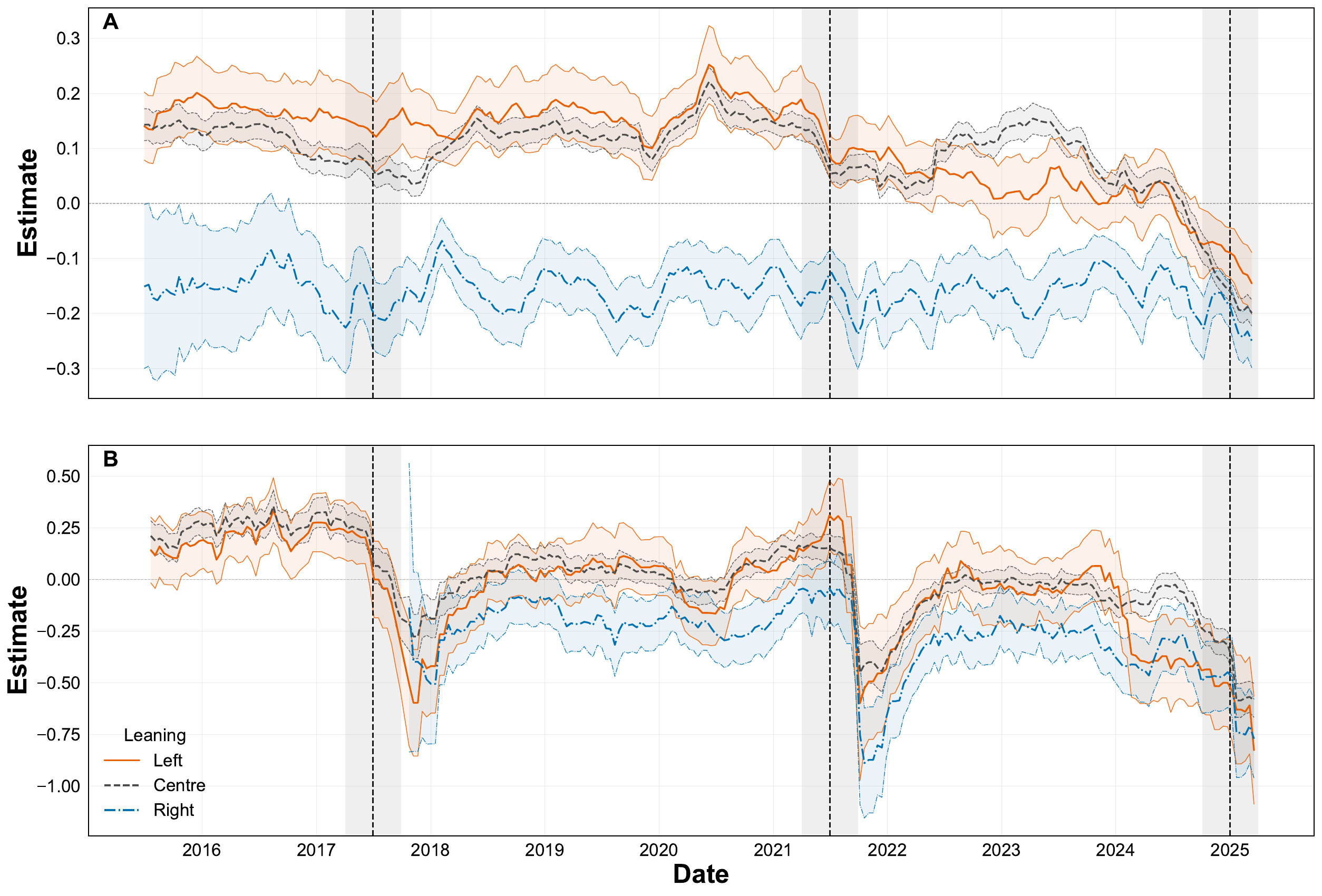}
    \caption{
        \textbf{Rolling-window estimates of evidence-minus-intuition
    (EMI) scores by ideological leaning on Twitter and in the German
    Parliament.}
    \textbf{A}, Leaning-specific conditional mean EMI on Twitter  and \textbf{B}, in
    Parliament, estimated from a linear mixed-effects model fitted within sliding 180-day windows
    advanced in 14-day steps. Lines show point estimates for Left, Centre and
    Right; shaded bands are 95\% Wald confidence intervals. Grey vertical
    shading spans one window width around each election breakpoint
    (1~July 2017, 1~July 2021, 1~January 2025), where window-mean estimates
    mix pre- and post-breakpoint observations.}
    \label{fig:rolling}
\end{figure*}

For parliamentary speeches (Extended Data Tables~\ref{tab:model2} and \ref{tab:model3}, Figure~\ref{fig:emi_parliament} Panel B), the breakpoint regression reveals significant level shifts at each electoral cycle. With breakpoint effects allowed to vary by camp, right-leaning actors exhibit substantially lower baseline EMI ($\beta = -0.347$, $P < 0.001$, 95\% CI $[-0.480, -0.213]$); left-leaning actors do not differ from the centre at baseline ($\beta = 0.015$, $P = 0.930$). The 2017 election is associated with a significant downward level shift ($\beta = -0.427$, $P < 0.001$, 95\% CI $[-0.485, -0.369]$) without a significant slope change ($P = 0.780$); the camp-specific deviations are not significant. The 2021 election is associated with a further negative level shift ($\beta = -0.218$, $P < 0.001$, 95\% CI $[-0.269, -0.167]$) and a negative slope change ($\beta = -0.007$, $P < 0.001$, 95\% CI $[-0.009, -0.005]$); the slope decline is further amplified among left-leaning actors ($\beta = -0.010$, $P = 0.006$, 95\% CI $[-0.017, -0.003]$). The 2025 election shows a strongly negative level shift ($\beta = -0.490$, $P < 0.001$, 95\% CI $[-0.600, -0.381]$) but a significantly positive slope change ($\beta = 0.277$, $P < 0.001$, 95\% CI $[0.137, 0.416]$), driven primarily by left-leaning actors ($\beta = 0.757$, $P = 0.001$, 95\% CI $[0.309, 1.205]$); these post-2025 slope estimates should be interpreted with caution because of the sparse post-breakpoint data. The lagged EMI term is again positive and significant ($\beta = 0.124$, $P < 0.001$, 95\% CI $[0.111, 0.137]$). Equations \ref{eq:2} and \ref{eq:3} impose a piecewise-linear functional form on the EMI trajectories and represent camp differences as constant offsets. To trace those differences without imposing this form, we fitted a simplified linear mixed-effects model within sliding 180-day windows advanced in 14-day steps, entering leaning without a global intercept so that each coefficient estimates that camp's covariate-adjusted mean EMI within the window (See Methods). The resulting trajectories reproduce the before exogenous assumed temporal patterns of the main analysis and make the contrast between the two arenas directly visible in Figure \ref{fig:rolling}. On Twitter (Panel A), the centrist and left-leaning camps decline steadily while right-leaning actors remain close to their initial level, narrowing the gap between the centre and the right over the observation period. In the Bundestag (Panel B), the camps instead step down together at the 2017 breakpoint, with the distance between them largely preserved; narrowing appears only later in the window. In neither arena does the ordering of the camps invert.

\section{Discussion}\label{sec3}

We examine the emergence of extreme right-wing actors in Germany and a concerning epistemic shift in which truth-seeking is increasingly dominated by intuition rather than evidence. Across two complementary arenas of elite political communication in Germany (a strategically curated social-media platform and a highly institutionalised parliamentary chamber), we observe a consistent shift towards intuition-based rhetoric between 2015 and 2025. Right-leaning actors, in particular the AfD, exhibit the lowest EMI values throughout the period, and the broader political discourse co-moves with their increasing visibility. On Twitter this co-movement takes the form of a gradual decline in EMI, consistent with low barriers to participation and faster diffusion of rhetorical conventions \cite{carrella-2025}. In the Bundestag it manifests as discrete level shifts that align temporally with each electoral cycle, with the most pronounced single shift coinciding with the AfD's parliamentary entry in 2017. This dual pattern is consistent with two complementary mechanisms of rhetorical convergence: continuous social diffusion online and compositional adaptation within the chamber following the entry of new actors.

These findings extend prior work that documented a long-term decline in evidence-based language in U.S.\ congressional speeches since the mid-1970s \cite{aroyehun-2025} and a parallel cross-country pattern in parliamentary discourse \cite{aroyehun-2026, lasser-2023}. They also extend the party-competition literature on far-right ``contagion.'' Prior work documented that mainstream parties respond to radical right success by accommodating its policy positions \cite{vanspanje-2010, abouchadi-krause-2020} and by adopting its issue priorities \cite{gessler-hunger-2022, voelker-saldivia-2024}. Our results suggest that contagion may extend to a third dimension that is conceptually orthogonal to both position and topic. we find a chamber-wide and largely party uniformly shift in 2017 shwon by the finding that both mainstream left and mainstream right shift positions once a radical right party enters parliament \cite{abouchadi-krause-2020}. Though mainstream parties differ considerably in their formal engagement strategies toward the AfD \cite{heinze-2022}. The German case is informative on three counts. First, it demonstrates that the shift is not specific to the U.S.\ two-party system but emerges within a multiparty European democracy. Second, the relatively recent and well-dated entry of a single far-right party into a previously six-party chamber provides a sharper temporal anchor than is available in most multiparty contexts. Third, the cross-platform replication, in two communicative arenas with very different participation rules, suggests the dynamic is not an artifact of a single platform's affordances. At the same time, the structural breaks at the 2017 electoral cycle are co-temporal with other shocks in our window, notably the COVID-19 pandemic, the Russian invasion of Ukraine, and pronounced changes in social-media platform governance, so we cannot attribute the observed shifts to AfD entry alone. The patterns we report are co-movements consistent with a diffusion or adaptation hypothesis. Stronger causal claims will require designs that exploit exogenous variation in actor visibility (for example, regional discontinuities in AfD success) and explicit network-exposure measures.

Several mechanisms could plausibly underlie the observed convergence towards intuition based language. First, strategic rhetorical mimicry: established actors may adopt elements of the populist rhetorical style in response to perceived voter demand or competitive pressure \cite{decker-2023, Bohmer2024Strack}, mirroring the accommodative strategies and positional contagion documented in the party-competition literature \cite{meguid-2005, vanspanje-2010, abouchadi-krause-2020}. Notably, this literature implies that even \emph{adversarial} engagement can pull the mainstream onto the challenger's terrain, since confronting a niche party raises the salience of its issues \cite{meguid-2005}. Second, issue-level agenda-setting: when far-right actors successfully shift the topical agenda toward immigration, identity, and contested values \cite{gessler-hunger-2022, voelker-saldivia-2024}, all participants in the debate must speak about topics that are more amenable to intuition-based framing. Third, platform amplification: algorithms that prioritise emotionally resonant content can disproportionately surface intuition-based posts, indirectly shaping the rhetorical environment to which other actors respond \cite{simchon-2022, lazer-2018}. Our data cannot fully adjudicate between these mechanisms, but two observations are informative. The contrast between the gradual Twitter decline and the discrete parliamentary shifts is more easily reconciled with strategic adaptation and issue-level agenda-setting in the chamber and with diffusion online. Moreover, because the time-trend and breakpoint estimates are robust to document-level topic fixed effects, a pure agenda-setting account (in which the shift reflects only a changed topic mix of the kind documented during the refugee crisis \cite{gessler-hunger-2022}) cannot fully explain our results: the movement toward intuition-based rhetoric occurs \emph{within} most topics, indicating a genuinely stylistic component.

Our findings also speak to ongoing debates about the relationship between rhetorical style and democratic deliberation. Evidence-based rhetoric supports a shared factual baseline that is argued to be necessary for meaningful democratic debate \cite{lewandowsky-2024}, and prior work has linked declining evidence-based language to higher polarisation and to the sharing of low-quality information sources \cite{aroyehun-2025, lasser-2022}. The German shift we document occurs alongside a documented rise in attacks against members of established parties \cite{welle-2025, BKA} and rising affective polarisation \cite{renstrom-2023, wagner-2020}. We do not claim a causal pathway from one to the other, but the joint pattern motivates closer study of how rhetorical drift, polarisation, and political violence co-evolve.

Our camp-level aggregation necessarily obscures heterogeneity within the left, centre and right blocs; the centrist camp in particular combines actors with substantively different rhetorical profiles. Party-level analyses recover the same clustering while preserving these party-specific differences, indicating that aggregation does not distort the pattern we report, and the position of the most extreme actors is unaffected. Our analysis is restricted to elite communication and cannot speak to whether comparable shifts have occurred in mass-public discourse. 

Future work should test whether network exposure to AfD content predicts within-actor shifts in EMI for other politicians, an analytic step that would strengthen claims about diffusion mechanisms beyond the co-movement results we report here. Cross-country comparisons that exploit different timings of far-right entry into national parliaments could provide a quasi-experimental test of the convergence hypothesis at scale.

Taken together, our results point to a dimension of elite discourse that changes alongside far-right visibility and that current accounts of democratic resilience largely overlook. Mainstream responses to radical right challengers have been studied as contests over positions and issues, and the corresponding remedies, from fact-checking to corrections and platform moderation, operate at the level of individual claims. What we observe operates at the level of epistemic reasoning. Between 2015 and 2025, the epistemic style of German elite discourse moved towards intuition in both arenas we examined, and it did so most sharply in the Bundestag, the arena with the strongest formal speech norms and the longest standing consensus against the far right. The 2017 shift was chamber-wide, with no significant differences between ideological camps, even though mainstream parties differ considerably in how they formally engage with the AfD \cite{heinze-2022}. Where the decline of evidence-based language in the U.S. Congress unfolded over roughly half a century \cite{aroyehun-2025}, comparable movement appears here within a single decade, in discrete steps aligned with changes in the composition of the chamber.

None of this argues for a politics of evidence alone. Intuition carries moral and experiential content that democratic debate cannot do without, and it is over-reliance on it at the expense of evidence that is of concern \cite{aroyehun-2026}. But where evidence can no longer adjudicate between competing positions, accountability, informed policy evaluation and cross-partisan deliberation lose their footing \cite{aroyehun-2026, aroyehun-2025, lasser-2023}. Democracies can withstand deep disagreement about what the facts are. It is less clear that they can withstand disagreement whether facts themselves are a foundational requirement of democratic deliberation. 

\section{Methods}\label{sec4}
\subsection{Twitter corpus}
The Twitter corpus combines two sources. The first is the German subset of the multi-country politician corpus assembled by Lasser and colleagues \cite{lasser-2023}, which was collected by identifying verified Twitter accounts of national politicians and retrieving their tweets via the Twitter API; we use the German subsample from 2008--2021. The second source extends this corpus forward in time by retrieving all tweets posted between 2015 and 2025 from the verified handles of members of the 20th German Bundestag elected in 2022, using the GESIS handle list \cite{ZA7721}, merged with the handles in the Lasser dataset to maximise coverage of currently and previously serving parliamentarians. Tweets were retrieved between 1~June~2025 and 30~September~2025 via the Twitter (X) Academic API operating under the European Union's Digital Services Act (DSA, Regulation (EU) 2022/2065) Article~40 access mechanism, which obliges very large online platforms to provide qualified researchers with legally mandated access to publicly-accessible platform data for the study of systemic risks. Where the same tweet was present in both source datasets, the entry was de-duplicated by tweet ID. The final dataset contains 4.5~million tweets from 2{,}067 politician accounts across 8 parties.

\subsection{Parliamentary speech corpora}
We use two parliamentary corpora. The \textit{historical corpus} is used solely to pre-train the domain-specific word-embedding model and merges the German Parliamentary Corpus covering 1867--1942 \cite{abrami_german_2022, abrami-etal-2024-german} with the Open Discourse Bundestag corpus covering 1949--2021 \cite{opendiscourse2020}, yielding 329{,}390 speeches by 3{,}786 politicians across 22 parties. The \textit{contemporary corpus} is the unit of analysis for the empirical results: it consists of all plenary protocols of the 18th--21st German Bundestag (2015--2025), downloaded from the official Bundestag document server \cite{bundestag_plenarprotokolle}. For each electoral term we retrieved the available XML transcripts and parsed session metadata (date, session number, electoral term), speaker information (name, faction, role) and the full text of each speech. Each speech was assigned a unique identifier and linked to its source PDF. The contemporary corpus comprises 59{,}170 speeches (after removing procedural speeches) by 1{,}391 politicians across 7 parties.

\subsection{Dictionary translation and validation}
The intuition and evidence dictionaries used in this study were originally developed  and validated in English by Aroyehun and colleagues \cite{aroyehun-2025}. Briefly, a manually selected seed list by experts (to illustrate, initial keywords for evidence included terms such as ``reality'', ``assess'', ``examine'', ``evidence'', ``fact'', ``truth'', ``proof'' and so on; for intuition, initial keywords were terms such as ``believe'', ``opinion'', ``consider'', ``feel'', ``intuition'' or ``common sense'') was expanded computationally using fastText word embeddings with a cosine-similarity threshold of 0.75 and refined using cross-linguistic colexification networks; overlapping, duplicate and inflected forms were removed, and a human validation study was used to remove or reassign terms that participants did not align with their assigned category.

For the present study we translated the validated English dictionaries into German and re-validated them using an online survey on Prolific in which participants scored each candidate term on two five-point Likert scales (representativeness for intuition; representativeness for evidence). Survey data were collected on 21 March 2024 from 47 German-speaking participants (35~male, 12~female; mean age 28.9 years, s.d.\ 7.4). For each term we computed paired $t$-tests of intuition versus evidence ratings, with $p$-values Holm-corrected for multiple comparisons. Of 113 candidate keywords, 86 were retained as significantly aligning with their assigned category; 27 did not reach the corrected significance threshold ($p < 0.05$) and were removed. The final dictionaries are given in Extended Data Table~\ref{tab:keywords}; full instructions and demographics of participants are provided in Supplementary Note 2.10.

\subsection{Identification of evidence- and intuition-based components in text}
We use the Distributed Dictionary Representations (DDR) approach, in which a predefined dictionary is used to quantify the semantic similarity between a text and a target rhetorical construct via word embeddings, rather than via raw word counts. We trained a domain-specific gensim word2vec model on the historical Bundestag corpus (1867--2025 combined) and fine-tuned the resulting model separately on the Twitter and contemporary parliamentary corpora to obtain platform-specific embeddings (training hyperparameters are reported in Supplementary Note 2). The EMI score for each text is computed in three steps.

First, for each text \(t\), we  calculate its distance \(D\) to the evidence and intuition dictionaries. The distance \(D_i(t)\) measures the semantic similarity between the averaged embeddings of the text and the dictionary:
\[
D_i(t) = \text{cos-sim}\left(\frac{1}{|t|} \sum_{w \in t} \mathbf{v}(w), \frac{1}{|i|} \sum_{w \in i} \mathbf{v}(w)\right),
\]
where:
\begin{itemize}
    \item \(t\) is the set of words in the input text,
    \item \(i\) is the set of words in the dictionary (\emph{evidence} or \emph{intuition}),
    \item \(\mathbf{v}(w)\) is the embedding vector of word \(w\),
    \item \(|t|\) and \(|i|\) are the number of words in \(t\) and \(i\), respectively.
\end{itemize}

The cosine similarity \(\text{cos-sim}\) is calculated as:
\[
\text{cos-sim}(\mathbf{u}, \mathbf{v}) = \frac{\mathbf{u} \cdot \mathbf{v}}{\|\mathbf{u}\| \|\mathbf{v}\|},
\]
where:
\begin{itemize}
    \item \(\mathbf{u} \cdot \mathbf{v} = \sum_{j=1}^n u_j v_j\) is the dot product of the two vectors,
    \item \(\|\mathbf{u}\| = \sqrt{\sum_{j=1}^n u_j^2}\) and \(\|\mathbf{v}\| = \sqrt{\sum_{j=1}^n v_j^2}\) are their magnitudes.
\end{itemize}

Next, we z-score normalize the distances to the evidence and intuition dictionaries across all texts. This normalization ensures comparability between scores by centering and scaling the values. Finally, we calculate the EMI score for each text as the difference between the z-scored distances:
\[
\text{EMI}(t) = z(D_\text{evidence}(t)) - z(D_\text{intuition}(t)),
\]
where \(z(\cdot)\) represents the z-score transformation:
\[
z(x) = \frac{x - \mu}{\sigma},
\]
with \( \mu \) and \( \sigma \) being the mean and standard deviation of \( x \), respectively. To analyse trends over time, we calculate the average EMI score for all texts in a given period:
\[
\langle \text{EMI} \rangle = \frac{1}{|T|} \sum_{t \in T} \text{EMI}(t),
\]
where:
\begin{itemize}
    \item \(T\) is the set of texts in the period,
    \item \(|T|\) is the total number of texts.
\end{itemize}

This approach allows us to quantify the rhetorical alignment of texts with evidence or intuition-based communication styles over time.

\subsection{Document-level validation}
The DDR pipeline was validated through three complementary procedures. First, all EMI scores were recomputed using the publicly available generic German fastText embeddings; the correlation between the gensim-based and fastText-based document-level EMI scores was $r = 0.93$ (95\% CI $[0.91, 0.94]$), indicating that the substantive results are not specific to the fine-tuned embedding model. Second, the contribution of individual dictionary terms to the document-level scores was assessed via 1{,}000 bootstrap resamples of the dictionary composition; document-level scores remained stable across resamples (Supplementary Note 2.8), indicating that no small subset of highly influential terms is driving the observed patterns. Third, we conducted a pre-registered human validation study on Prolific in which 284 German-speaking participants annotated a stratified sample of 1{,}300 parliamentary passages (each 10--50 tokens long), drawn equally across four EMI bins for each of the three constructs (evidence, intuition, EMI), with no duplicates. We followed the validation procedure of Lasser and colleagues \cite{lasser-2023}; full instructions are reproduced in Supplementary Note 2.11. An attention check (an example item for each construct, repeated within the survey) was used to screen participants; 32 participants failed and were excluded, leaving 252 valid responses. Survey data were collected on 11~February~2025. Single-rater intraclass correlation coefficients (ICC$_\text{single}$) were 0.14 (95\% CI $[0.10, 0.18]$) for evidence and 0.16 (95\% CI $[0.12, 0.21]$) for intuition, rising to ICC$_\text{average}$ values of 0.62 (95\% CI $[0.53, 0.69]$) and 0.65 (95\% CI $[0.57, 0.72]$), respectively, when averaged across raters. These modest single-rater ICCs are consistent with the inherent ambiguity of short political passages and motivate the use of aggregate-level analyses throughout the paper. To obtain per-dimension ground truth, individual ratings were dichotomised into endorsement (agree or strongly agree) versus non-endorsement, and a chunk was labelled evidence-based (or intuition-based) if a majority of its annotators endorsed that dimension. The EMI ground truth was constructed differently: rather than dichotomising each rating, we averaged the raw ratings per dimension across annotators, took their difference (survey EMI = mean evidence - mean intuition) and labelled a chunk evidence-dominant where this difference exceeded zero. The computed scores discriminated the human labels above chance: AUC = 0.68 for intuition and 0.59 for evidence, and 0.72 for EMI against the binarized survey-based EMI (See Supplementary Note 2.11). That the difference score outperforms either component is expected if annotators, like the embeddings, treat the two constructs as partly overlapping rather than mutually exclusive; the contrast is more reliably identified than either level.

\subsection{Regression}
We chose a quarterly data aggregation to reduce noise, capture long-term trends in rhetorical shifts, and align with political and media cycles while ensuring model stability and interpretability. To examine how Evidence Minus Intuition (EMI) scores relate to time and political leaning, and whether German federal elections act as structural breakpoints in those trajectories, we estimated a sequence of three mixed-effects regression models of increasing complexity.

\textbf{Baseline model.}
We first established a baseline by regressing EMI scores on time, political leaning, and their interaction, while controlling for within-actor temporal autocorrelation and stable between-actor differences:
\begin{align}
\label{eq:1}
\text{EMI}_{ij} &= \beta_0 + \beta_1 \cdot \text{time}_{i} + \beta_2 \cdot \text{leaning}_{j} \notag \\
&\quad + \beta_3 \cdot (\text{time}_{i} \cdot \text{leaning}_{j}) + \lambda \cdot \overline{\text{EMI}}_{j,t-1} + u_j + \epsilon_{ij}.
\end{align}
Here $\text{EMI}_{ij}$ is the Evidence Minus Intuition score for document $i$ produced by actor $j$ (e.g.,\ a speech by a parliamentarian or a tweet by a political account); $\beta_0$ is the intercept reflecting the baseline EMI score for the reference group at the mean observation time; $\beta_1 \cdot \text{time}_{i}$ captures the overall linear trend in EMI over time (measured in quarters); $\beta_2 \cdot \text{leaning}_{j}$ encodes the main effect of an actor's political leaning (left, centre, or right, with centre as the reference category); and $\beta_3 \cdot (\text{time}_{i} \cdot \text{leaning}_{j})$ allows the temporal trend to differ across ideological groups. The autoregressive term $\lambda \cdot \overline{\text{EMI}}_{j,t-1}$ is the actor's mean EMI in the quarter preceding document $i$, which absorbs within-actor serial correlation and ensures that time-trend estimates reflect genuine change rather than carry-over from earlier behaviour. The random intercept $u_j \sim \mathcal{N}(0, \sigma^2_u)$ accounts for stable, unmeasured between-actor differences in baseline EMI, acknowledging, for example, that individual politicians or content creators may differ systematically in their reliance on evidence- versus intuition-based language independently of time or leaning. $\epsilon_{ij}$ is the residual error, assumed to be normally distributed.

\textbf{Breakpoint model.}
Building on the baseline, we next tested whether EMI trajectories exhibit structural breaks around each German federal election. To do so, we aggregated EMI to quarterly actor-level means $\overline{\text{EMI}}_{jt}$ and introduced piecewise linear regressors for each election year $k \in \{2017, 2021, 2025\}$, with breakpoint $\tilde{b}_k$ set approximately one month before the election date. Time was mean-centred ($\tilde{t}$) to improve numerical stability and interpretability of the intercept:
\begin{align}
\label{eq:2}
\overline{\text{EMI}}_{jt} &= \alpha + \beta_1 \tilde{t} + \beta_2 \cdot \text{leaning}_{j} + \beta_3 \cdot (\tilde{t} \cdot \text{leaning}_{j}) \notag \\
&\quad + \sum_{k} \Bigl[
    \gamma_k \cdot \max(0,\, \tilde{t} - \tilde{b}_k)
  + \delta_k \cdot \mathbf{1}(\tilde{t} \geq \tilde{b}_k)
\Bigr] \notag \\
&\quad + \lambda \cdot \overline{\text{EMI}}_{j,t-1} + u_j + \epsilon_{jt}.
\end{align}
The baseline linear trend and leaning effects from Equation~\ref{eq:1} are retained. For each breakpoint $k$, two complementary regressors are added: $\gamma_k \cdot \max(0,\, \tilde{t} - \tilde{b}_k)$ is a piecewise slope-change term that equals zero before the breakpoint and increases linearly thereafter, capturing a durable shift in the rate of EMI change following an election cycle; $\delta_k \cdot \mathbf{1}(\tilde{t} \geq \tilde{b}_k)$ is a level-shift indicator capturing an abrupt step change in mean EMI at the breakpoint. In this model the breakpoint effects are constrained to be uniform across ideological groups, providing an estimate of the average electoral shock on the EMI trajectory irrespective of leaning.

\textbf{Full breakpoint model with leaning interactions.}
Finally, to test whether electoral breakpoints modulate EMI differently for left-, centre-, and right-leaning actors, we extended Equation~\ref{eq:2} by fully interacting both breakpoint regressors with political leaning:
\begin{align}
\label{eq:3}
\overline{\text{EMI}}_{jt} &= \alpha + \beta_1 \tilde{t} + \beta_2 \cdot \text{leaning}_{j} + \beta_3 \cdot (\tilde{t} \cdot \text{leaning}_{j}) \notag \\
&\quad + \sum_{k} \Bigl[
    \gamma_k \cdot \max(0,\, \tilde{t} - \tilde{b}_k)
  + \delta_k \cdot \mathbf{1}(\tilde{t} \geq \tilde{b}_k) \notag \\
&\quad\qquad
  + \gamma_k^{\ell} \cdot \max(0,\, \tilde{t} - \tilde{b}_k) \cdot \text{leaning}_{j}
  + \delta_k^{\ell} \cdot \mathbf{1}(\tilde{t} \geq \tilde{b}_k) \cdot \text{leaning}_{j}
\Bigr] \notag \\
&\quad + \lambda \cdot \overline{\text{EMI}}_{j,t-1} + u_j + \epsilon_{jt}.
\end{align}
This model is identical to Equation~\ref{eq:2} except that $\gamma_k^{\ell}$ and $\delta_k^{\ell}$ are leaning-specific slope-change and level-shift coefficients, respectively, estimated as deviations from the reference group (centre). A significant $\gamma_k^{\ell}$ indicates that the post-election trend in EMI diverges for a given ideological group relative to the centre; a significant $\delta_k^{\ell}$ indicates a differential immediate shift in mean EMI at the breakpoint. The lagged outcome $\lambda \cdot \overline{\text{EMI}}_{j,t-1}$ and random intercept $u_j \sim \mathcal{N}(0, \sigma^2_u)$ serve the same purposes as in Equations~\ref{eq:1} and~\ref{eq:2}. All three models were estimated by maximum likelihood (REML\,=\,False) using \texttt{statsmodels MixedLM}. Model fit was assessed via likelihood-ratio test against an intercept-only null, AIC/BIC, intraclass correlation coefficient $\text{ICC} = \sigma^2_u / (\sigma^2_u + \sigma^2_\varepsilon)$, and the Durbin--Watson statistic on model residuals to verify that the lagged term adequately addressed residual autocorrelation.

\textbf{Rolling-window analysis.}
Equations~\ref{eq:1}--\ref{eq:3} impose a piecewise-linear functional form on
the EMI trajectories. To verify that our conclusions do not depend on this
parametrisation, we additionally traced the temporal evolution of EMI with a
rolling-window analysis. We slid a 180-day window across each corpus in
14-day steps and, within each window, fitted a simplified linear mixed-effects
model regressing document-level EMI on political leaning, the actor's mean EMI
in the preceding quarter and, for Twitter, the log-transformed like count,
with a random actor intercept $u_j$ as in Equation \ref{eq:1}. Leaning was entered without a
global intercept, so that each leaning coefficient directly estimates that
group's covariate-adjusted mean EMI within the window. Windows were estimated
only if they contained at least 300 (Parliament) or 500 (Twitter) documents
from at least ten actors spanning at least two ideological groups; all window
models were fitted by maximum likelihood, with 95\% Wald confidence intervals.

\subsection{Topic modelling}
To enable the topic-controlled robustness analysis described below, we estimated a separate topic model for each corpus using BERTopic \cite{grootendorst-2022}. Documents were embedded with the multilingual sentence-transformer \texttt{paraphrase-multilingual-mpnet-base-v2}, which produces 768-dimensional representations and offers strong German-language coverage. The embedding space was then dimensionally reduced with UMAP and clustered with HDBSCAN under the BERTopic defaults; the resulting clusters were reduced to a target of $K = 30$ topics with a minimum cluster size of 50 documents, a setting chosen to balance topical resolution against the requirement that each retained topic be estimable as a fixed effect in the downstream regression. Topics were labelled by their top class-based TF-IDF (c-TF-IDF) terms, and each cluster was then manually inspected and assigned to one of a smaller set of substantive policy domains; this aggregation was used to verify that the retained topics span the main areas of political debate rather than fragmenting a single domain. Full term lists and domain assignments per topic are provided in Supplementary Note 4.

The two corpora differ in size by roughly two orders of magnitude and were fit accordingly. For the contemporary Bundestag corpus ($n \approx 59{,}000$ speeches), BERTopic was fit on the full corpus. For the Twitter corpus ($n \approx 4.5$~million tweets), BERTopic was fit on a stratified subsample of 500{,}000 tweets that preserves the joint distribution of ideological leaning and party and the fitted model was then used to transform the full corpus. In both corpora, each document received a hard topic assignment $T_d \in \{-1, 1, \ldots, K\}$, with $T_d = -1$ denoting the BERTopic noise cluster of documents that the model could not confidently assign to any coherent topic. Soft topic probabilities were not retained because the downstream fixed-effects design only requires the identity of each document's assigned topic, not its degree of membership.

\subsection{Topic-controlled robustness analysis}
To address the concern that observed temporal shifts in EMI could reflect changes in document-level topic composition rather than in epistemic style, we re-estimated each baseline regression with document-level topic fixed effects, using the topic assignments obtained as described above. The largest topic was chosen as the reference category. Each augmented specification adds the same $K - 1$ document-level topic dummies to the corresponding baseline:
\begin{align}
\label{eq:4}
\overline{\text{EMI}}_{djt}^{\text{(topic-controlled)}} &= \overline{\text{EMI}}_{djt}^{\text{(baseline)}} + \sum_{m=1}^{K-1} \pi_m \cdot \mathbf{1}(T_d = m),
\end{align}
where $\pi_m$ is the mean EMI shift for documents in topic $m$ relative to the reference. The augmentation is applied to the linear baseline (Equation~\ref{eq:1}) for the Twitter corpus, and to both the constrained and the leaning-interactive breakpoint baselines (Equations~\ref{eq:2} and~\ref{eq:3}) for the Bundestag corpus, with the dispatch matching the baseline that was originally fit to each dataset. Because the FE design absorbs all between-topic variation, the time, leaning and breakpoint coefficients in the augmented models are estimated from \emph{within-topic} variation only, so any shift in the rhetorical signal that is driven by changes in topic mix over time can no longer drive them. Side-by-side comparisons of the baseline and topic-controlled coefficients for each specification are reported in Supplementary Note 4 and confirm that the time-trend, breakpoint, and leaning estimates are robust to topic-mix controls.

\clearpage
\bmhead{Data availability}
The validated German evidence and intuition dictionaries, the contemporary Bundestag speech dataset, the tweet IDs of the Twitter corpus (in line with the X Developer Agreement), and all aggregated time-series used in the figures and regression analyses are deposited on the Open Science Framework (OSF) at \url{https://osf.io/x3zpc/} (DOI to be assigned upon acceptance). The original Lasser tweet IDs are available from Lasser and colleagues \cite{lasser-2023}. The historical Bundestag corpus is available from Abrami and colleagues \cite{abrami-etal-2024-german} and from the Open Discourse project \cite{opendiscourse2020}. The contemporary Bundestag XML protocols are publicly available from the Bundestag document server \cite{bundestag_plenarprotokolle}. Access to raw tweet content is restricted by the X Developer Agreement and by the conditions of the DSA Article~40  framework. Researchers vetted to access this dataset under the DSA can have independent access to it, queries about access after approval should be directed to the corresponding author.

\bmhead{Code availability}
All code used to construct the corpora, train the embedding models, compute EMI scores, run the validation surveys and estimate the regression models is released under the MIT License on GitHub at \url{https://github.com/peersal/German-EMI}. Analyses were performed using Python~3.9.1.

\bmhead{Ethics declarations}
The two online survey studies were approved by the Ethics Committee of the University of Konstanz (approval no.\ 03/2023). All participants were recruited via Prolific, were informed of the study's purpose, and provided informed digital consent prior to participation. Participants were compensated at the Prolific median rate. Tweet content was processed under the conditions of the EU DSA Article~40; only anonymised, aggregated metrics are reported and no personally identifiable information (PII) beyond publicly visible verified-account names of elected politicians is included in any derived dataset.

\bmhead{Competing interests}
The authors declare no competing interests.

\bmhead{Author contributions}
P.S., S.T.A., S.L.\ and D.G.\ designed the study. P.S.\ collected and curated the corpora, implemented the embedding pipeline and the regression analyses, and ran the document-level validation survey. C.M.A.\ designed and ran the EMI keyword validation survey, and F.C.\ conducted the statistical validation analysis of the keyword survey. S.T.A.\ contributed the DDR pipeline and provided the historical-corpus modelling framework. S.L.\ provided theoretical input on the evidence/intuition framework. D.G.\ supervised the project. P.S.\ wrote the first draft of the manuscript; all authors contributed to revisions and approved the final version.

\bmhead{Funding}
Funded by the Deutsche Forschungsgemeinschaft (DFG -- German Research Foundation) under Germany's Excellence Strategy -- EXC-2035/2 -- 390681379. S.L., D.G., S.T.A and C.M.A.\ acknowledge financial support from the European Research Council (ERC Advanced Grant 101020961 PRODEMINFO).

\bmhead{Supplementary information}
A Supplementary Information document accompanies this submission, covering corpus construction (Note 1), EMI score validation through embedding-model comparisons, dictionary analyses, bootstrapping, two annotator surveys, and a historical validation (Note 2), Regressions (Note 3), and BERTopic diagnostics controlling for agenda-setting (Note 4).

\newpage
\begin{appendices}

\section{Extended Data}\label{secA1}

\begin{table}[h!]
\caption{Validated German keywords for the intuition and evidence dictionaries.}\label{tab:keywords}%
\begin{tabular}{@{}ll@{}}
\toprule
\textbf{Intuition Keywords} & \textbf{Evidence Keywords} \\
\midrule
Ansicht & Akkurat \\
Bauchgefühl & Analyse \\
Behauptung & Analysieren \\
Ehrlichkeit & Auskunft \\
Fake News & Bildung \\
Gefühl & Daten \\
Instinkt & Dossier \\
Intuition & Ergebnisse \\
Meinung & Evidenz \\
Misstrauen & Experte \\
Perspektive & Fakt \\
Propaganda & Forschung \\
Rat & Grund \\
Ratschlag & Hinweis \\
Sichtweise & Information \\
Standpunkt & Intelligenz \\
Täuschung & Korrektur \\
Unehrlichkeit & Labor \\
Verdacht & Logik \\
Zweifel & Methode \\
aufrichtig & Nachweis \\
behaupten & Prozess \\
beteuern & Prüfung \\
ehrlich & Punkt \\
gesunder Menschenverstand & Statistik \\
glauben & Studie \\
irreführen & Untersuchung \\
lügen & Verfahren \\
offensichtlich & Wahrheit \\
relativ & Wissen \\
täuschen & Wissenschaft \\
unehrlich & echt \\
unterstützen & erforschen \\
vermuten & ermitteln \\
vertrauen & exakt \\
vertrauenswürdig & gefälscht \\
verwechselt & genau bestimmen \\
Überzeugung & inkorrekt \\
 & korrekt \\
 & lernen \\
 & lesen \\
 & logisch \\
 & präzise \\
 & suchen \\
 & untersuchen \\
 & wahr \\
 & wahrheitsgemäß \\
 & wissenschaftlich \\
\botrule
\end{tabular}
\end{table}

\bigskip
\clearpage
\newpage
\begin{table}[h!]
    \caption{Illustrative intuition-based passages drawn from the German
    Twitter and Bundestag speech corpora (2015--2025), shown in the
    original German with English translation. Passages are sentences from
    the documents with the highest z-standardised intuition scores in
    each corpus.}
    \label{tab:intuition_passages}%
    \begin{tabular}{@{}p{5.5cm}p{5.5cm}@{}}
        \toprule
        \textbf{Original (German)} & \textbf{English Translation} \\
        \midrule
        \multicolumn{2}{@{}l}{\emph{Twitter}} \\
        \addlinespace
        ``{`}Meinung{'} haben Sie schon\dots\ Aber keine Sorge: {`}Meinung{'} aber wenig {`}Wissen{'} ist gerade Zeitgeist\dots'' & ``You do have an {`}opinion{'}\dots\ But don't worry: {`}opinion{'} with little {`}knowledge{'} is the zeitgeist right now\dots'' \\
        \addlinespace
        ``Alle wissen: wenn man es nur oft genug behauptet, fangen die ersten an, es zu glauben. Dann ist es eine Meinung. Dann wird die Meinung h\"aufiger vertreten, unabh\"angig von den Fakten [\dots]'' & ``Everyone knows: if you just assert it often enough, the first people start to believe it. Then it becomes an opinion. Then the opinion is held more and more widely, regardless of the facts [\dots]'' \\
        \addlinespace
        ``Ich denke, dass er sehr genau wei{\ss} was er tut. Ich habe begr\"undeten Verdacht zur Annahme, dass er den Trottel nur spielt [\dots]'' & ``I think he knows very well what he is doing. I have well-founded suspicion to assume that he is merely playing the fool [\dots]'' \\
        \addlinespace
        \multicolumn{2}{@{}l}{\emph{Bundestag speeches}} \\
        \addlinespace
        ``Ich verstehe die Absicht, aber ich vermisse jeglichen Sinn in Ihren Ausf\"uhrungen. Ich finde es atemberaubend, mit welchen Girlanden und relativ abstrakten Verrenkungen Sie es geschafft haben, uns hier zu erkl\"aren, warum wir keinen Ausschluss der Einb\"urgerung bei Mehr- und Vielehe vornehmen sollten.'' & ``I understand the intention, but I fail to see any sense in your remarks. I find it breathtaking with what garlands and rather abstract contortions you have managed to explain to us why we should not exclude naturalisation in cases of polygamy.'' \\
        \addlinespace
        ``[\dots] frage ich mich schon: Was sind das f\"ur verwirrte Geister, die hier am Rednerpult stehen? Es sind undurchdachte Antr\"age, die Sie hier vorgelegt haben.'' & ``[\dots] I really do ask myself: what kind of confused minds are standing at the lectern here? These are ill-considered motions that you have put forward.'' \\
        \addlinespace
        ``Ich kenne Sie als jemanden, der eher juristisch oberfl\"achlich und nicht an Fakten orientiert argumentiert und arbeitet. Das lehne ich schon aus tiefster Grund\"uberzeugung ab.'' & ``I know you as someone who argues and works in a legally superficial manner rather than being oriented towards facts. I reject that out of my deepest conviction.'' \\
        \botrule
    \end{tabular}
\end{table}

\bigskip

\begin{table}[h!]
    \caption{Illustrative evidence-based passages drawn from the German
    Twitter and Bundestag speech corpora (2015--2025), shown in the
    original German with English translation. Passages are sentences from
    the documents with the highest z-standardised evidence scores in each
    corpus.}
    \label{tab:evidence_passages}%
    \begin{tabular}{@{}p{5.5cm}p{5.5cm}@{}}
        \toprule
        \textbf{Original (German)} & \textbf{English Translation} \\
        \midrule
        \multicolumn{2}{@{}l}{\emph{Twitter}} \\
        \addlinespace
        ``Sie nehmen also eine Studie, welche Ihnen genehm ist, ohne die Annahmen und Methoden zu pr\"ufen und f\"uhren das als vermeintlichen Beweis f\"ur eine L\"uge an? Ich kenne viele Studien dazu, und die meisten haben M\"angel.'' & ``So you take a study that suits you, without checking its assumptions and methods, and present it as supposed proof of a lie? I know many studies on this topic, and most of them have shortcomings.'' \\
        \addlinespace
        ``6000 Tote sind nicht erfunden sondern wissenschaftliche Absch\"atzung auf 100ten von Studien basierend.'' & ``6,000 deaths are not invented but a scientific estimate based on hundreds of studies.'' \\
        \addlinespace
        ``Jede wissenschaftliche Arbeit sollte in Bezug auf Studiendesign und Auswertung von mehr als einer Person begutachtet werden.'' & ``Every scientific study should be reviewed by more than one person with regard to study design and analysis.'' \\
        \addlinespace
        \multicolumn{2}{@{}l}{\emph{Bundestag speeches}} \\
        \addlinespace
        ``Sie berufen sich hier auf eine Studie von John Cook aus dem Jahr 2013. [\dots] Schauen wir uns diese Studie einmal an. Was wurde da gemacht? John Cook hat 11\,944 wissenschaftliche Arbeiten untersucht.'' & ``You are relying here on a study by John Cook from 2013. [\dots] Let us take a look at this study. What was done there? John Cook examined 11,944 scientific papers.'' \\
        \addlinespace
        ``Dabei ist der NIPT zun\"achst nur eine Suche und keine Diagnose. Eine Diagnose erfolgt erst durch eine invasive Fruchtwasseruntersuchung, die das Risiko auf Fehlgeburten erh\"oht. Eine Vielzahl der Tests liefert falsch-positive Ergebnisse.'' & ``Yet the NIPT is initially only a screening, not a diagnosis. A diagnosis is only established by an invasive amniocentesis, which increases the risk of miscarriage. A large number of the tests yield false-positive results.'' \\
        \addlinespace
        ``Die fehlenden Zitate und nicht belegten Formulierungen verdeutlichen doch ganz klar, dass es sich eben nicht um eine glaubhafte, wissenschaftlich fundierte Analyse handelt [\dots]'' & ``The missing citations and unsubstantiated statements make it quite clear that this is not a credible, scientifically sound analysis [\dots]'' \\
        \botrule
    \end{tabular}
\end{table}

\clearpage
\begin{figure}[h!]
\centering\includegraphics[width=\textwidth]{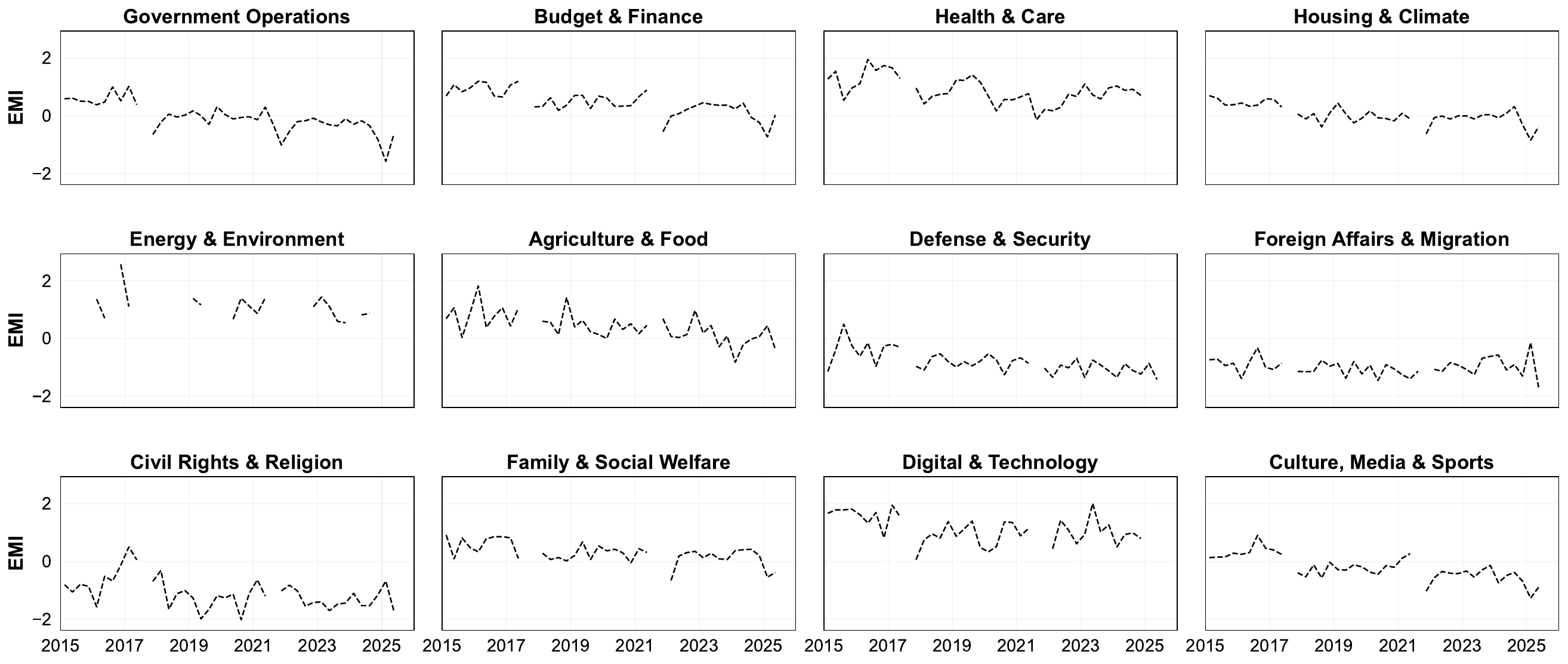}
\caption{\textbf{Evidence-minus-intuition (EMI) scores in the German
    Parliament by topic, 2015--2025.}
    Each panel shows the quarterly mean EMI of parliamentary speeches
    assigned to one semantic topic cluster (dashed lines). Topics were
    obtained with BERTopic ($K = 30$) and aggregated into twelve semantic
    clusters; documents in the BERTopic outlier cluster were excluded.
    Quarters with fewer than five speeches in a cluster are left blank.
    Axes are shared across panels.}
\label{fig:parliament_topic}
\end{figure}

\begin{figure}[h!]
\centering\includegraphics[width=\textwidth]{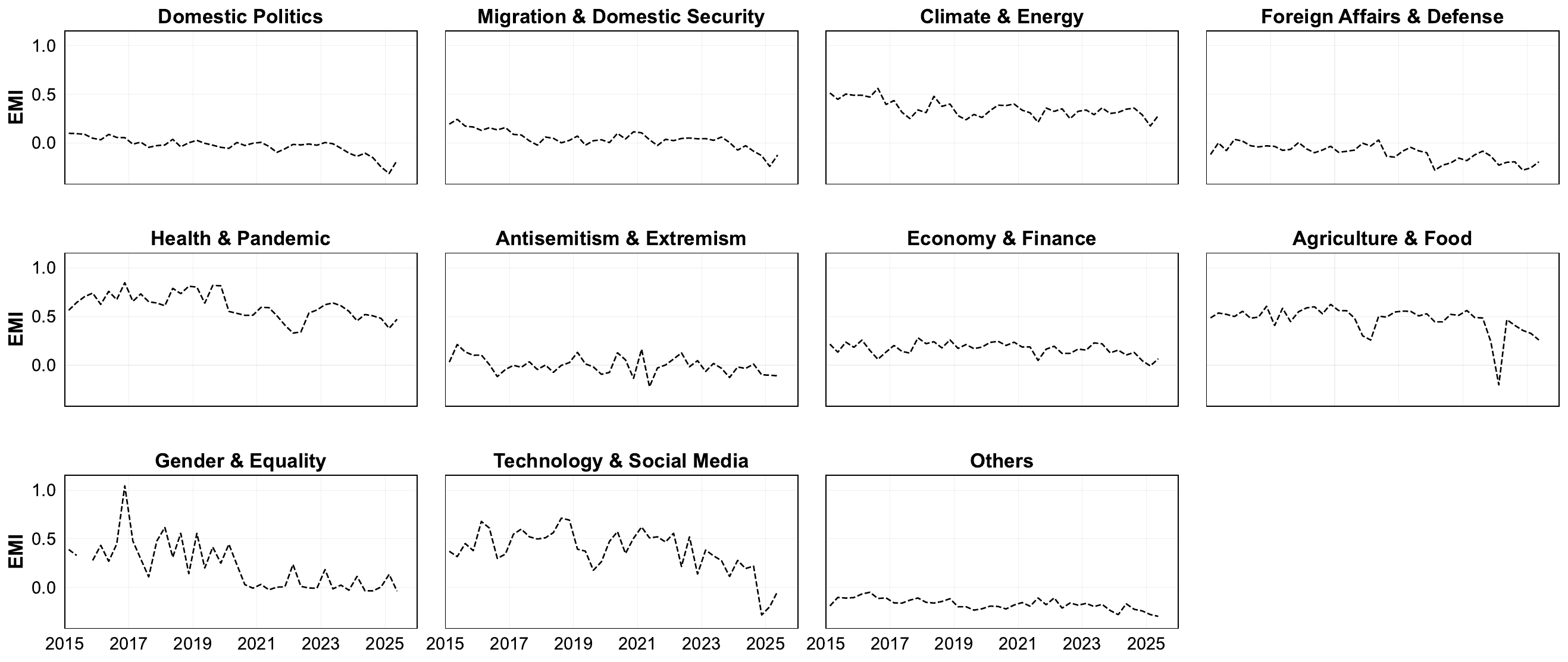}
\caption{\textbf{Evidence-minus-intuition (EMI) scores on Twitter by
    topic, 2015--2025.}
    Each panel shows the quarterly mean EMI of tweets assigned to one
    semantic topic cluster (dashed lines). Topics were obtained with
    BERTopic ($K = 30$) and aggregated into eleven semantic clusters; tweets in the BERTopic outlier cluster were excluded, and topics not assigned to a named cluster are collected under Others. Quarters with fewer than five tweets in a cluster are left blank. Axes are shared
    across panels.}
\label{fig:twitter_topic}
\end{figure}

\begin{table}[h!]
\caption{Mean Evidence Minus Intuition (EMI) scores on Twitter by party and ideological leaning, 2015--2025. Values report the unweighted mean across all tweets by all accounts affiliated with each party; standard deviations, standard errors and 95\% confidence intervals are derived from the document-level distribution. Rows in italics give the aggregate across all parties of the corresponding leaning, following the Chapel Hill Expert Survey (CHES) classification. The left and right leanings each comprise a single party, so their aggregates equal the respective party values.}\label{tab:emi_twitter}%
\begin{tabular}{@{}lccccc@{}}
\toprule
 & Mean EMI & Std & SE & 95\% CI & Sample size \\
\midrule
The Left Party & 0.049 & 0.734 & 0.001 & (0.047, 0.051) & 425246 \\
\textit{All left-leaning} & \textit{0.049} & \textit{0.734} & \textit{0.001} & \textit{(0.047, 0.051)} & \textit{425246} \\
\midrule
Alliance 90/The Greens & 0.135 & 0.749 & 0.001 & (0.133, 0.136) & 727984 \\
FDP & 0.011 & 0.751 & 0.001 & (0.009, 0.012) & 914924 \\
SPD & -0.019 & 0.763 & 0.001 & (-0.020, -0.017) & 608149 \\
CDU/CSU & -0.041 & 0.752 & 0.001 & (-0.042, -0.039) & 590891 \\
\textit{All centre} & \textit{0.026} & \textit{0.756} & \textit{0.000} & \textit{(0.025, 0.026)} & \textit{2841948} \\
\midrule
AfD & -0.268 & 0.688 & 0.001 & (-0.270, -0.266) & 508933 \\
\textit{All right-leaning} & \textit{-0.268} & \textit{0.688} & \textit{0.001} & \textit{(-0.270, -0.266)} & \textit{508933} \\
\botrule
\end{tabular}
\end{table}

\begin{table}[h!]
\caption{Mean Evidence Minus Intuition (EMI) scores in the German Bundestag by party and ideological leaning, 18th--21st electoral terms (2015--2025). Values report the unweighted mean across all speeches by all members affiliated with each party; standard deviations, standard errors and 95\% confidence intervals are derived from the document-level distribution. Rows in italics give the aggregate across all parties of the corresponding leaning, following the Chapel Hill Expert Survey (CHES) classification. The BSW estimate is based on a small sample ($n = 153$) and is reported descriptively only; the right leaning comprises a single party, so its aggregate equals the party values.}\label{tab:emi_speeches}%
\begin{tabular}{@{}lccccc@{}}
\toprule
 & Mean EMI & Std & SE & 95\% CI & Sample size \\
\midrule
The Left Party & 0.051 & 1.268 & 0.015 & (0.021, 0.080) & 7038 \\
BSW & -1.098 & 1.346 & 0.109 & (-1.311, -0.885) & 153 \\
\textit{All left-leaning} & \textit{0.026} & \textit{1.281} & \textit{0.015} & \textit{(-0.003, 0.056)} & \textit{7191} \\
\midrule
SPD & 0.113 & 1.264 & 0.011 & (0.092, 0.135) & 13262 \\
CDU/CSU & 0.108 & 1.258 & 0.010 & (0.089, 0.127) & 16690 \\
Alliance 90/The Greens & 0.039 & 1.229 & 0.013 & (0.014, 0.065) & 9165 \\
FDP & -0.110 & 1.187 & 0.015 & (-0.140, -0.081) & 6146 \\
\textit{All centre} & \textit{0.066} & \textit{1.247} & \textit{0.006} & \textit{(0.054, 0.077)} & \textit{45263} \\
\midrule
AfD & -0.473 & 1.236 & 0.015 & (-0.502, -0.443) & 6716 \\
\textit{All right-leaning} & \textit{-0.473} & \textit{1.236} & \textit{0.015} & \textit{(-0.502, -0.443)} & \textit{6716} \\
\botrule
\end{tabular}
\end{table}

\bigskip
\begin{table}
\caption{Linear mixed-effects regression of Evidence Minus Intuition (EMI) on time, ideological leaning and their interaction for the Twitter corpus (Equation~\ref{eq:1}). Random intercepts at the actor level. Reference category for leaning is centre.}
\label{tab:model1}%
\centering
\begin{tabular}{@{}llll@{}}
\toprule
Model:            & MixedLM & Dependent Variable: & emi\_w2vparliament  \\
No. Observations: & 3707690 & Method:             & ML                  \\
No. Groups:       & 1063    & Scale:              & 0.5062              \\
Min. group size:  & 1       & Log-Likelihood:     & -4000976.3513       \\
Max. group size:  & 80191   & Converged:          & Yes                 \\
Mean group size:  & 3487.9  &                     &                     \\
\botrule
\end{tabular}

\medskip
\begin{tabular}{@{}lrrrrrr@{}}
\toprule
                                &  Coef. & Std.Err. &        z & P$> |$z$|$ & [0.025 & 0.975]  \\
\midrule
Intercept                       &  0.050 &    0.007 &    6.929 &       0.000 &  0.036 &  0.064  \\
leaning[T.right]                & -0.111 &    0.020 &   -5.620 &       0.000 & -0.149 & -0.072  \\
leaning[T.left]                 &  0.001 &    0.020 &    0.069 &       0.945 & -0.039 &  0.041  \\
time\_centered                  & -0.001 &    0.000 &  -29.184 &       0.000 & -0.001 & -0.001  \\
time\_centered:leaning[T.right] &  0.001 &    0.000 &   20.468 &       0.000 &  0.001 &  0.001  \\
time\_centered:leaning[T.left]  & -0.000 &    0.000 &   -0.684 &       0.494 & -0.000 &  0.000  \\
log\_like\_count                & -0.040 &    0.000 & -149.524 &       0.000 & -0.040 & -0.039  \\
emi\_lag1                       &  0.495 &    0.002 &  202.166 &       0.000 &  0.490 &  0.500  \\
Group Var                       &  0.038 &          &          &             &        &         \\
\botrule
\end{tabular}
\end{table}

\bigskip
\begin{table}
\caption{Constrained breakpoint linear mixed-effects regression of Evidence Minus Intuition (EMI) on time, ideological leaning and election-year structural breaks (2017, 2021, 2025) for the Bundestag corpus (Equation~\ref{eq:2}). Breakpoint level shifts ($\delta_k$) and slope changes ($\gamma_k$) are constrained to be uniform across ideological camps. Random intercepts at the actor level. Reference category for leaning is centre.}
\label{tab:model2}%
\centering
\begin{tabular}{@{}llll@{}}
\toprule
Model:            & MixedLM & Dependent Variable: & emi\_w2vparliament  \\
No. Observations: & 56066   & Method:             & ML                  \\
No. Groups:       & 1218    & Scale:              & 1.0245              \\
Min. group size:  & 1       & Log-Likelihood:     & -81804.4067         \\
Max. group size:  & 362     & Converged:          & Yes                 \\
Mean group size:  & 46.0    &                     &                     \\
\botrule
\end{tabular}

\medskip
\begin{tabular}{@{}lrrrrrr@{}}
\toprule
                                &  Coef. & Std.Err. &       z & P$> |$z$|$ & [0.025 & 0.975]  \\
\midrule
Intercept                       &  0.623 &    0.066 &   9.485 &       0.000 &  0.494 &  0.752  \\
leaning[T.left]                 & -0.075 &    0.067 &  -1.115 &       0.265 & -0.207 &  0.057  \\
leaning[T.right]                & -0.342 &    0.065 &  -5.260 &       0.000 & -0.469 & -0.214  \\
time\_centered                  &  0.003 &    0.001 &   2.197 &       0.028 &  0.000 &  0.006  \\
time\_centered:leaning[T.left]  &  0.002 &    0.001 &   3.080 &       0.002 &  0.001 &  0.003  \\
time\_centered:leaning[T.right] &  0.001 &    0.001 &   1.217 &       0.224 & -0.001 &  0.002  \\
slope\_elec2017                 & -0.001 &    0.002 &  -0.638 &       0.524 & -0.004 &  0.002  \\
level\_elec2017                 & -0.421 &    0.027 & -15.787 &       0.000 & -0.473 & -0.368  \\
slope\_elec2021                 & -0.008 &    0.001 &  -7.287 &       0.000 & -0.010 & -0.006  \\
level\_elec2021                 & -0.207 &    0.023 &  -9.180 &       0.000 & -0.251 & -0.162  \\
slope\_elec2025                 &  0.303 &    0.062 &   4.883 &       0.000 &  0.182 &  0.425  \\
level\_elec2025                 & -0.502 &    0.050 & -10.061 &       0.000 & -0.600 & -0.404  \\
emi\_lag1                       &  0.125 &    0.007 &  19.036 &       0.000 &  0.112 &  0.138  \\
Group Var                       &  0.390 &    0.019 &         &             &        &         \\
\botrule
\end{tabular}
\end{table}

\bigskip
\begin{table}
\caption{Full breakpoint linear mixed-effects regression of Evidence Minus Intuition (EMI) on time, ideological leaning and election-year structural breaks (2017, 2021, 2025) with leaning-specific breakpoint interactions for the Bundestag corpus (Equation~\ref{eq:3}). Random intercepts at the actor level. Reference category for leaning is centre. Post-2025 estimates are based on a sparse post-breakpoint sample and should be interpreted with caution.}
\label{tab:model3}%
\centering
\begin{tabular}{@{}llll@{}}
\toprule
Model:            & MixedLM & Dependent Variable: & emi\_w2vparliament  \\
No. Observations: & 56066   & Method:             & ML                  \\
No. Groups:       & 1218    & Scale:              & 1.0238              \\
Min. group size:  & 1       & Log-Likelihood:     & -81786.6693         \\
Max. group size:  & 362     & Converged:          & Yes                 \\
Mean group size:  & 46.0    &                     &                     \\
\botrule
\end{tabular}

\medskip
\begin{tabular}{@{}lrrrrrr@{}}
\toprule
                                &  Coef. & Std.Err. &       z & P$> |$z$|$ & [0.025 & 0.975]  \\
\midrule
Intercept                       &  0.607 &    0.072 &   8.428 &       0.000 &  0.466 &  0.748  \\
leaning[T.left]                 &  0.015 &    0.172 &   0.088 &       0.930 & -0.323 &  0.353  \\
leaning[T.right]                & -0.347 &    0.068 &  -5.071 &       0.000 & -0.480 & -0.213  \\
time\_centered                  &  0.003 &    0.002 &   1.652 &       0.099 & -0.000 &  0.006  \\
time\_centered:leaning[T.left]  &  0.005 &    0.004 &   1.341 &       0.180 & -0.002 &  0.012  \\
time\_centered:leaning[T.right] & -0.000 &    0.002 &  -0.087 &       0.931 & -0.004 &  0.004  \\
slope\_elec2017                 & -0.000 &    0.002 &  -0.279 &       0.780 & -0.004 &  0.003  \\
level\_elec2017                 & -0.427 &    0.030 & -14.429 &       0.000 & -0.485 & -0.369  \\
slope\_elec2017\_left           & -0.002 &    0.004 &  -0.404 &       0.686 & -0.010 &  0.006  \\
level\_elec2017\_left           &  0.018 &    0.073 &   0.252 &       0.801 & -0.125 &  0.162  \\
slope\_elec2021                 & -0.007 &    0.001 &  -5.616 &       0.000 & -0.009 & -0.005  \\
level\_elec2021                 & -0.218 &    0.026 &  -8.387 &       0.000 & -0.269 & -0.167  \\
slope\_elec2021\_left           & -0.010 &    0.004 &  -2.741 &       0.006 & -0.017 & -0.003  \\
level\_elec2021\_left           &  0.088 &    0.075 &   1.174 &       0.240 & -0.059 &  0.234  \\
slope\_elec2021\_right          &  0.000 &    0.003 &   0.050 &       0.960 & -0.006 &  0.006  \\
level\_elec2021\_right          &  0.030 &    0.064 &   0.472 &       0.637 & -0.096 &  0.156  \\
slope\_elec2025                 &  0.277 &    0.071 &   3.879 &       0.000 &  0.137 &  0.416  \\
level\_elec2025                 & -0.490 &    0.056 &  -8.754 &       0.000 & -0.600 & -0.381  \\
slope\_elec2025\_left           &  0.757 &    0.228 &   3.315 &       0.001 &  0.309 &  1.205  \\
level\_elec2025\_left           & -0.493 &    0.176 &  -2.802 &       0.005 & -0.838 & -0.148  \\
slope\_elec2025\_right          & -0.301 &    0.178 &  -1.690 &       0.091 & -0.651 &  0.048  \\
level\_elec2025\_right          &  0.277 &    0.156 &   1.774 &       0.076 & -0.029 &  0.583  \\
emi\_lag1                       &  0.124 &    0.007 &  18.868 &       0.000 &  0.111 &  0.137  \\
Group Var                       &  0.391 &    0.019 &         &             &        &         \\
\botrule
\end{tabular}
\end{table}

\clearpage

\end{appendices}

\clearpage
\newpage

\bibliography{sources.bib}%

\end{document}


\begin{titlepage}
    \centering
    \vspace*{2cm}
    
    \Huge \textbf{Supplementary Materials}\\
    \vspace{0.5cm}
    \Large for the Paper:\\
    \vspace{0.5cm}
    \Large \textbf{German parties shifted towards intuition-based rhetoric after the far right's parliamentary breakthrough}\\
    
    \vspace{2cm}
    \Large Author: \textbf{Peer Saleth} \\
    \Large Organization: \textbf{University of Constance} \\
    \Large Date: \textbf{\today} \\

    \vfill
\end{titlepage}

\newpage

\tableofcontents

\newpage

\listoffigures

\newpage

\section{Data}
In this section we give a detailed description of the data generation process, preprocessing, anonymization steps and descriptive. All Data received by us is publicly available at the OSF repository. All Data used is publicly available at the corresponding link/source.

\subsection{German Parliament Data}
We leverage existing data from the German
Parliament. Therefore, we combine two publicly available historical datasets: GerParCor \cite{abrami-2024} (1867–1942) and Open Discourse \cite{unknown-author-no-dateF} (1949–2021).

We combined the two datasets into a unified corpus encompassing all speeches delivered in the German parliament between 1867 and 2022. The preprocessing workflow, adapted from the pipeline described by Aroyehun et al. \cite{aroyehun-2024A}, included steps such as removing special characters and stop words, with manual adjustments to tailor the process to the dataset’s specific requirements. Entries shorter than 10 tokens were excluded, while lengthy speeches were segmented into chunks containing 50 to 150 words. To maintain the integrity of the original content, the preprocessed chunks were matched with their corresponding original text, ensuring a reliable basis for validation in subsequent analyses.

\begin{figure}[H]
    \centering    \includegraphics[width=\textwidth]{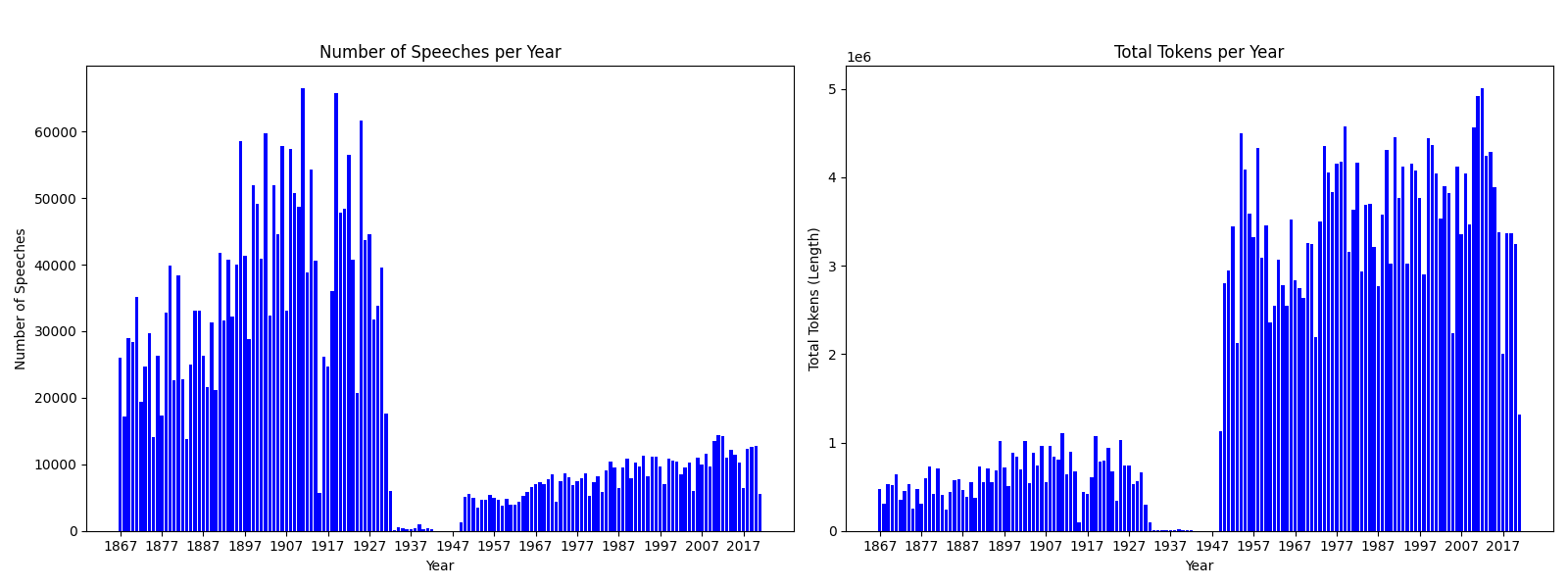}
    \caption{Speeches Descriptives}
    \label{fig:speeches1}
\end{figure}

\begin{figure}[H]
    \centering    
    \includegraphics[width=12cm]{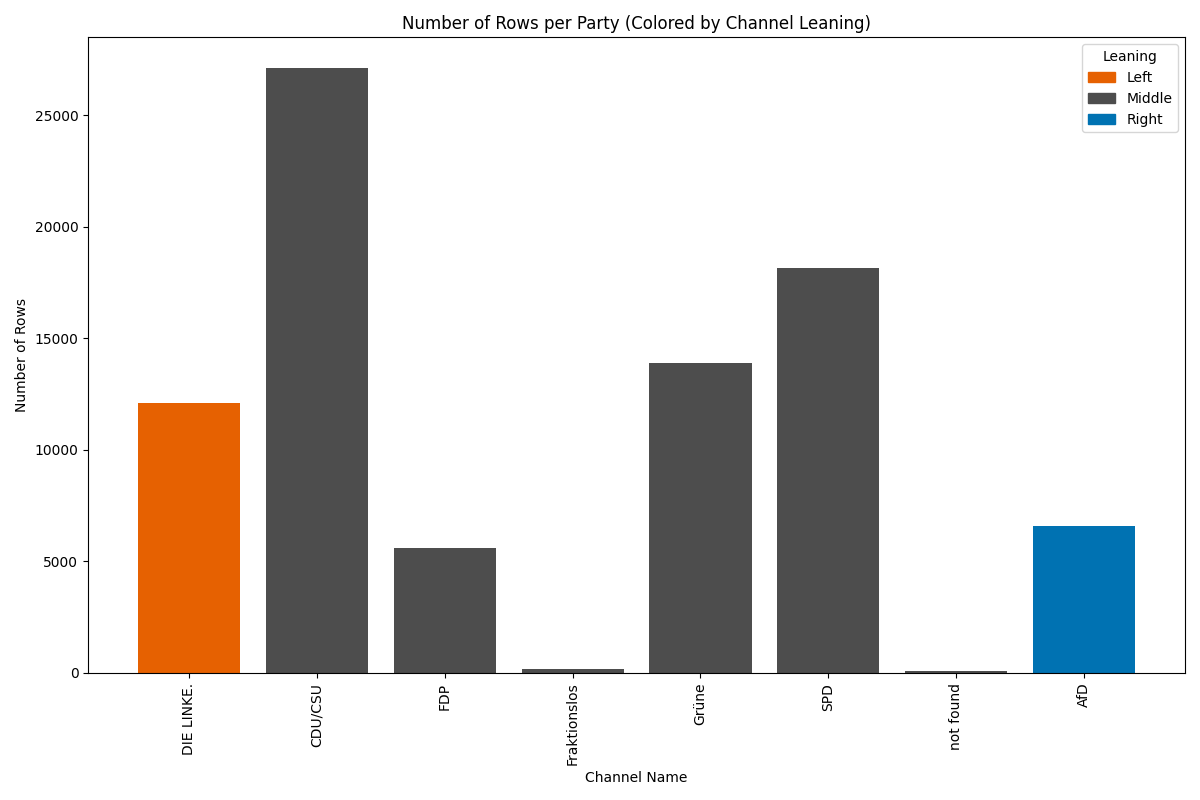}
    \caption{Party Distribution Parliament Speeches 2014 to 2022}
    \label{fig:speeches2}
\end{figure}

Descriptive visualizations (Figures \ref{fig:speeches1} and \ref{fig:speeches2}) shows the number and total length of speeches over Time. Binning confirms concentration in a specific range, with few outliers. Speech frequency fluctuates over time, peaking around major political events and sharply declining during periods of war. Parties were only analyzed for the time period we define for our analysis (2014 to 2022). Centrist parties (CDU/CSU, SPD, Greens) dominate the dataset, while DIE LINKE, AfD, and FDP contribute fewer entries. A small portion is labeled "non-bund" (uncategorized/independent).

\subsection{Twitter Data}
\begin{figure}[h!]
    \centering    \includegraphics[width=\textwidth]{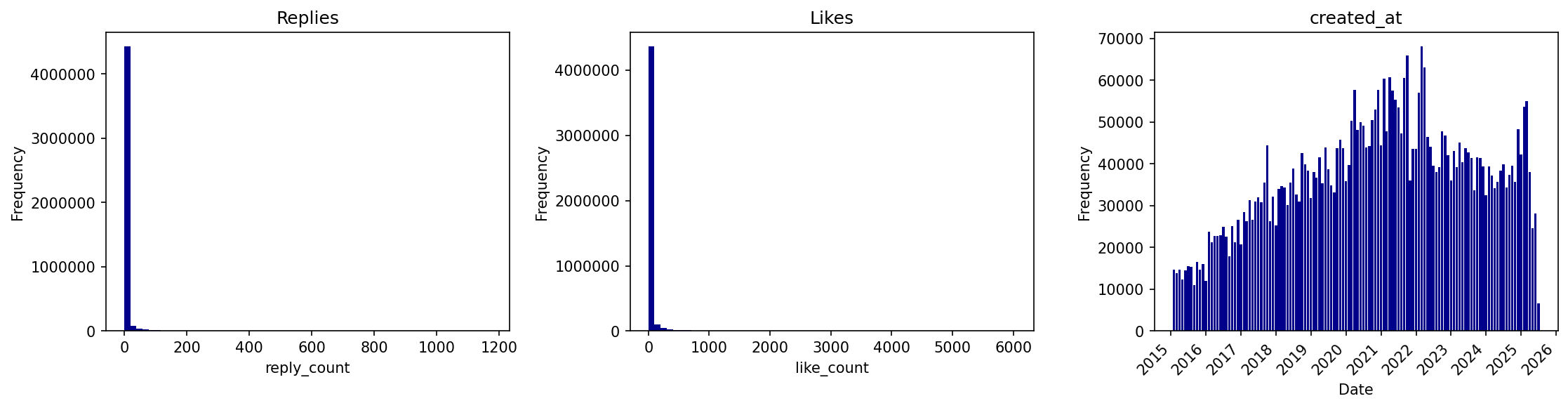}
    \caption{Descriptives Tweets}
    \label{fig:tweets2}
\end{figure}

\begin{figure}[h!]
    \centering    \includegraphics[width=\textwidth]{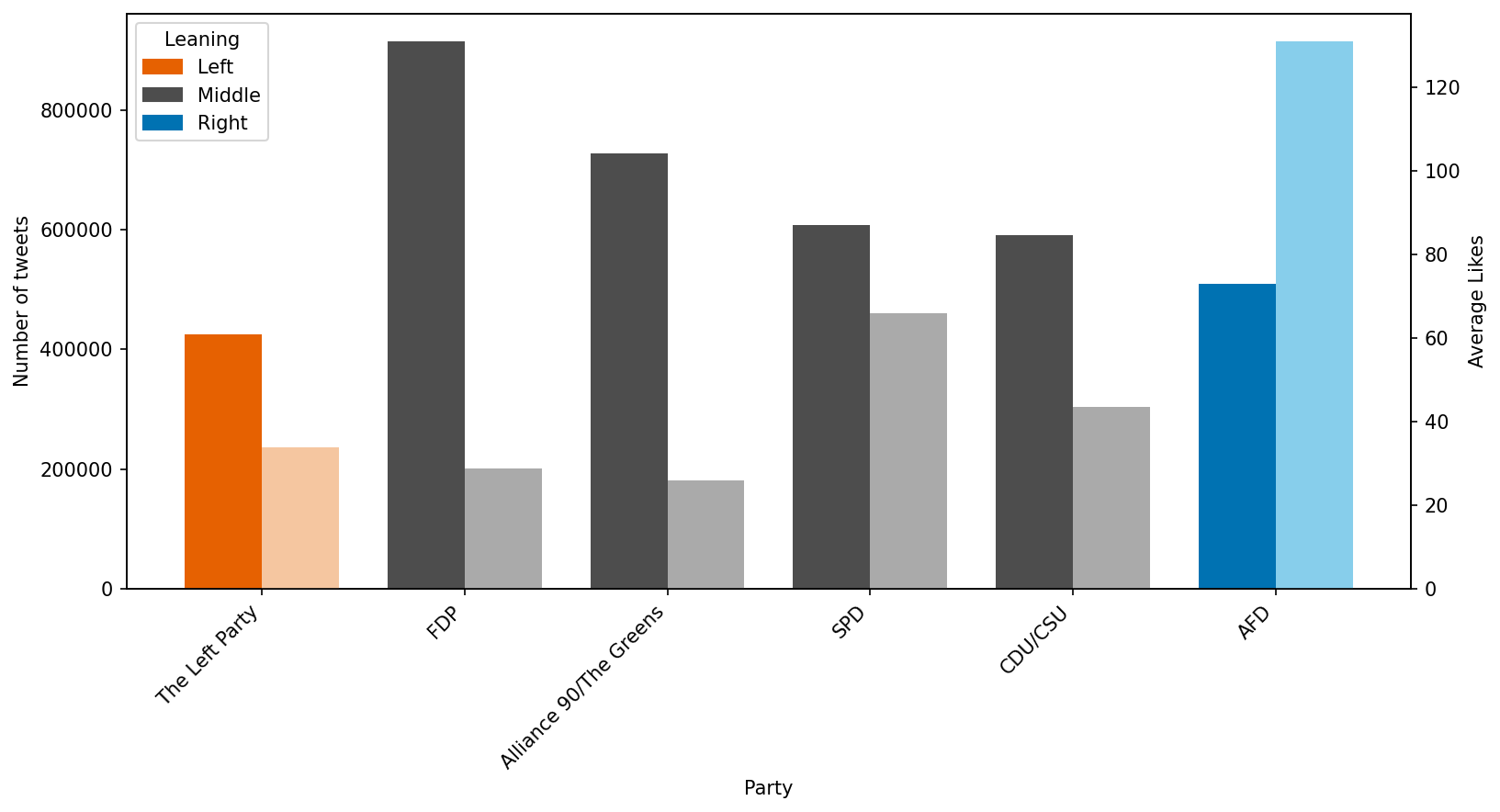}
    \caption{Party Distribution Twitter}
    \label{fig:tweets1}
\end{figure}

We use a German sample from Lasser et al. \cite{lasser-2023}, who collected tweets from verified politician accounts (2016–2022) using the Twitter API, excluding retweets and non-informative URLs. Figure \ref{fig:tweets1} shows tweet volume and engagement by party: left-leaning parties (orange) tweet frequently; centrist parties (gray) are more balanced; AfD (blue) shows moderate volume but highest average likes per tweet, indicating strong engagement. Figure \ref{fig:tweets2} illustrates engagement metrics—most tweets receive few replies and likes, though a small subset gains high interaction. Tweet frequency rises steadily from 2015 onward. Overall, the data highlight differing engagement levels and long-term trends in party communication on Twitter.

\newpage
\section{Evidence Minus Intuition Scores}

For ensuring robust findings, we validate the Evidence Minus Intuition (EMI) score through multiple approaches. What follows is an explanation and short interpretation of all validation steps (see the methods for details on how the scores are derived). Distributed representation methods, such as Word2Vec, GloVe, and transformer-based models like BERT or SBERT, are essential in text analysis, capturing semantic relationships beyond simple keyword matching. They enable flexible, context-aware classification but face challenges in interpretability and domain adaptation. 

\subsection{Embedding Model Variation}
\begin{landscape}  %
    \begin{figure}
        \centering
        \includegraphics[width=700pt]{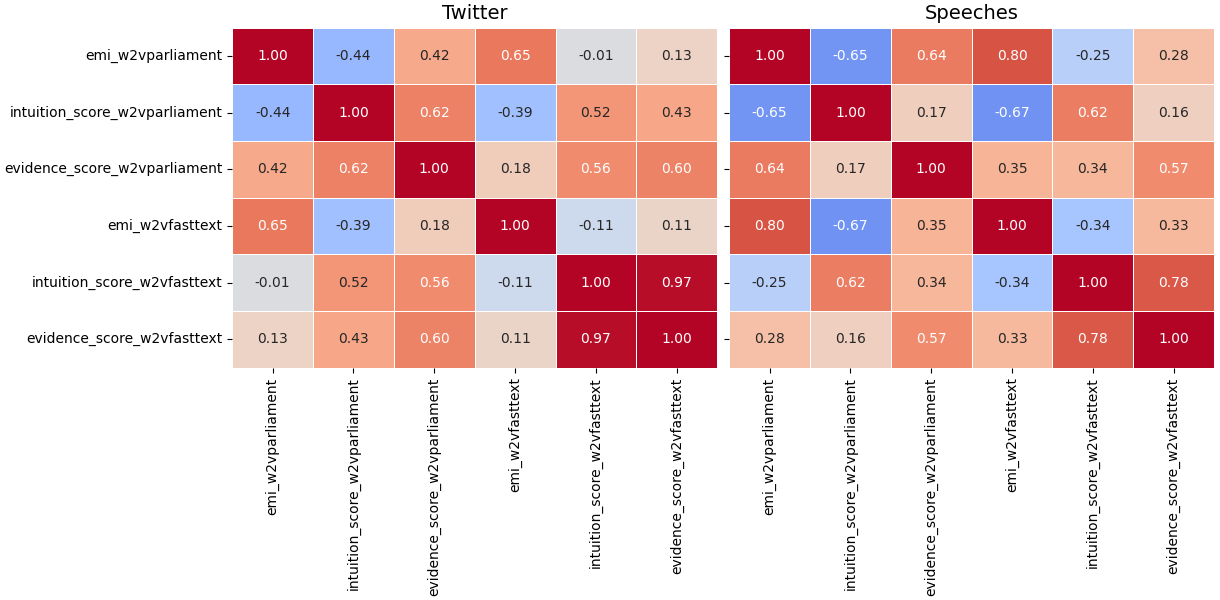}
        \caption{Model Correlation}
        \label{fig:model_correlation}
    \end{figure}
\end{landscape}

To ensure validity across embedding models, we computed EMI scores using two approaches, both implemented with the \texttt{gensim} library\footnote{\url{https://pypi.org/project/gensim/}}. All models and code are available in the GitHub repository. The first model, \textbf{w2v\_parliament}, was trained on 23 million words from German parliamentary speeches (1870–2023) over 10 epochs (window size 10, 100 dimensions). It was then fine-tuned separately on 10 million Twitter words. Fine-tuning continued training on new domains while preserving the original model’s representations, resulting in three models: a base model and two domain-specific variants. The second model, \textbf{wtv\_fasttext}, uses pre-trained FastText embeddings\footnote{\url{https://fasttext.cc/docs/en/crawl-vectors.html}} (CBOW with position weights, 300 dimensions, n-grams of length 5, window size 5, 10 negatives), trained on Common Crawl and Wikipedia. To enable average pooling and GPU acceleration, we converted the fine-tuned Word2Vec models into \texttt{Sentence-BERT}\footnote{\url{https://sbert.net/}} (SBERT) format. Word vectors were saved in plain-text and loaded using SBERT’s \texttt{WordEmbeddings} module, with whitespace tokenization and stopword filtering. A \texttt{Pooling} layer then applied mean pooling to generate sentence-level embeddings. Figure \ref{fig:model_correlation} shows Pearson correlations between EMI scores from both models. Correlations were strong across all domains: 0.80 for Speeches (95\% CI [0.80, 0.81]) and 0.65 for Twitter (95\% CI [0.65, 0.65]); all $p < 0.001$. Evidence and intuition scores were also highly correlated across models, both at the document and word level.
The p-values were calculated using Pearson's test, and confidence intervals were derived using Fisher's z-transformation and bootstrapping to account for variability.

\newpage

\newpage

\subsection{Dictionary Analysis}
Evidence-based and intuition-based reasoning represent distinct yet interconnected approaches to truth-seeking in discourse. Though often seen as oppositional, both can coexist and complement each other in complex discussions. However, when using distributed dictionary methods, overlaps between these constructs can emerge due to context-dependent word meanings and corpus-specific biases. Words appearing in both reasoning styles may develop dual associations, especially in corpora where intuition and evidence frequently co-occur. Since word embeddings reflect statistical co-occurrences rather than fixed categories, polysemous or conceptually linked terms (e.g., "belief," "experience") may drift semantically between categories.

To assess overlap, we computed cosine similarity between embeddings of words from the evidence and intuition dictionaries. Mean similarities between averaged dictionary embeddings were 0.3241 (speeches) and 0.4443 (Twitter). Higher values indicate greater semantic proximity, suggesting a blurrier boundary between rhetorical styles in informal contexts (Twitter) than in formal ones (parliamentary speeches), where the styles are more distinct.

Figures \ref{fig:dic1}–\ref{fig:dic3} show pairwise cosine similarity heatmaps between dictionary terms. Some intuition words (e.g., "Ehrlichkeit" – honesty) align closely with evidence words like "Wahrheit" (truth), reflecting contextual convergence. Others diverge strongly, such as "Labor" (lab) vs. "aufrichtig" (sincere), indicating domain-specific use. Embedding differences across models further reflect their training data: the fine-tuned Twitter model show higher semantic overlap due to the opinion-rich nature of social media, compared to the more structured language of parliamentary texts. 
\begin{landscape}  %
    \begin{figure}
        \centering
        \includegraphics[width=500pt]{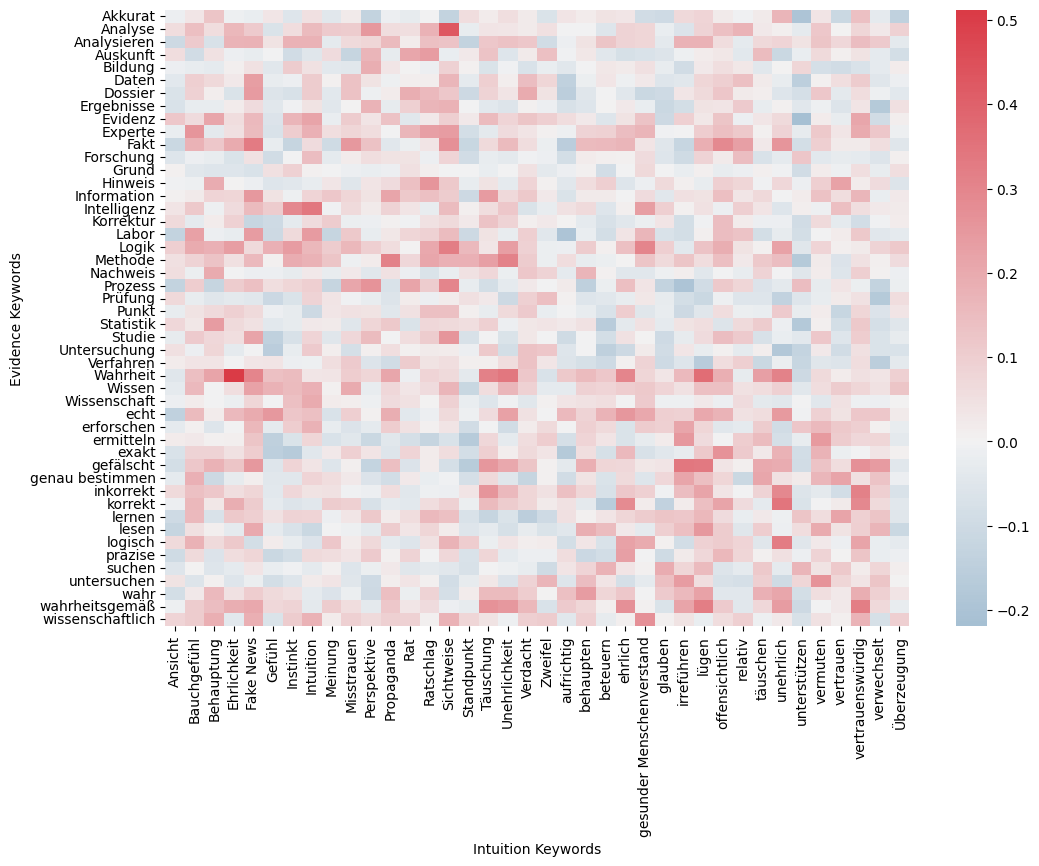}
        \caption{Dictionary Correlation Speeches Model}
        \label{fig:dic1}
    \end{figure}

    \begin{figure}
        \centering
        \includegraphics[width=500pt]{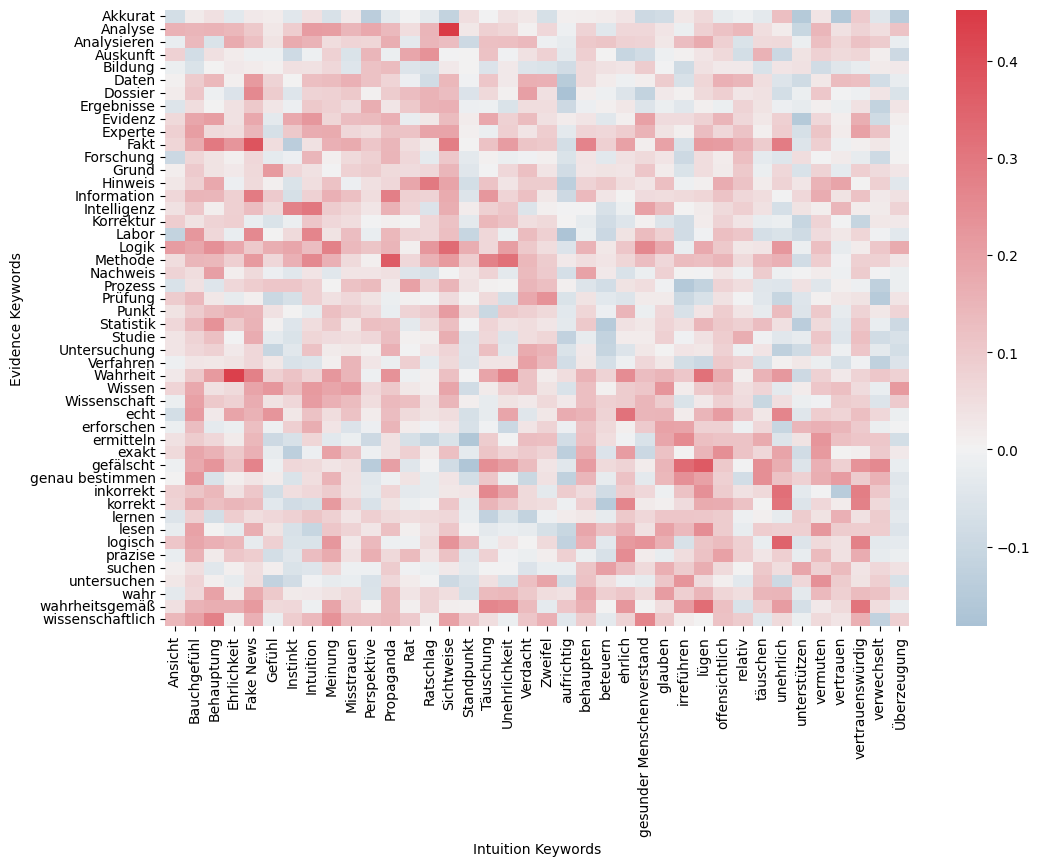}
        \caption{Dictionary Correlation Twitter Model}
        \label{fig:dic3}
    \end{figure}
\end{landscape}

\subsection{Descriptive and Length Correlation}
\begin{figure}[h!]
    \centering
    \includegraphics[width=\textwidth]{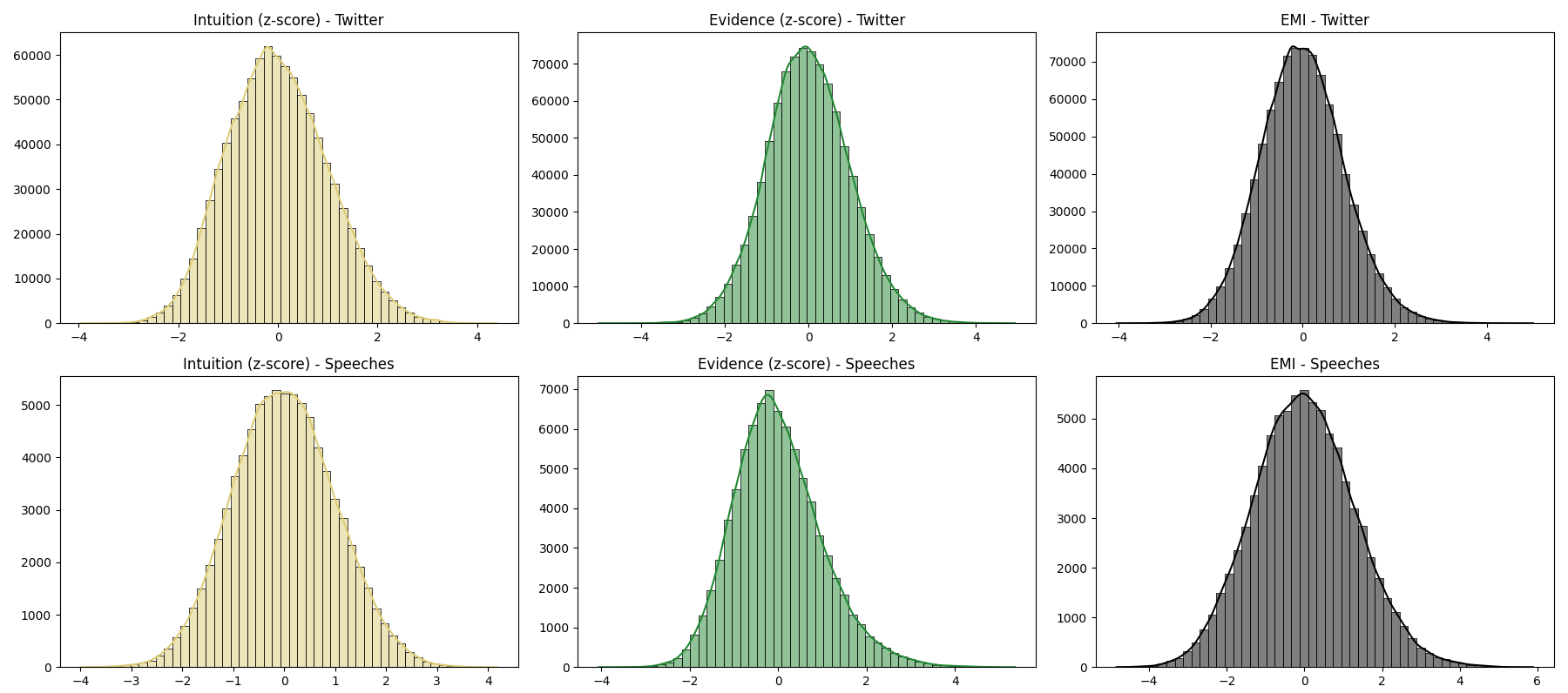}
    \caption{Truth Histograms}
    \label{fig:truthhist}
\end{figure}

\begin{figure}[h!]
    \centering
    \includegraphics[width=\textwidth]{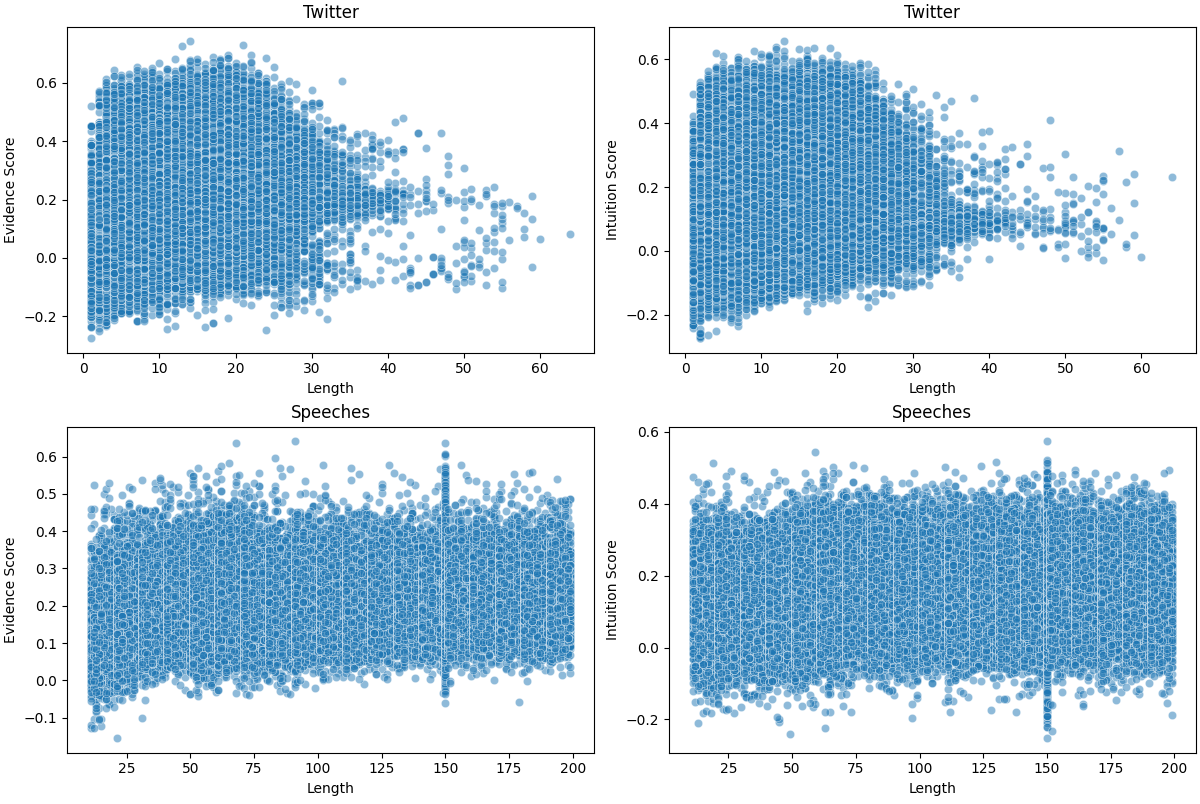}
    \caption{Text-length correlation}
    \label{fig:length}
\end{figure}

As recent research suggests there is a correlation between text length and embedded cosine similarities \cite{lasser-2023, lasser-2022, aroyehun-2024A} we test and control for length. Looking at the descriptive in Figure \ref{fig:truthhist}  reveals normal distributions for Twitter and Parliament Speeches. Looking into the scatterplot in Figure \ref{fig:length} reveals no distinct linear correlation between evidence similarities and text length. The same holds for intuition.  After adjusting for text length by binning and z-scoring, the correlation with the EMI score is effectively zero (0.000, $p = 0.999$, 95\% CI: [-0.009, 0.010]).

\begin{landscape}  %
    \begin{figure}
        \centering
        \includegraphics[width=700pt]{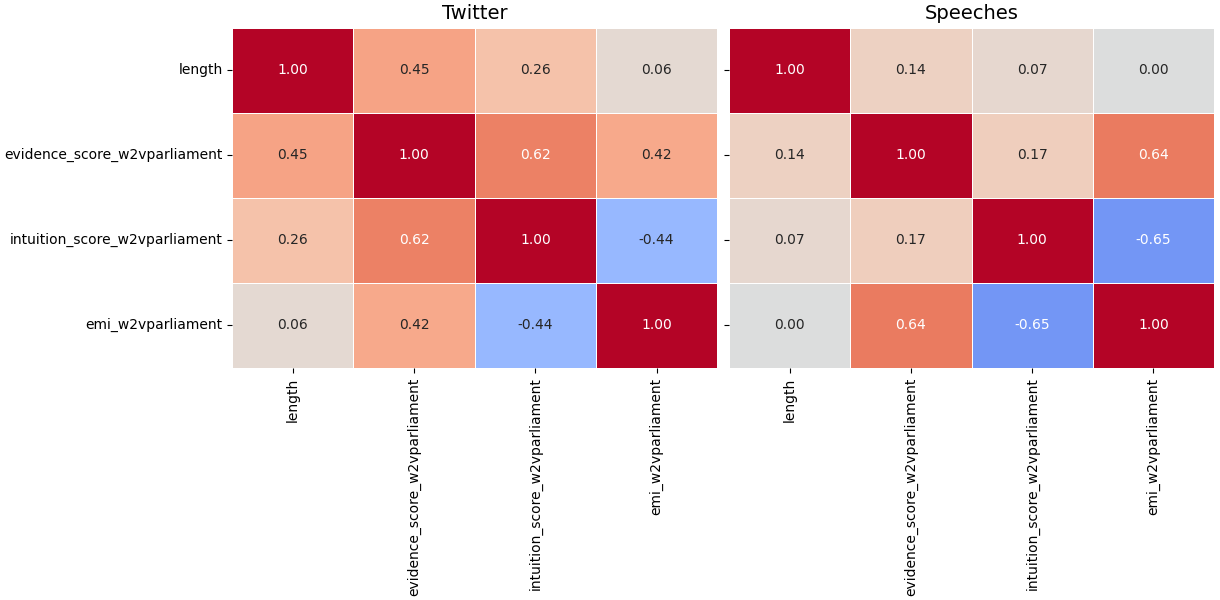}
        \caption{Text -length Correlationmatrix}
        \label{fig:lengt_corrmatrix}
    \end{figure}
\end{landscape}

\clearpage
\subsection{Corpus analysis}
\begin{figure}[h!]
    \centering
    \includegraphics[width=\textwidth]{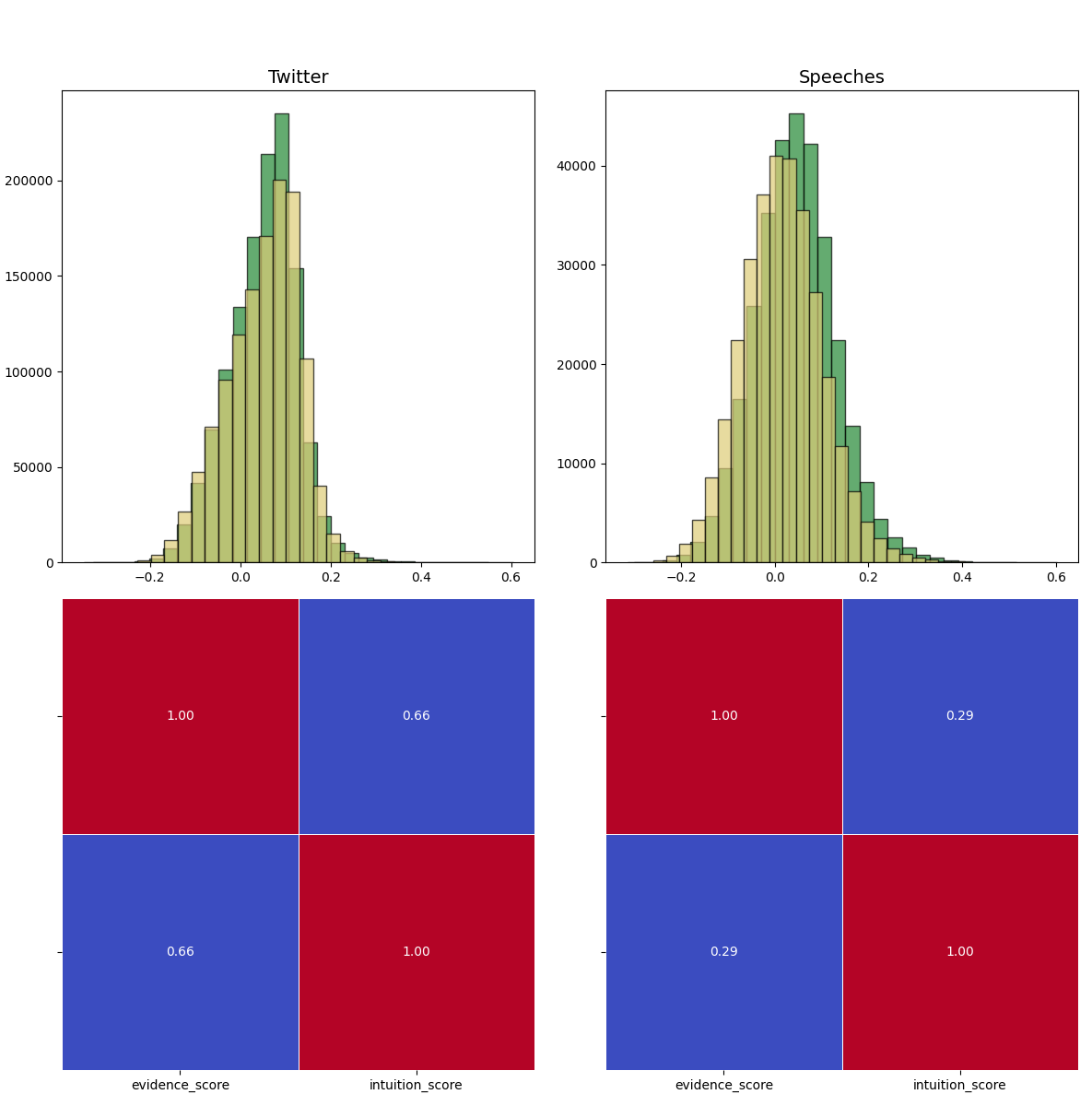}
    \caption{Word Similarity Distributions}
    \label{fig:word_hist}
\end{figure}

Next we analysed influential words in the corpus. This process was used iteratively to define fine-tuned stop words. We calculate frequency, intuition similarity and evidence similarity for each single word in the corpus. The histograms in Figure \ref{fig:word_hist} reveal that similarities are normally distributed at a word level. We find high correlations between evidence and intuition on the word level. 

\subsection{EMI Components}
\begin{figure}[h!]
\centering\includegraphics[width=\textwidth]{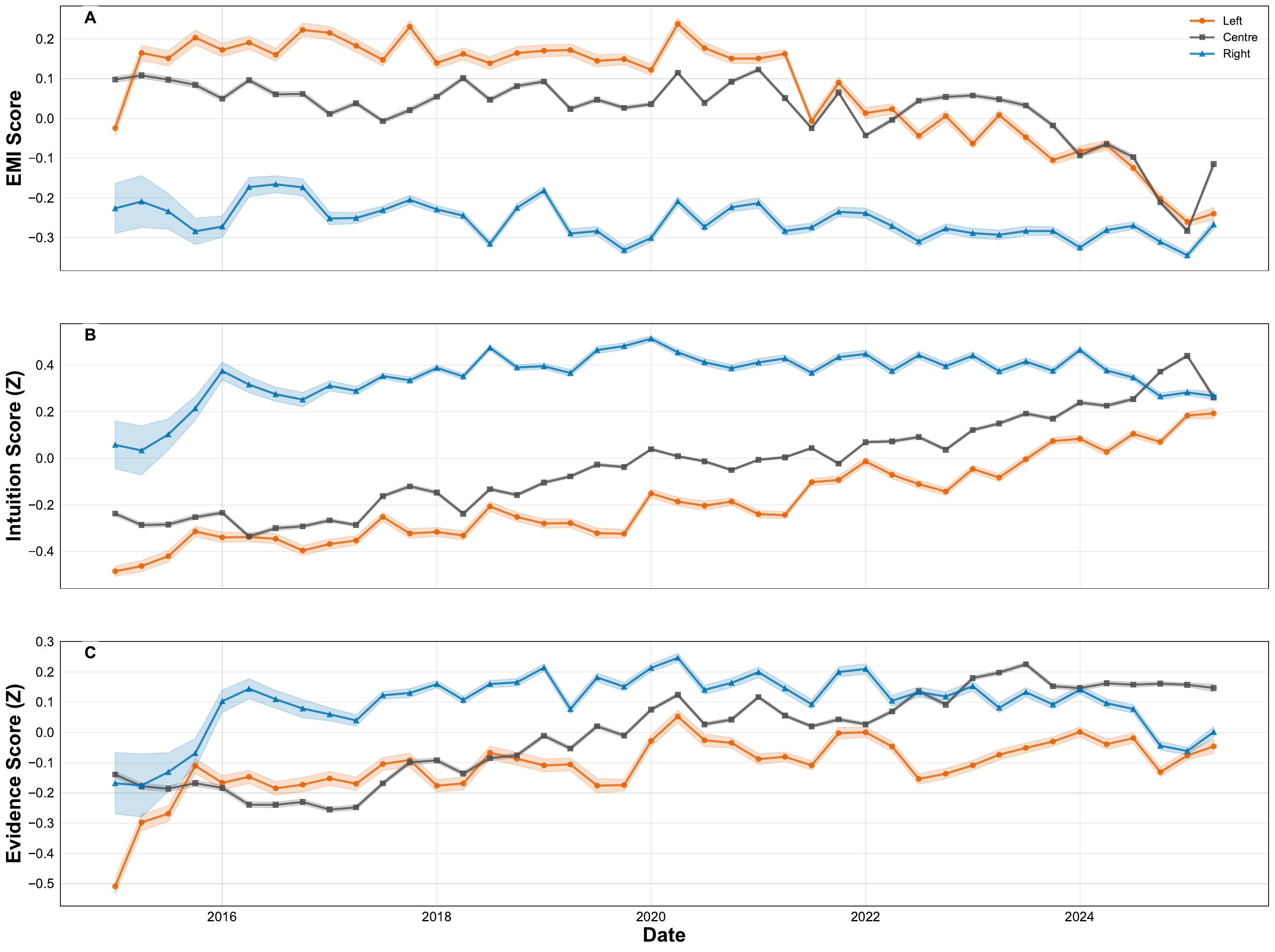}
\caption{
\textbf{Component-level rhetorical styles in tweets by political leaning, 2015--2025.}
\textbf{A}, Quarterly averages of the Evidence Minus Intuition (EMI) score, \textbf{B}, intuition-based language similarity, and \textbf{C}, evidence-based language similarity, with parties grouped by ideological leaning (left, centre, right). Shaded regions represent 95\% confidence intervals from quarterly aggregation.
}
\label{fig:components_twitter}
\end{figure}

\begin{figure}[h!]
\centering\includegraphics[width=\textwidth]{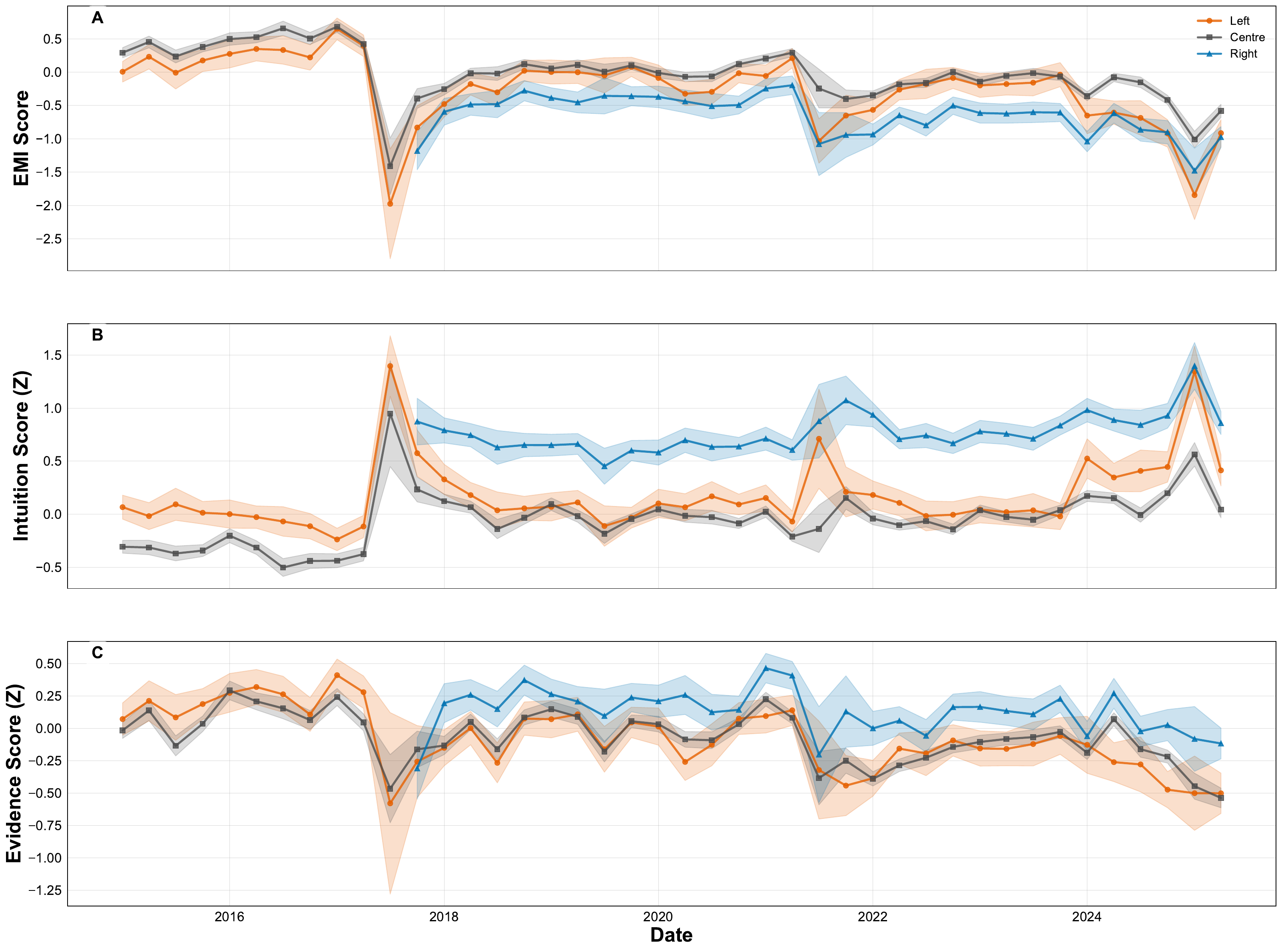}
\caption{
\textbf{Component-level rhetorical styles in parliamentary speeches by political leaning, 2015--2025.}
\textbf{A}, Quarterly averages of the Evidence Minus Intuition (EMI) score, \textbf{B}, intuition-based language similarity, and \textbf{C} evidence-based language similarity, with parties grouped by ideological leaning (left, centre, right). Shaded regions represent 95\% confidence intervals from quarterly aggregation.
}
\label{fig:components_parliament}
\end{figure}

The EMI score is a difference measure, so a decline can in principle arise from 
falling evidence-based language, rising intuition-based language, or both. To 
disentangle these components, Figures~\ref{fig:components_twitter} 
and~\ref{fig:components_parliament} display the quarterly averages of the two 
underlying $z$-standardised similarity scores separately, alongside the composite 
EMI, for the Twitter and Bundestag corpora. In both arenas the downward trend in 
EMI is driven by movement in both components rather than a single one: 
intuition-based similarity rises over the observation window while evidence-based 
similarity stagnates or declines, and the ordering across ideological camps is 
mirrored on both dimensions, with right-leaning actors showing the highest 
intuition and lowest evidence loadings throughout. This indicates that the 
convergence documented in the main text reflects a genuine shift in the balance 
of epistemic styles rather than an artefact of one component's variance.

\clearpage
\newpage
\subsection{Word Scatter and Keyness}
\begin{figure}
\centering\includegraphics[width=\textwidth]{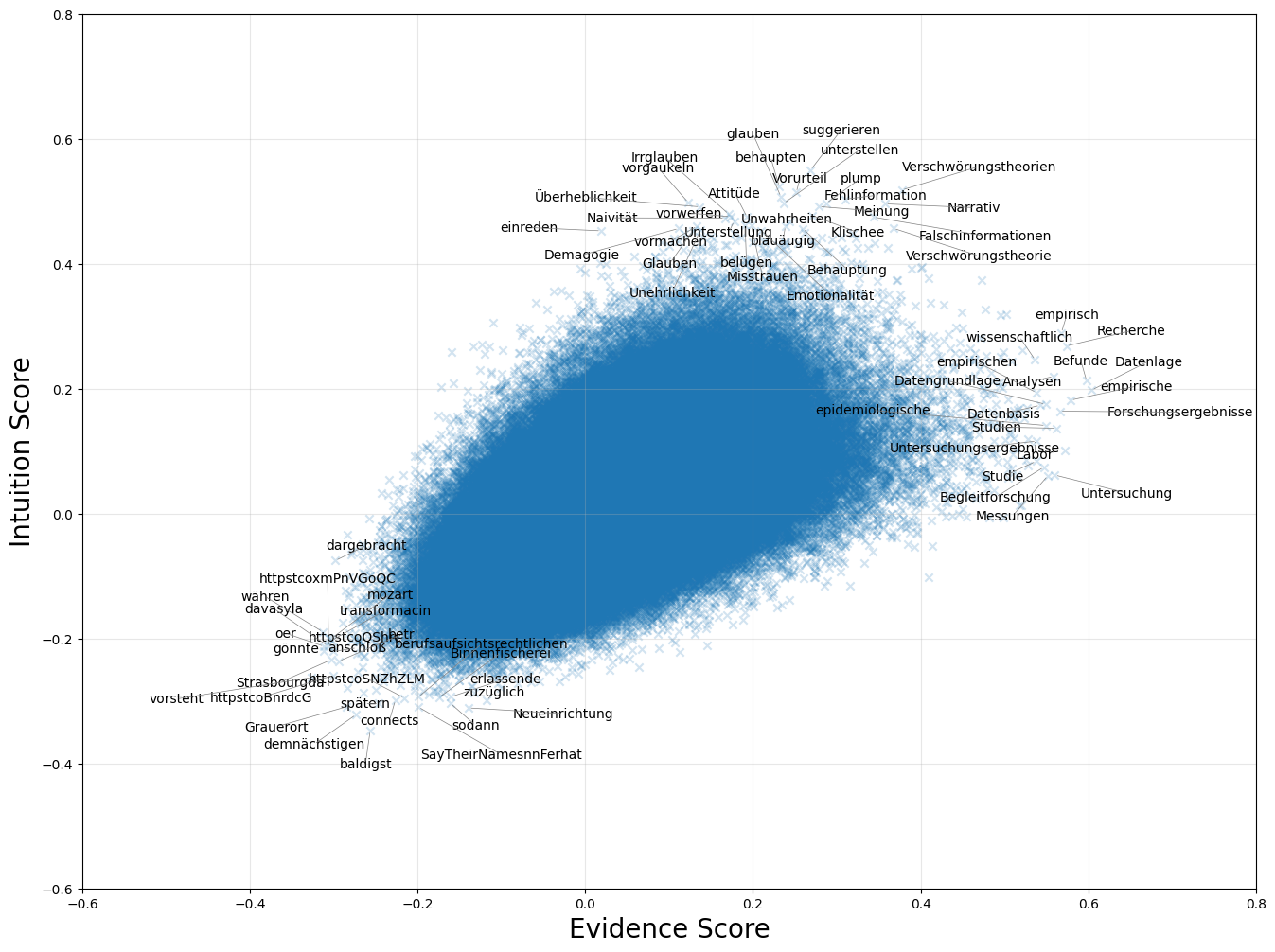}
    \caption{Scatter Twitter Corpus}
     \label{fig:sactter2}
\end{figure}

\begin{figure}
\centering\includegraphics[width=\textwidth]{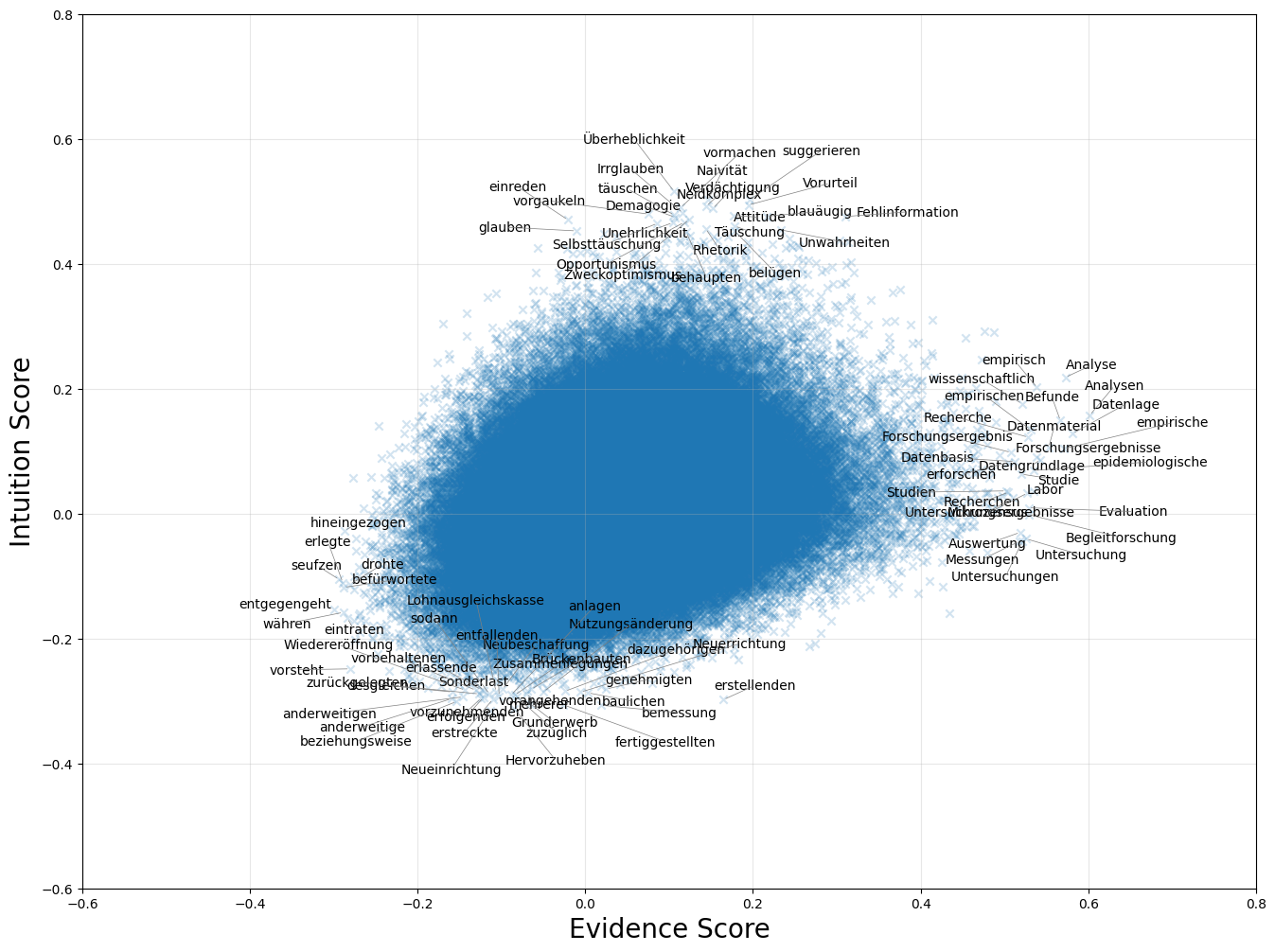}
    \caption{Scatter Parliament Corpus}
     \label{fig:sactter3}
\end{figure}

To visualize influential words we plot scatter plots where each word is represented by its raw evidence and intuition similarity. The plots \ref{fig:sactter2} and \ref{fig:sactter3} show that most words cluster around 0 for both constructs. Words with negative similarity do not tend to have a semantic connection to either evidence or intuition. For highly positive similarities distinct patterns can be observed.

\vspace{0.5cm}
Next, we compute two keyness measures: truth keyness and political keyness. These measures quantify the association of words with honesty-related and political discourse, respectively. The results were visualized using scatterplots where each word's coordinates correspond to its computed keyness values (see methods). To visualize the relationship between political and truth keyness, we generated scatterplots where each word \( w \) is represented as a point with coordinates \((PK_w, SFS_w)\). Points positioned towards the left or right on the x-axis indicate stronger political keyness in the respective direction, while points positioned towards the top or bottom on the y-axis indicate association with evidence-based or intuition-based rhetoric, respectively. The most extreme words, determined by their absolute keyness scores, were labeled to facilitate interpretation. Horizontal and vertical reference lines were added at \( x = 0 \) and \( y = 0 \) to denote the neutral position in both dimensions.
\vspace{0.5cm}
The keyness scatter plots in Figures \ref{fig:key2} and \ref{fig:key3} illustrate the distribution of words in relation to their political and rhetorical significance across different corpora: Twitter and parliamentary speeches. Highly influential words for evidence-based rhetoric include \textit{"Fakten"} (facts), \textit{"Studien"} (studies), \textit{"Daten"} (data), and \textit{"Beweise"} (evidence). These words frequently appear in factual statements, scientific discourse, and political debates grounded in empirical reasoning. In contrast, words associated with intuition-based rhetoric include \textit{"glauben"} (believe), \textit{"fühlen"} (feel), \textit{"Meinung"} (opinion), and \textit{"wahrnehmen"} (perceive). These terms reflect subjective viewpoints, personal beliefs, and less structured argumentation. The scatterplots further differentiate political speech patterns across platforms. For instance, \textbf{Twitter discourse} tends to contain more polarizing and emotionally charged words, reflecting the informal and often reactive nature of the platform. \textbf{Parliamentary speeches} show a distinct pattern where high-truth-keyness words cluster around legislative and policy-related discussions.

\vspace{0.5cm}
The Word Mover’s Distance (WMD) plots seen in Figure \ref{fig:wmd_words} complement these scatter plots by quantifying semantic distances between words within different corpora. The results show that speeches tend to have the highest internal coherence, whereas social media platforms exhibit greater variance in WMD scores due to informal language, satire, and emotionally charged rhetoric. Across all datasets, words highly ranked in keyness (such as \textit{"Beweise"} for evidence and \textit{"Meinung"} for intuition) show consistency across the scatter and WMD analyses, reinforcing their relevance in distinguishing rhetorical styles.

\newpage

\begin{landscape}  %
    \begin{figure}
        \centering
        \includegraphics[width=700pt]{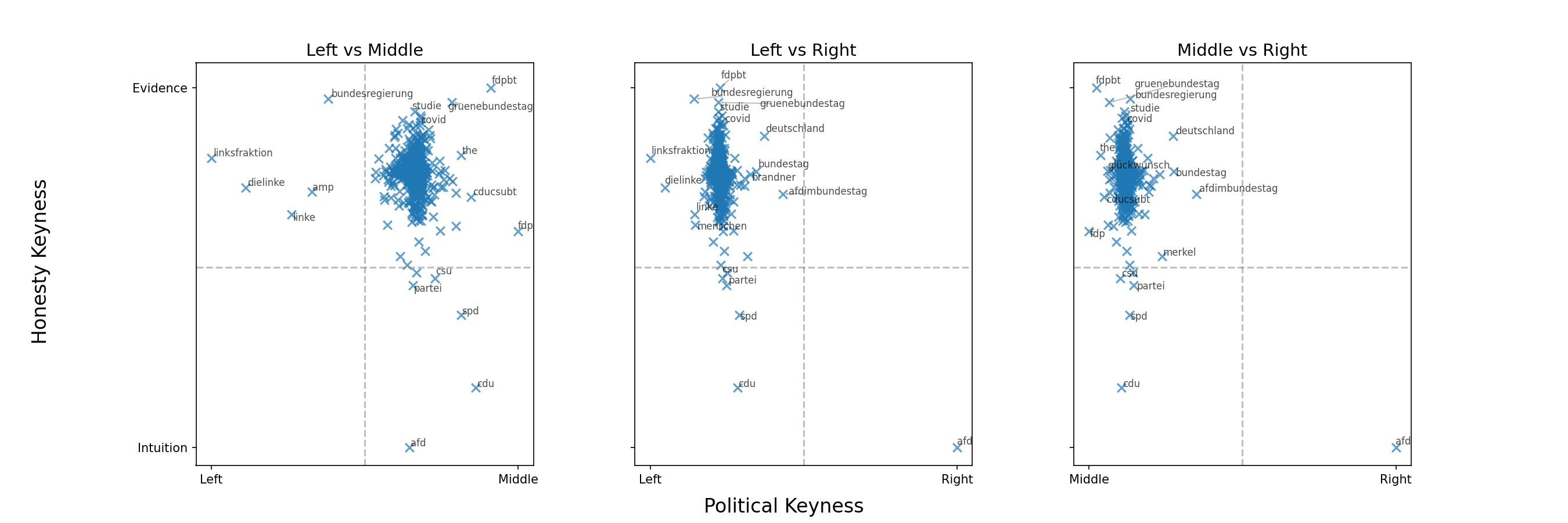}
        \caption{Scatterplots of Political vs. Honesty Keyness Twitter}
        \label{fig:key2}
    \end{figure}

    \begin{figure}
        \centering
        \includegraphics[width=700pt]{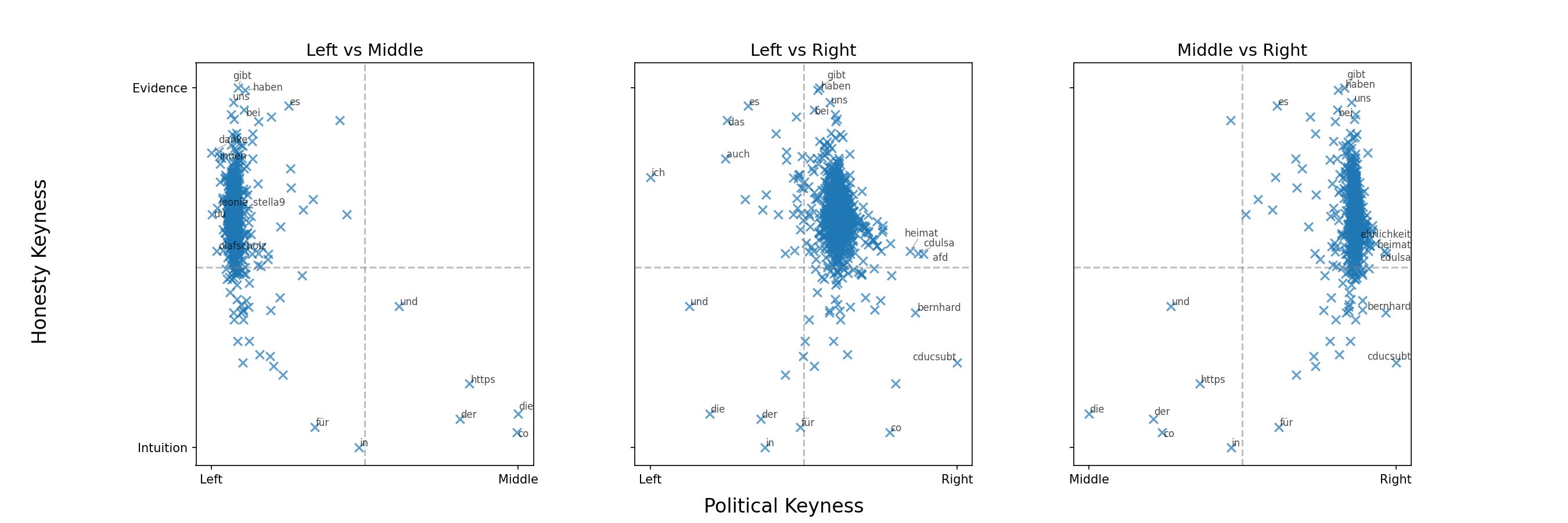}
        \caption{Scatterplots of Political vs. Honesty Keyness Speeches}
         \label{fig:key3}
    \end{figure}

    \begin{figure}
        \centering
        \includegraphics[width=600pt]{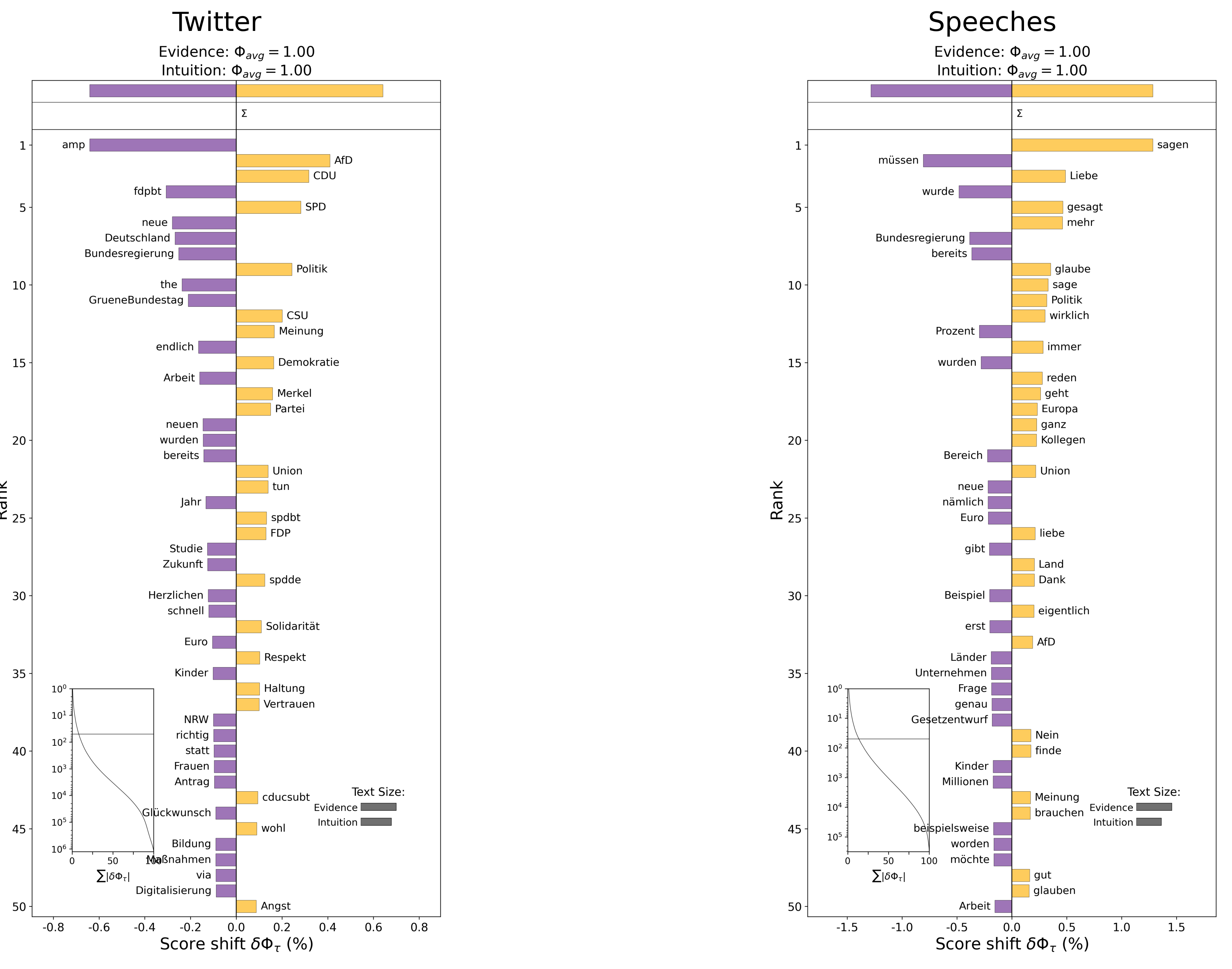}
        \caption{Word mover graphs}
        \label{fig:wmd_words}
    \end{figure}
    
\end{landscape}

\subsection{Clipping Negative Similarities}
When working with word embeddings, cosine similarity is commonly used to measure semantic relatedness. However, negative cosine similarities pose an interpretational challenge. While positive similarities indicate some degree of semantic alignment between the compared embeddings, negative similarities lack a straightforward linguistic interpretation. In our scatter plots, we observed that negative similarities did not exhibit a clear semantic pattern, further complicating their meaningful interpretation. At the document level, we first computed the cosine similarity between each word embedding in the document and its corresponding dictionary embedding. Instead of averaging all similarities, we retained only positive values for aggregation. This robustness check was performed across all both datasets.

\begin{figure}[]
    \centering
    \includegraphics[width=\textwidth]{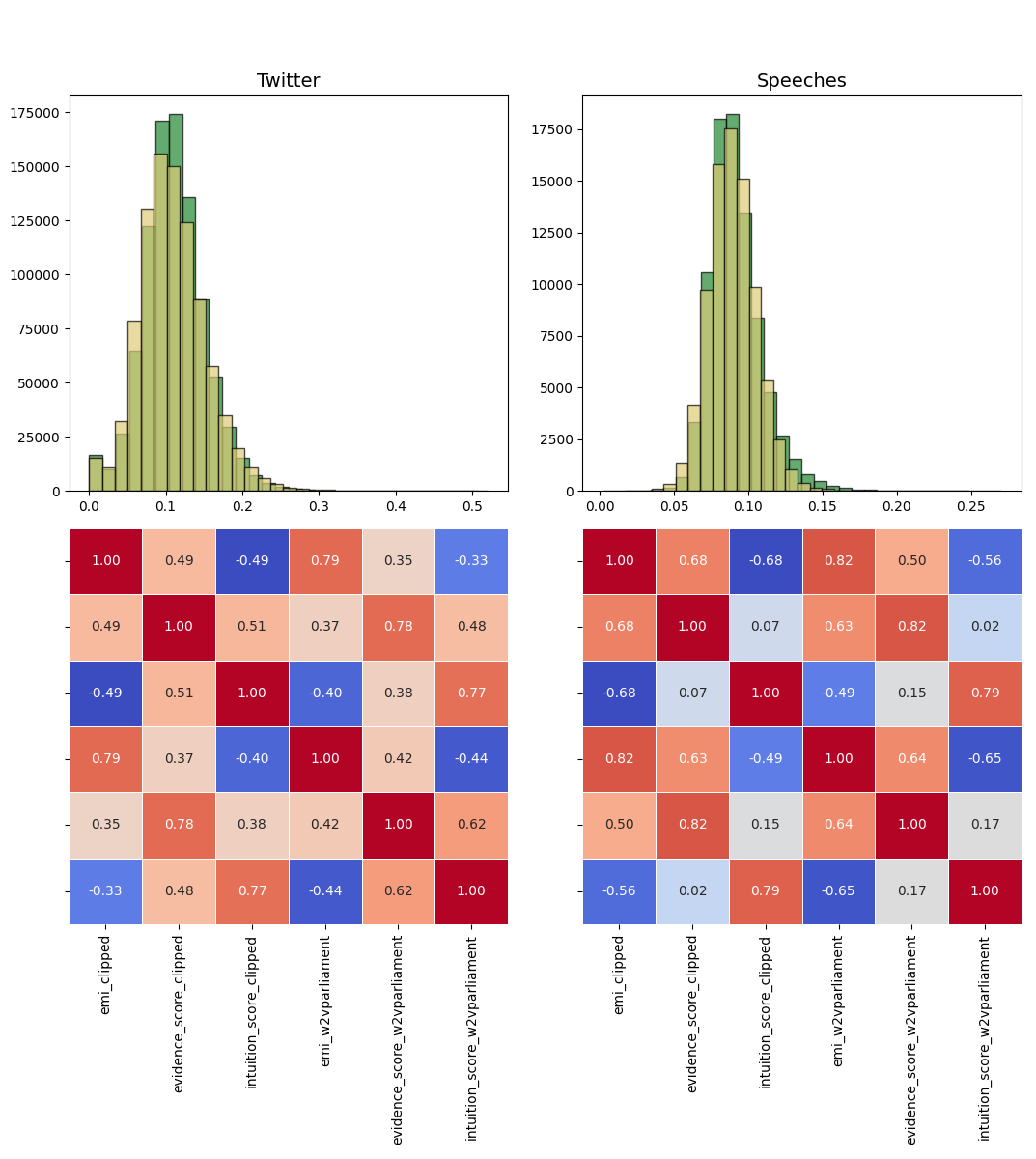}
    \caption{Clipping Correlation}
    \label{fig:clipping_corr}
\end{figure}

Figure \ref{fig:clipping_corr} presents a comparison of evidence and intuition scores across two datasets: Twitter and Speeches. The histograms at the top illustrate the distribution of evidence and intuition scores after removing negative similarity words for averaging embeddings, showing that both distributions follow a roughly normal shape, with intuition scores generally being slightly lower than evidence scores. Below the histograms, correlation heatmaps display the relationships between different scoring methods, including the clipped and non-clipped EMI scores.
The heatmaps demonstrate that clipping leads to consistent lower correlations between intuition and evidence across both datasets, ensuring that evidence and intuition scores remain interpretable and aligned with meaningful semantic relationships. Notably, the correlation between clipped and non-clipped scores remains strong, indicating that while clipping could reduce noise, it does not distort the underlying ranking of documents in terms of their evidence-based or intuition-based associations. One possible reason for this is that negative similarities may arise due to the high-dimensional nature of word embeddings, where orthogonality and unrelated meanings can lead to spurious negative values. Additionally, embedding models trained on large corpora do not necessarily enforce a strict symmetry between opposite meanings, meaning that a negative similarity does not always imply conceptual opposition. Instead, negative values could emerge from random distributional effects, adding noise rather than conveying structured semantic information. This could also explain why the correlation between intuition and evidence is lower for clipped scores. When we compare the bimodular distribution of original scores, one can assume that there are frequent high negative occurring words for both dimensions. This consistency across datasets indicates that the clipping procedure does not distort the general trajectory of evidence-based and intuition-driven rhetoric but rather enhances interpretability by limiting the influence of extreme cosine similarity values.

\newpage

\subsection{Bootstrapping Word Level}
To ensure robustness, we bootstrapped the intuition and evidence similarities for all words in the corpus. This was achieved by first sampling a dictionary length between 30 and the maximum possible length minus one across 1000 iterations. Subsequently, words were randomly drawn to match the selected length for both the evidence and intuition dictionary. This resampling process was repeated across different dictionary sizes to confirm that the observed similarities were not driven by specific words or dictionary length but rather reflected broader trends in the corpus.
To quantify robustness, we calculated the mean variance of evidence and intuition similarities across words and compared it to the total variance. The key values computed include the Mean Variance (MV), representing the average variance of the bootstrapped samples across words, and the Total Variance (TV), which denotes the variance of the mean values across all words in the full dataset. The Bootstrap Variance Ratio (BVR) is computed as the ratio of MV to TV:
\begin{equation}
BVR_{X} = \frac{MV_{X}}{TV_{X}}, \quad X \in {\text{evidence}, \text{intuition}}
\end{equation}
A lower BVR suggests stable bootstrap estimates, whereas a higher BVR indicates greater variability in resampling.

\vspace{0.5cm}
The Twitter dataset showed a Mean Variance for Evidence of 0.0001 (95\% CI = [0.000107, 0.000107]), Total Variance of 0.0055, and a BVR of 0.0195 (95\% CI = [0.019490, 0.019532]). For Intuition, the Mean Variance was 0.0001 (95\% CI = [0.000069, 0.000069]), the Total Variance was 0.0063, and the BVR was 0.0109 (95\% CI = [0.010918, 0.010941]). The Speeches dataset had a Mean Variance for Evidence of 0.0002 (95\% CI = [0.000227, 0.000228]), with a Total Variance of 0.0072, leading to a BVR of 0.0316 (95\% CI = [0.031588, 0.031694]). Similarly, the Mean Variance for Intuition was 0.0001 (95\% CI = [0.000142, 0.000142]), with a Total Variance of 0.0073, resulting in a BVR of 0.0195 (95\% CI = [0.019445, 0.019512]). The consistently low BVR values across datasets indicate that bootstrap samples are stable, reinforcing the robustness of the methodology and confirming that resampling does not introduce excessive variability in estimates. Additionally, the confidence intervals demonstrate that evidence and intuition similarities are not artifacts of a few highly influential words in the dictionary but remain stable across different dictionary compositions. Figures \ref{fig:bxplt1} and \ref{fig:bxplt3} shows the Box plots of the most influential words for 0.25 quantiles. It can be observed that, for most words, the range is really low. This analysis demonstrates that evidence and intuition similarities are not artifacts of a few highly influential words in the dictionary but remain stable across different dictionary compositions.

\begin{landscape}  %
    \begin{figure}
        \centering
        \includegraphics[width=500pt]{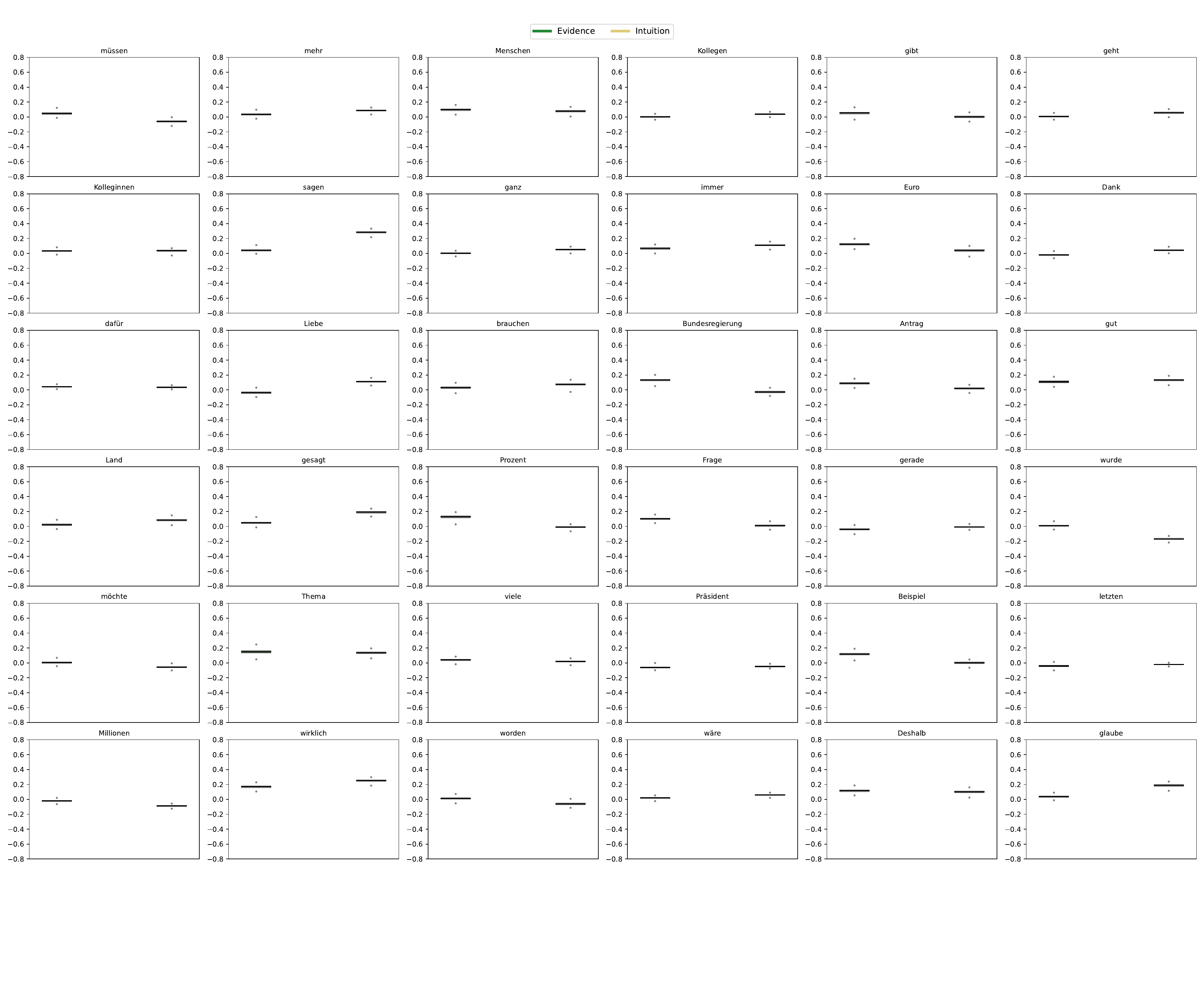}
        \caption{Boxplots Parliament Corpus}
        \label{fig:bxplt1}
    \end{figure}

    \begin{figure}
        \centering
        \includegraphics[width=500pt]{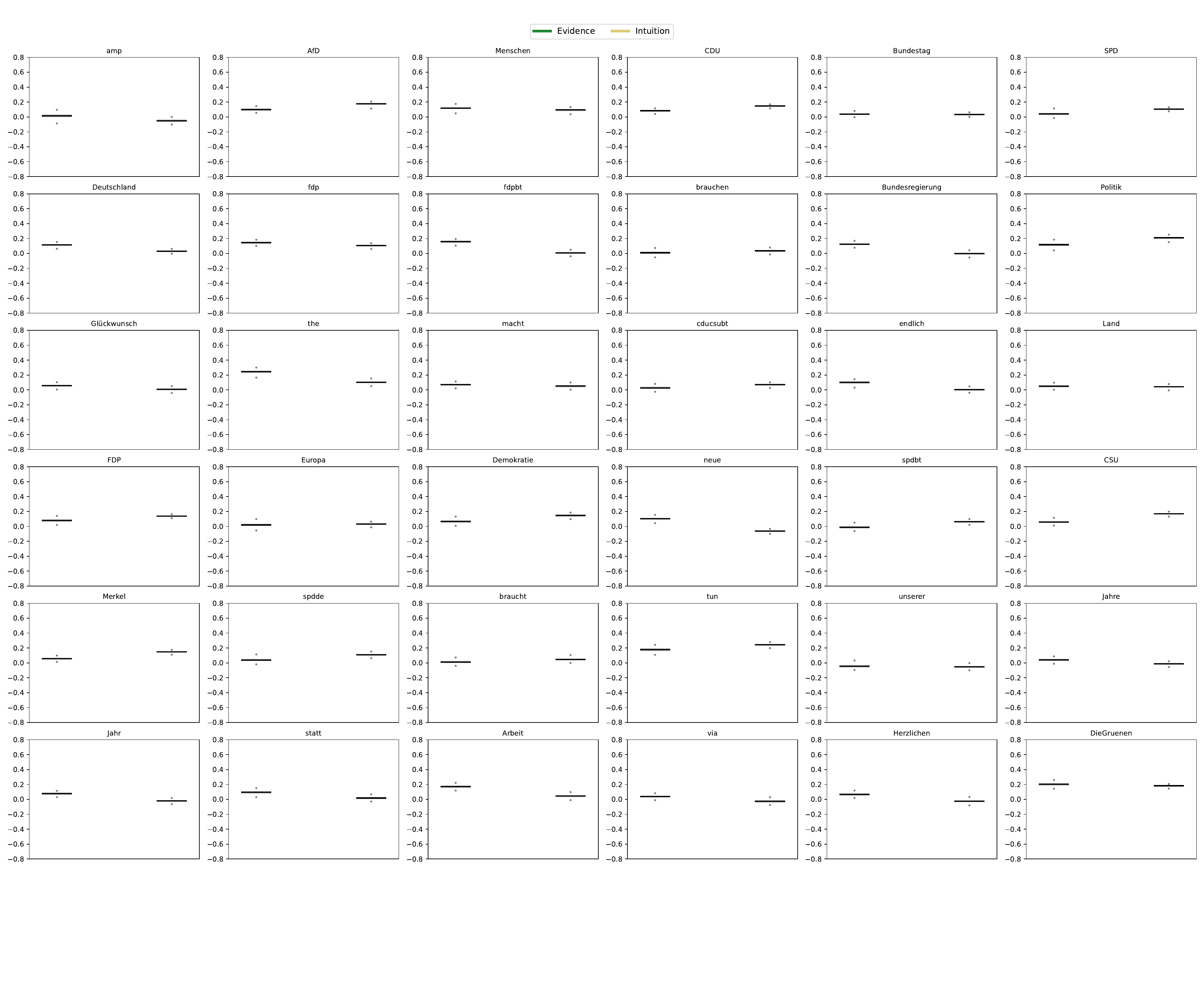}
        \caption{Boxplots Twitter Corpus}
        \label{fig:bxplt3}
    \end{figure}
    
\end{landscape}

\subsection{Bootstrapping Document Level}
\begin{figure}[h!]
    \centering
    \includegraphics[width=\textwidth]{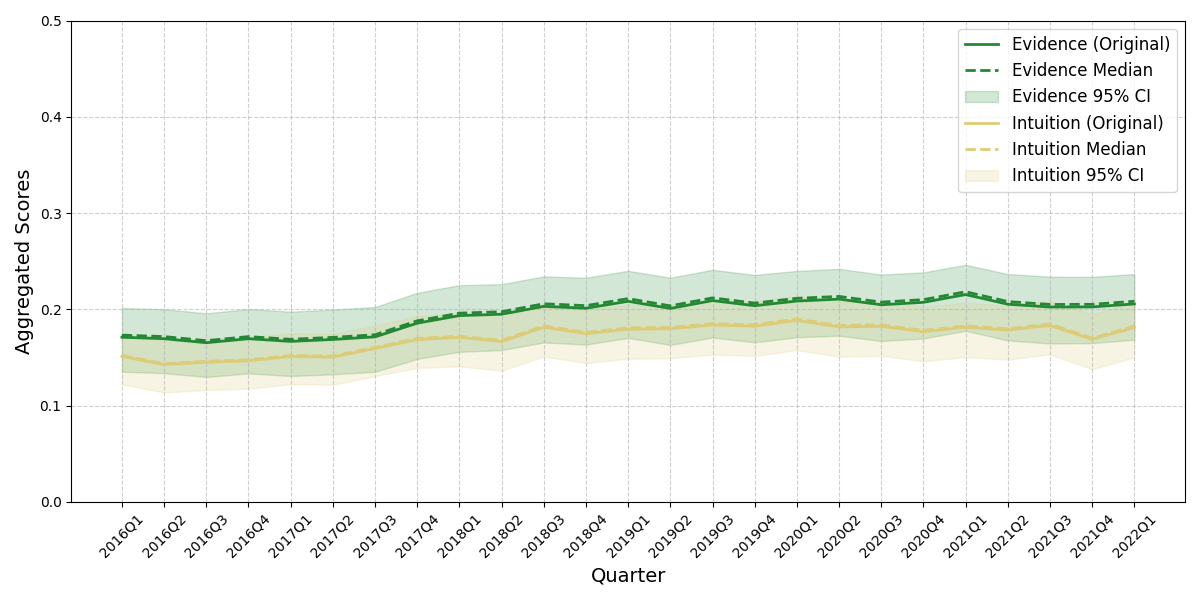}
    \caption{Intuition and Evidence Timeline with Bootstrapped CIs Twitter}
    \label{fig:bci2}
\end{figure}

\begin{figure}[h!]
    \centering
    \includegraphics[width=\textwidth]{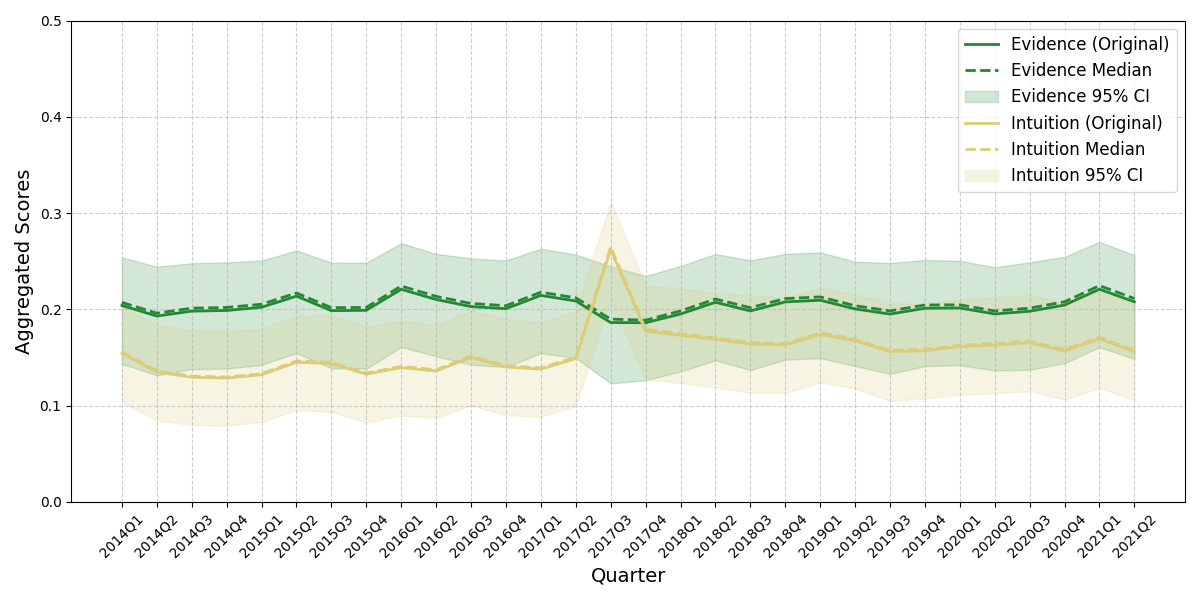}
    \caption{Intuition and Evidence Timeline with Bootstrapped CIs Parliament Speeches}
    \label{fig:bci3}
\end{figure}

To ensure robustness at the document level, we applied bootstrapping to document similarity calculations. This involved recalculating the document scores for each bootstrap iteration done on the word level. Figures \ref{fig:bci2} and \ref{fig:bci3} show similarities in intuition and evidence over time, this time incorporating confidence intervals derived from the bootstrapping procedure described above, rather than simple mean aggregation. The confidence intervals are relatively small, indicating low variability across bootstrapped samples. Furthermore, the median of all 1000 iterations closely aligns with the original value, demonstrating that the estimated similarities remain highly stable across resampled dictionaries. This consistency across iterations underscores the robustness of the EMI score, confirming that its trends over time are not driven by specific word selections but rather reflect broader pattern

\newpage

\subsection{Historical Parliament Speeches Analysis}
\begin{figure}[H]
    \centering
    \includegraphics[width=15cm]{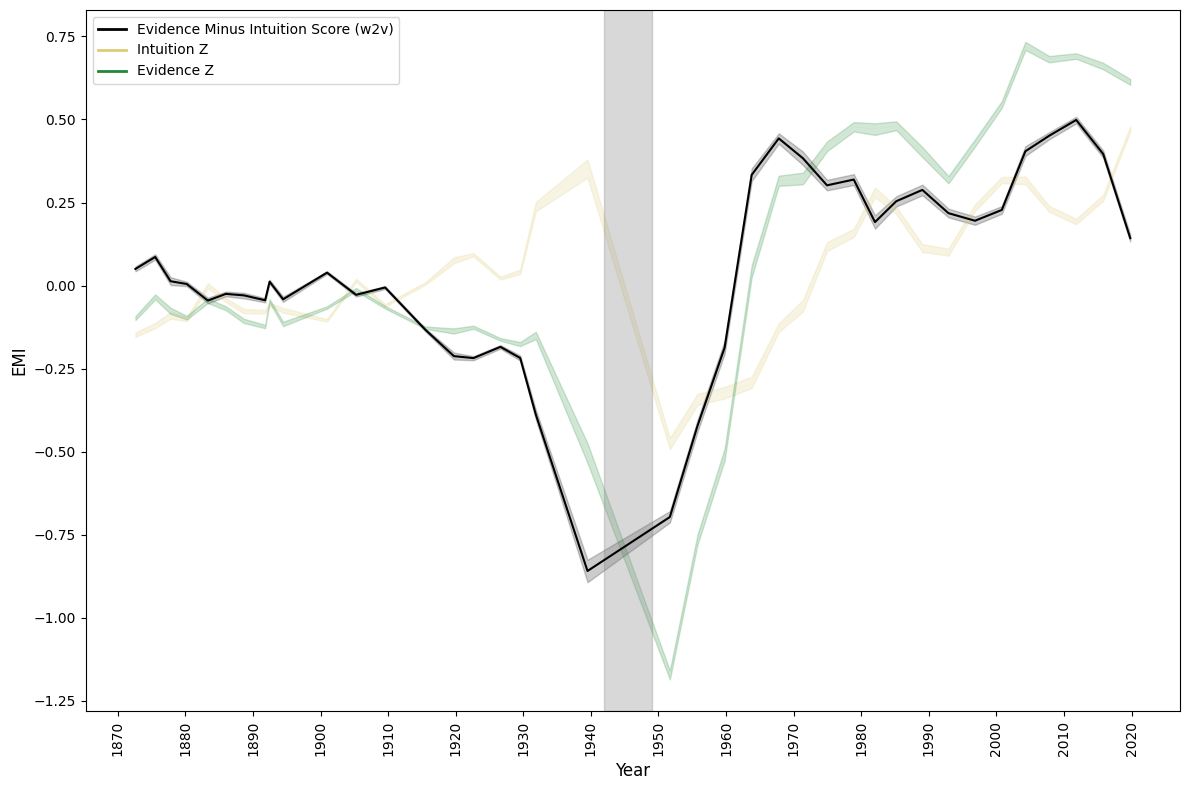}
    \caption{Evidence, Intuition, and EMI in Historical Speeches Averaged by Legislative Period}
    \label{fig:histo}
\end{figure}

We computed the EMI score for all speeches in the parliamentary dataset. Plot \ref{fig:histo} reveals a sharp decline in EMI around the 1930s–1940s, driven by a rise in intuition and a drop in evidence-based language. This aligns with theories about Nazi-era rhetoric, which emphasized internal, emotional "truths" over factual reasoning. The period serves as a striking example of how overreliance on intuition can endanger democratic discourse, and offers a compelling validation for the EMI metric.

\newpage
\subsection{Validation Survey Keyword Level}
What follows is a verbatim copy of the instructions provided to the participants who rated the German keywords.
The instructions were presented in German.
The two language dimensions are referred to by their German equivalents \textit{Evidenz} and \textit{Intuition}, corresponding to the evidence-based and intuition-based components of the EMI score used in this work.

\bigskip
\noindent\rule{\textwidth}{0.4pt}
\bigskip

\noindent Menschen können unterschiedliche Vorstellungen davon haben, was es bedeutet, die \textit{Wahrheit} zu sagen.
Wir konzentrieren uns auf zwei solche Vorstellungen.

\medskip
\noindent Die eine basiert auf Intuition, Bauchgefühl und Authentizität.
Gemäß dieser Vorstellung sagen Menschen die Wahrheit, wenn sie \glqq sagen, was sie im Moment für wahr hielten\grqq.
Ob Aussagen eine korrekte Widerspiegelung der Realität sind, ist dabei weniger wichtig.
Wir nennen diese Vorstellung von Wahrheit \textbf{Intuition}.

\medskip
\noindent Die andere basiert auf Evidenz, Analyse und Faktentreue.
Gemäß dieser Vorstellung sagen Menschen die Wahrheit, wenn ihre Aussagen mit den vorhandenen Belegen übereinstimmen.
Ob Aussagen eine authentische Widerspiegelung der eigenen Gefühle sind, ist dabei weniger wichtig.
Wir nennen diese Vorstellung von Wahrheit \textbf{Evidenz}.

\medskip
\noindent Ihre Aufgabe ist es, für jedes der folgenden Wörter zu beurteilen, welcher Vorstellung von Wahrheit es am ehesten entspricht.
Wenn jemand dieses Wort verwendet -- spiegelt es eher \textbf{Intuition} wieder?
Oder spiegelt das Wort eher \textbf{Evidenz} wieder?

\medskip
\noindent Bitte geben Sie für jedes Wort an, welcher Vorstellung es am nächsten ist, indem Sie für jede Spalte einen Wert von 1 bis 5 wählen, wobei 1 bedeutet, dass das Wort \textit{am wenigsten} repräsentativ für diese Kategorie ist, und 5 bedeutet, dass das Wort \textit{sehr repräsentativ} für diese Kategorie ist.
Es gibt keine richtigen oder falschen Antworten, wir sind an Ihrer Einschätzung der Bedeutung dieser Wörter interessiert.

\bigskip
\noindent\rule{\textwidth}{0.4pt}
\bigskip

\begin{figure}[H]
    \centering
    \includegraphics[width=15cm]{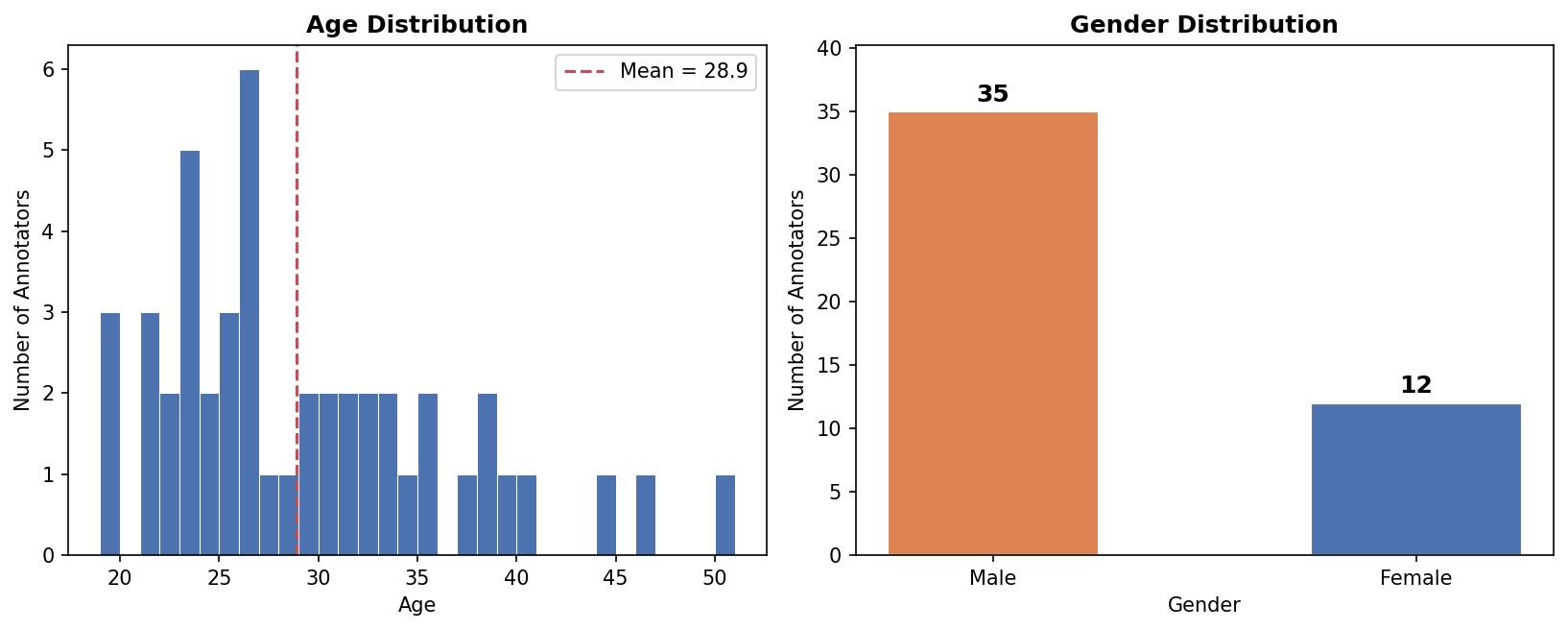}
    \caption{Demographics of the 47 German-speaking annotators recruited via Prolific. \textbf{Left:} Age distribution (mean\,=\,28.9 years, s.d.\,=\,7.4 years, range 19--50). \textbf{Right:} Gender composition (35 women, 12 non-binary/diverse).}
\label{fig:annotator_demographics}
    \label{fig:dist_key}
\end{figure}

\begin{figure}[H]
    \centering
    \includegraphics[width=15cm]{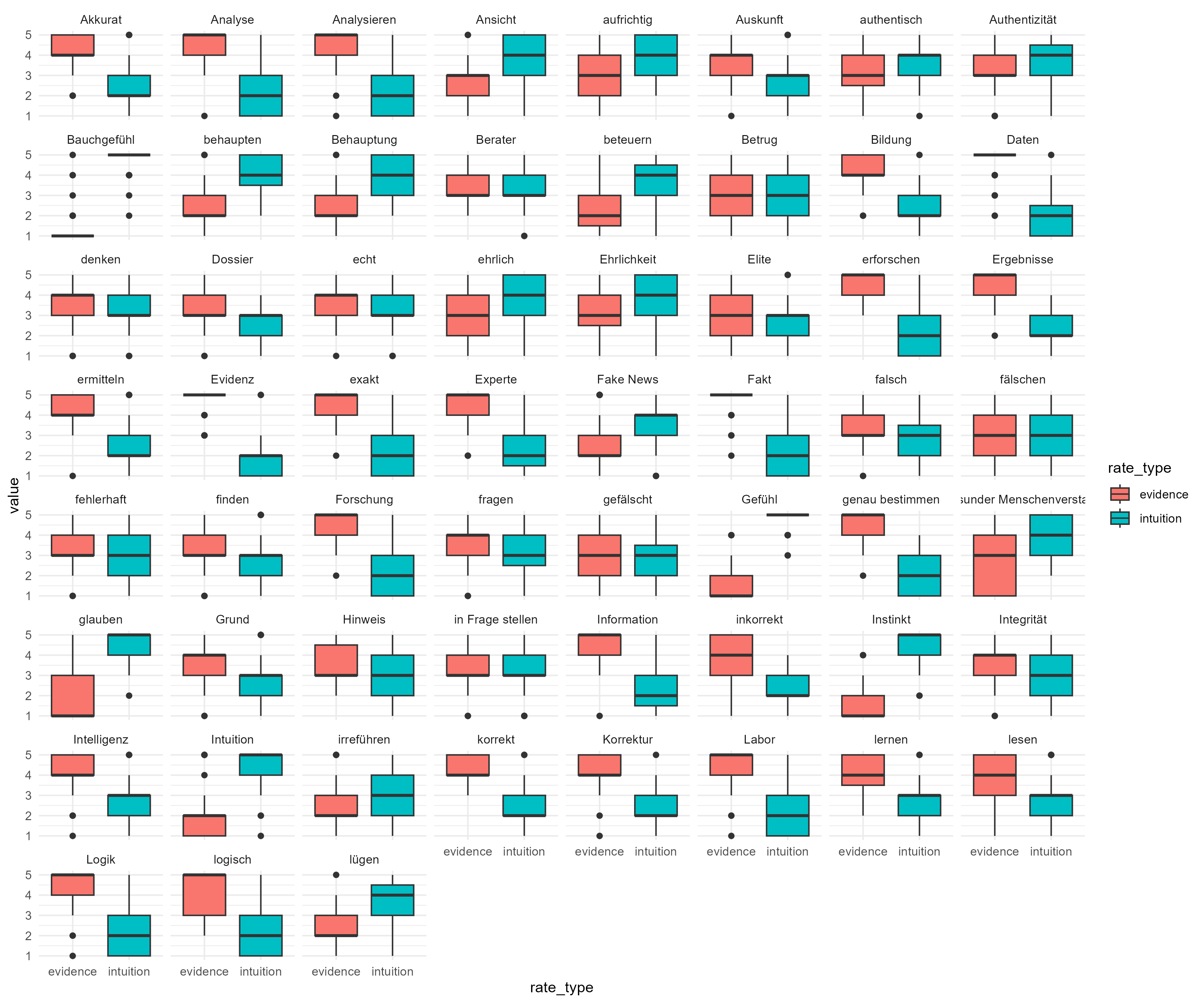}
    \caption{Distributions of \textit{Evidenz} (orange) and \textit{Intuition} (blue) representativeness ratings for German keywords beginning with letters A--L ($N = 47$ annotators, five-point Likert scale). Each panel shows one keyword; boxes indicate the interquartile range, horizontal lines the median, and whiskers extend to 1.5\,IQR. Keywords for which the two rating distributions differed significantly ($p < .05$, Holm-corrected paired $t$-test) were retained in their respective dictionaries.}
    \label{fig:box_key1}
\end{figure}

\begin{figure}[H]
    \centering
    \includegraphics[width=15cm]{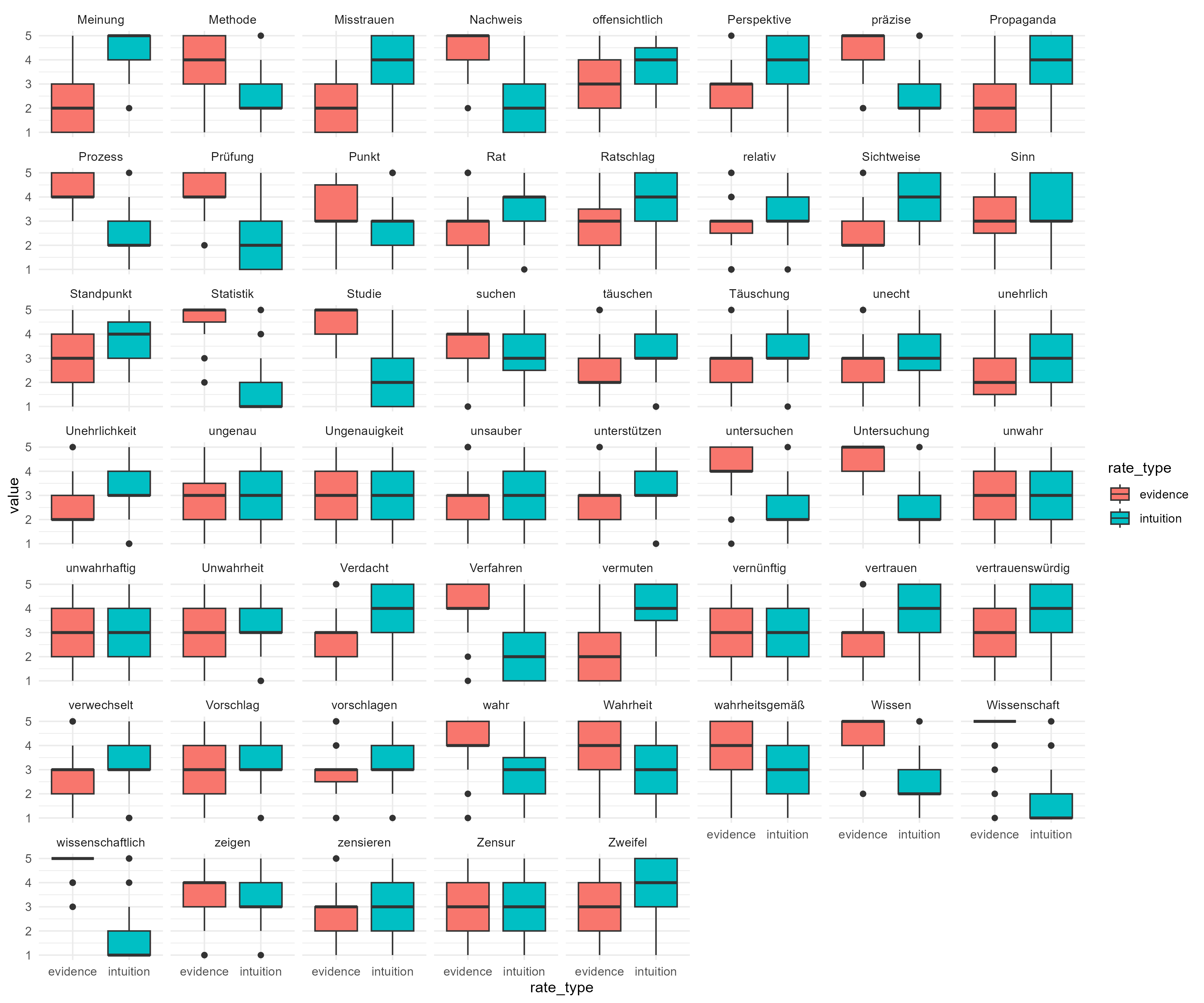}
    \caption{Distributions of \textit{Evidenz} (orange) and \textit{Intuition} (blue) representativeness ratings for German keywords beginning with letters M--Z ($N = 47$ annotators, five-point Likert scale). Each panel shows one keyword; boxes indicate the interquartile range, horizontal lines the median, and whiskers extend to 1.5\,IQR. Keywords for which the two rating distributions differed significantly ($p < .05$, Holm-corrected paired $t$-test) were retained in their respective dictionaries.}
    \label{fig:box_key2}
\end{figure}

Participants were presented with all 113 German candidate keywords one at a time in randomised order.
For each keyword they completed two rating tasks on the same screen: they rated how representative the word was of \textit{Evidenz}-based language and of \textit{Intuition}-based language, each on a five-point Likert scale (1\,=\,not at all representative, 5\,=\,highly representative).
The survey was administered online via the Prolific platform and took approximately [X] minutes to complete; participants received monetary compensation at the standard Prolific rate. Data were collected on [DATE] from 47 German-speaking individuals.
Figure~\ref{fig:annotator_demographics} shows the demographic composition of the sample. Annotators ranged in age from 19 to 50 years (mean\,=\,28.9, s.d.\,=\,7.4) and comprised 35 women and 12 non-binary/diverse participants. To assess whether annotators differentiated the two rating dimensions within each keyword, we performed Holm-corrected paired $t$-tests comparing \textit{Evidenz} and \textit{Intuition} ratings for each of the 113 keywords. The full results are reported in Table~\ref{tab:kw_validation_german}, and the rating distributions for each keyword are displayed in Figures~\ref{fig:box_key1} and~\ref{fig:box_key2}, grouped alphabetically (A--L and M--Z, respectively). Of the 113 keywords, 86 showed a significant difference between the two rating dimensions ($p < .05$, Holm-corrected), with $\Delta M$ ranging from $-3.43$ (\textit{Bauchgefühl}; \textit{Intuition} $\gg$ \textit{Evidenz}) to $2.87$ (\textit{wissenschaftlich}; \textit{Evidenz} $\gg$ \textit{Intuition}). All 86 significant keywords were confirmed in the dictionary to which they had originally been assigned; no keyword was reclassified to the opposite category.
The remaining 27 keywords that did not reach the significance threshold were removed from the dictionaries, yielding final lists of 48 \textit{Evidenz} terms and 38 \textit{Intuition} terms.

\begin{longtable}{l l r r r r r}
\caption{Paired $t$-test results: evidence vs.\ intuition ratings for German keywords ($N=47$ annotators, Holm-adjusted).}\\
\label{tab:kw_validation_german}\\
\toprule
Keyword & Component & $\Delta M$ & $t$ & $df$ & 95\,\% CI & $p$ \\
\midrule
\endfirsthead
\multicolumn{7}{c}{\tablename\ \thetable{} -- \textit{continued}} \\
\toprule
Keyword & Component & $\Delta M$ & $t$ & $df$ & 95\,\% CI & $p$ \\
\midrule
\endhead
\midrule
\multicolumn{7}{r}{\textit{Continued on next page}} \\
\endfoot
\bottomrule
\endlastfoot
\multicolumn{7}{l}{\textit{Evidence keywords}} \\
\midrule
Akkurat & Evidence & 1.49 & 6.87 & 46 & [1.05,\ 1.93] & \textless{}.001$^{***}$ \\
Analyse & Evidence & 2.62 & 11.98 & 46 & [2.18,\ 3.06] & \textless{}.001$^{***}$ \\
Analysieren & Evidence & 2.32 & 10.00 & 46 & [1.85,\ 2.79] & \textless{}.001$^{***}$ \\
Auskunft & Evidence & 0.81 & 3.21 & 46 & [0.30,\ 1.32] & 0.002$^{**}$ \\
Berater & Evidence & 0.19 & 0.89 & 46 & [-0.24,\ 0.62] & 0.376 \\
Betrug & Evidence & 0.09 & 0.28 & 46 & [-0.52,\ 0.69] & 0.779 \\
Bildung & Evidence & 1.77 & 8.64 & 46 & [1.35,\ 2.18] & \textless{}.001$^{***}$ \\
Daten & Evidence & 2.74 & 12.70 & 46 & [2.31,\ 3.18] & \textless{}.001$^{***}$ \\
Dossier & Evidence & 1.02 & 4.39 & 46 & [0.55,\ 1.49] & \textless{}.001$^{***}$ \\
Elite & Evidence & 0.11 & 0.43 & 46 & [-0.40,\ 0.61] & 0.672 \\
Ergebnisse & Evidence & 2.26 & 11.25 & 46 & [1.85,\ 2.66] & \textless{}.001$^{***}$ \\
Evidenz & Evidence & 2.81 & 13.82 & 46 & [2.40,\ 3.22] & \textless{}.001$^{***}$ \\
Experte & Evidence & 2.04 & 8.33 & 46 & [1.55,\ 2.54] & \textless{}.001$^{***}$ \\
Fakt & Evidence & 2.74 & 12.01 & 46 & [2.28,\ 3.20] & \textless{}.001$^{***}$ \\
Forschung & Evidence & 2.55 & 10.59 & 46 & [2.07,\ 3.04] & \textless{}.001$^{***}$ \\
Grund & Evidence & 0.79 & 3.03 & 46 & [0.26,\ 1.31] & 0.004$^{**}$ \\
Hinweis & Evidence & 0.70 & 2.72 & 46 & [0.18,\ 1.22] & 0.009$^{**}$ \\
Information & Evidence & 1.96 & 7.99 & 46 & [1.46,\ 2.45] & \textless{}.001$^{***}$ \\
Integrität & Evidence & 0.38 & 1.53 & 46 & [-0.12,\ 0.89] & 0.132 \\
Intelligenz & Evidence & 1.49 & 6.73 & 46 & [1.04,\ 1.93] & \textless{}.001$^{***}$ \\
Korrektur & Evidence & 1.55 & 6.30 & 46 & [1.06,\ 2.05] & \textless{}.001$^{***}$ \\
Labor & Evidence & 2.23 & 8.31 & 46 & [1.69,\ 2.78] & \textless{}.001$^{***}$ \\
Logik & Evidence & 1.74 & 6.10 & 46 & [1.17,\ 2.32] & \textless{}.001$^{***}$ \\
Methode & Evidence & 1.51 & 5.70 & 46 & [0.98,\ 2.04] & \textless{}.001$^{***}$ \\
Nachweis & Evidence & 2.32 & 11.48 & 46 & [1.91,\ 2.73] & \textless{}.001$^{***}$ \\
Prozess & Evidence & 1.85 & 10.15 & 46 & [1.48,\ 2.22] & \textless{}.001$^{***}$ \\
Prüfung & Evidence & 1.91 & 7.05 & 46 & [1.37,\ 2.46] & \textless{}.001$^{***}$ \\
Punkt & Evidence & 0.74 & 3.10 & 46 & [0.26,\ 1.23] & 0.003$^{**}$ \\
Statistik & Evidence & 2.74 & 11.60 & 46 & [2.27,\ 3.22] & \textless{}.001$^{***}$ \\
Studie & Evidence & 2.47 & 10.96 & 46 & [2.01,\ 2.92] & \textless{}.001$^{***}$ \\
Untersuchung & Evidence & 2.23 & 11.58 & 46 & [1.85,\ 2.62] & \textless{}.001$^{***}$ \\
Verfahren & Evidence & 2.02 & 9.26 & 46 & [1.58,\ 2.46] & \textless{}.001$^{***}$ \\
Wahrheit & Evidence & 0.89 & 3.30 & 46 & [0.35,\ 1.44] & 0.002$^{**}$ \\
Wissen & Evidence & 1.98 & 9.87 & 46 & [1.57,\ 2.38] & \textless{}.001$^{***}$ \\
Wissenschaft & Evidence & 2.68 & 10.60 & 46 & [2.17,\ 3.19] & \textless{}.001$^{***}$ \\
Zensur & Evidence & 0.09 & 0.27 & 46 & [-0.56,\ 0.73] & 0.792 \\
denken & Evidence & 0.06 & 0.25 & 46 & [-0.45,\ 0.57] & 0.802 \\
echt & Evidence & 0.49 & 2.21 & 46 & [0.04,\ 0.93] & 0.032$^{*}$ \\
erforschen & Evidence & 2.23 & 9.31 & 46 & [1.75,\ 2.72] & \textless{}.001$^{***}$ \\
ermitteln & Evidence & 1.87 & 7.49 & 46 & [1.37,\ 2.38] & \textless{}.001$^{***}$ \\
exakt & Evidence & 2.11 & 8.15 & 46 & [1.59,\ 2.63] & \textless{}.001$^{***}$ \\
falsch & Evidence & 0.45 & 1.81 & 46 & [-0.05,\ 0.94] & 0.077 \\
fehlerhaft & Evidence & 0.45 & 1.69 & 46 & [-0.09,\ 0.98] & 0.098 \\
finden & Evidence & 0.49 & 1.86 & 46 & [-0.04,\ 1.02] & 0.069 \\
fragen & Evidence & 0.32 & 1.26 & 46 & [-0.19,\ 0.83] & 0.213 \\
fälschen & Evidence & 0.13 & 0.43 & 46 & [-0.46,\ 0.72] & 0.666 \\
gefälscht & Evidence & 0.66 & 2.10 & 46 & [0.03,\ 1.29] & 0.041$^{*}$ \\
genau bestimmen & Evidence & 2.30 & 10.10 & 46 & [1.84,\ 2.76] & \textless{}.001$^{***}$ \\
in Frage stellen & Evidence & 0.21 & 0.82 & 46 & [-0.31,\ 0.74] & 0.417 \\
inkorrekt & Evidence & 1.17 & 4.80 & 46 & [0.68,\ 1.66] & \textless{}.001$^{***}$ \\
korrekt & Evidence & 1.77 & 8.19 & 46 & [1.33,\ 2.20] & \textless{}.001$^{***}$ \\
lernen & Evidence & 1.36 & 6.13 & 46 & [0.91,\ 1.81] & \textless{}.001$^{***}$ \\
lesen & Evidence & 1.09 & 4.47 & 46 & [0.60,\ 1.57] & \textless{}.001$^{***}$ \\
logisch & Evidence & 1.70 & 6.15 & 46 & [1.14,\ 2.26] & \textless{}.001$^{***}$ \\
prÃ¤zise & Evidence & 1.87 & 7.60 & 46 & [1.38,\ 2.37] & \textless{}.001$^{***}$ \\
suchen & Evidence & 0.47 & 2.20 & 46 & [0.04,\ 0.90] & 0.033$^{*}$ \\
untersuchen & Evidence & 1.91 & 7.76 & 46 & [1.42,\ 2.41] & \textless{}.001$^{***}$ \\
wahr & Evidence & 1.13 & 5.06 & 46 & [0.68,\ 1.58] & \textless{}.001$^{***}$ \\
wahrheitsgemäß & Evidence & 0.72 & 2.89 & 46 & [0.22,\ 1.23] & 0.006$^{**}$ \\
wissenschaftlich & Evidence & 2.87 & 12.33 & 46 & [2.40,\ 3.34] & \textless{}.001$^{***}$ \\
zeigen & Evidence & 0.17 & 0.81 & 46 & [-0.26,\ 0.60] & 0.425 \\
\midrule
\multicolumn{7}{l}{\textit{Intuition keywords}} \\
\midrule
Ansicht & Intuition & -1.43 & -5.74 & 46 & [-1.93,\ -0.93] & \textless{}.001$^{***}$ \\
AuthentizitÃ¤t & Intuition & -0.32 & -1.18 & 46 & [-0.86,\ 0.23] & 0.244 \\
BauchgefÃ¼hl & Intuition & -3.43 & -17.64 & 46 & [-3.82,\ -3.03] & \textless{}.001$^{***}$ \\
Behauptung & Intuition & -1.45 & -5.54 & 46 & [-1.97,\ -0.92] & \textless{}.001$^{***}$ \\
Ehrlichkeit & Intuition & -0.74 & -2.78 & 46 & [-1.28,\ -0.21] & 0.008$^{**}$ \\
Fake News & Intuition & -0.89 & -2.86 & 46 & [-1.52,\ -0.27] & 0.006$^{**}$ \\
Gefühl & Intuition & -3.04 & -15.83 & 46 & [-3.43,\ -2.66] & \textless{}.001$^{***}$ \\
Instinkt & Intuition & -2.91 & -14.48 & 46 & [-3.32,\ -2.51] & \textless{}.001$^{***}$ \\
Intuition & Intuition & -2.40 & -7.51 & 46 & [-3.05,\ -1.76] & \textless{}.001$^{***}$ \\
Meinung & Intuition & -2.34 & -10.44 & 46 & [-2.79,\ -1.89] & \textless{}.001$^{***}$ \\
Misstrauen & Intuition & -1.94 & -7.75 & 46 & [-2.44,\ -1.43] & \textless{}.001$^{***}$ \\
Perspektive & Intuition & -1.15 & -4.55 & 46 & [-1.66,\ -0.64] & \textless{}.001$^{***}$ \\
Propaganda & Intuition & -1.55 & -5.68 & 46 & [-2.10,\ -1.00] & \textless{}.001$^{***}$ \\
Rat & Intuition & -0.83 & -3.85 & 46 & [-1.26,\ -0.40] & \textless{}.001$^{***}$ \\
Ratschlag & Intuition & -0.98 & -4.10 & 46 & [-1.46,\ -0.50] & \textless{}.001$^{***}$ \\
Sichtweise & Intuition & -1.94 & -8.06 & 46 & [-2.42,\ -1.45] & \textless{}.001$^{***}$ \\
Sinn & Intuition & -0.21 & -0.70 & 46 & [-0.82,\ 0.40] & 0.488 \\
Standpunkt & Intuition & -0.74 & -2.71 & 46 & [-1.30,\ -0.19] & 0.009$^{**}$ \\
Täuschung & Intuition & -0.66 & -2.61 & 46 & [-1.17,\ -0.15] & 0.012$^{*}$ \\
Unehrlichkeit & Intuition & -0.91 & -3.71 & 46 & [-1.41,\ -0.42] & \textless{}.001$^{***}$ \\
Ungenauigkeit & Intuition & -0.09 & -0.29 & 46 & [-0.67,\ 0.50] & 0.770 \\
Unwahrheit & Intuition & -0.49 & -1.73 & 46 & [-1.06,\ 0.08] & 0.091 \\
Verdacht & Intuition & -0.96 & -3.71 & 46 & [-1.48,\ -0.44] & \textless{}.001$^{***}$ \\
Vorschlag & Intuition & -0.34 & -1.37 & 46 & [-0.84,\ 0.16] & 0.176 \\
Zweifel & Intuition & -1.02 & -3.63 & 46 & [-1.59,\ -0.46] & \textless{}.001$^{***}$ \\
aufrichtig & Intuition & -0.96 & -3.91 & 46 & [-1.45,\ -0.46] & \textless{}.001$^{***}$ \\
authentisch & Intuition & -0.45 & -1.87 & 46 & [-0.93,\ 0.03] & 0.068 \\
behaupten & Intuition & -1.47 & -5.24 & 46 & [-2.03,\ -0.90] & \textless{}.001$^{***}$ \\
beteuern & Intuition & -1.62 & -6.52 & 46 & [-2.12,\ -1.12] & \textless{}.001$^{***}$ \\
ehrlich & Intuition & -0.85 & -3.77 & 46 & [-1.31,\ -0.40] & \textless{}.001$^{***}$ \\
gesunder Menschenverstand & Intuition & -1.40 & -4.58 & 46 & [-2.02,\ -0.79] & \textless{}.001$^{***}$ \\
glauben & Intuition & -2.57 & -10.36 & 46 & [-3.07,\ -2.07] & \textless{}.001$^{***}$ \\
irreführen & Intuition & -0.66 & -2.27 & 46 & [-1.24,\ -0.07] & 0.028$^{*}$ \\
lügen & Intuition & -1.09 & -4.69 & 46 & [-1.55,\ -0.62] & \textless{}.001$^{***}$ \\
offensichtlich & Intuition & -0.62 & -2.06 & 46 & [-1.22,\ -0.02] & 0.045$^{*}$ \\
relativ & Intuition & -0.47 & -2.02 & 46 & [-0.93,\ -0.00] & 0.049$^{*}$ \\
täuschen & Intuition & -0.85 & -3.23 & 46 & [-1.38,\ -0.32] & 0.002$^{**}$ \\
unecht & Intuition & -0.36 & -1.47 & 46 & [-0.86,\ 0.13] & 0.148 \\
unehrlich & Intuition & -0.85 & -3.25 & 46 & [-1.38,\ -0.32] & 0.002$^{**}$ \\
ungenau & Intuition & -0.38 & -1.25 & 46 & [-1.00,\ 0.23] & 0.218 \\
unsauber & Intuition & -0.55 & -1.98 & 46 & [-1.12,\ 0.01] & 0.054 \\
unterstützen & Intuition & -0.79 & -3.19 & 46 & [-1.28,\ -0.29] & 0.003$^{**}$ \\
unwahr & Intuition & -0.28 & -1.03 & 46 & [-0.82,\ 0.27] & 0.311 \\
unwahrhaftig & Intuition & -0.26 & -0.94 & 46 & [-0.80,\ 0.29] & 0.354 \\
vermuten & Intuition & -1.96 & -7.29 & 46 & [-2.50,\ -1.42] & \textless{}.001$^{***}$ \\
vernünftig & Intuition & -0.15 & -0.45 & 46 & [-0.81,\ 0.51] & 0.653 \\
vertrauen & Intuition & -1.57 & -5.26 & 46 & [-2.18,\ -0.97] & \textless{}.001$^{***}$ \\
vertrauenswürdig & Intuition & -0.64 & -2.02 & 46 & [-1.28,\ -0.00] & 0.050$^{*}$ \\
verwechselt & Intuition & -0.66 & -2.33 & 46 & [-1.23,\ -0.09] & 0.024$^{*}$ \\
vorschlagen & Intuition & -0.47 & -2.01 & 46 & [-0.94,\ 0.00] & 0.051 \\
zensieren & Intuition & -0.09 & -0.30 & 46 & [-0.65,\ 0.48] & 0.764 \\
Überzeugung & Intuition & -1.64 & -5.86 & 46 & [-2.20,\ -1.08] & \textless{}.001$^{***}$ \\
\end{longtable}
\newpage

\subsection{Validation Survey Document Level}
The Final validation step includes a annotation survey on prolific for a corpus of 1300 chunks of parliament speeches. These were sampled in five bins for intuition, evidence and emi over decades. The survey was conducted using the Potato Annotation Tool\footnote{\url{https://potato-annotation.readthedocs.io/en/latest/}} integrated with Prolific for participant recruitment. Following prior research \cite{lasser-2023}, participants were asked to evaluate the extent to which a given text is evidence-based or intuition-based using a Likert scale. The survey aimed at German-speaking participants to ensure linguistic and cultural relevance in assessing rhetorical styles. Given the challenges of subjective text interpretation, we applied careful screening measures to recruit german speaking participants. We incorporated two attention checks by asking participants about examples presented in the instructions. In total, we recruited 284 participants, each rating 20 speeches and answering 2 test questions. Of these, 32 participants failed the attention check, and their responses were subsequently excluded from the validation set.

\subsubsection{User Agreement}
\begin{figure}[H]
    \centering
    \includegraphics[width=\textwidth]{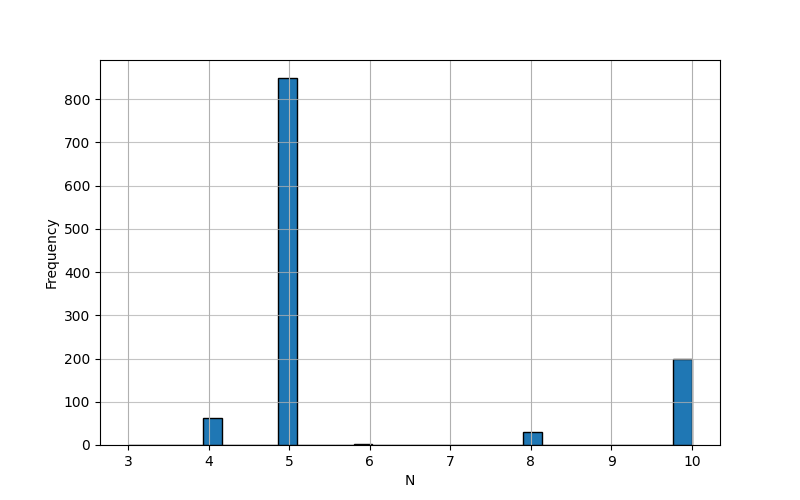}
    \caption{Distribution of Annotations}
    \label{fig:cons}
\end{figure}

We conducted five human annotations for 1000 texts and ten annotations for 300 texts. Figure \ref{fig:cons} shows the distibtuion of ratings after exluding failed participants.  

Participants rated statements on two dimensions: evidence and intuition using a 5-point Likert scale from -2 to 2. Each statement was independently rated in two separate columns: one for evidence and one for intuition. To facilitate classification and comparison with EMI scores, we transformed these ordinal ratings into binary labels. Ratings of 1 (Agree) and 2 (Strongly Agree) were mapped to 1, while ratings of 0 (Neutral), -1 (Disagree), and -2 (Strongly Disagree) were mapped to 0. A majority voting scheme was applied, assigning 1 if at least three out of five annotators rated it as 1; otherwise, the label was 0. This transformation ensured consistency with EMI scores. 
\newpage

To assess annotation reliability, we calculated user agreement and Intraclass Correlation Coefficients (ICCs). Agreement on the original Likert-scale ratings was 48.58\% for evidence and 51.04\% for intuition, representing the proportion of annotators selecting the most common score. After binarization, agreement rose to 71.56\% and 71.51\%, respectively, indicating ~70\% average consensus. We used ICCs to quantify inter-rater reliability, considering three models: \begin{itemize} \item \textbf{ICC(1)} (One-Way Random): assumes different raters per item, suitable when raters are drawn from a larger population. \item \textbf{ICC(2)} (Two-Way Random): assumes raters and items are random samples; reflects consistency across interchangeable raters. \item \textbf{ICC(3)} (Two-Way Mixed): assumes fixed raters across all items; less relevant here due to varying rater assignment. \end{itemize}

Given each rater annotated only 20 out of 1300 texts, \textbf{ICC(1)} is most appropriate, with \textbf{ICC(2)} serving as an additional consistency check. Since ratings were averaged, we also report the corresponding \textbf{k-versions} (ICC1k, ICC2k), which account for increased reliability through aggregation.For evidence, single-rater ICCs were 0.14 (95\% CI: [0.10, 0.18]), increasing to 0.62 (95\% CI: [0.53, 0.69]) when averaged. For intuition, single-rater ICCs were 0.16 (95\% CI: [0.12, 0.21]), rising to 0.65 (95\% CI: [0.57, 0.72]). These results indicate low individual agreement but strong reliability through aggregated ratings.

\begin{table}[h]
    \centering
    \caption{Intraclass Correlation Coefficients for Evidence Annotations}
    \label{tab:icc_evidence}
    \begin{tabular}{lcccccc}
        \toprule
        ICC Type & ICC Value & F-Value & df1 & df2 & p-value & 95\% CI \\
        \midrule
        ICC1  & 0.1388 & 2.6117 & 197  & 1782 & \(4.00 \times 10^{-25}\) & [0.10, 0.18] \\
        ICC2  & 0.1389 & 2.6139 & 197  & 1773 & \(3.79 \times 10^{-25}\) & [0.10, 0.18] \\
        ICC3  & 0.1390 & 2.6139 & 197  & 1773 & \(3.79 \times 10^{-25}\) & [0.10, 0.18] \\
        ICC1k & 0.6171 & 2.6117 & 197  & 1782 & \(4.00 \times 10^{-25}\) & [0.53, 0.69] \\
        ICC2k & 0.6172 & 2.6139 & 197  & 1773 & \(3.79 \times 10^{-25}\) & [0.53, 0.69] \\
        ICC3k & 0.6174 & 2.6139 & 197  & 1773 & \(3.79 \times 10^{-25}\) & [0.53, 0.69] \\
        \bottomrule
    \end{tabular}
\end{table}

\begin{table}[h]
    \centering
    \caption{Intraclass Correlation Coefficients for Intuition Annotations}
    \label{tab:icc_intuition}
    \begin{tabular}{lcccccc}
        \toprule
        ICC Type & ICC Value & F-Value & df1 & df2 & p-value & 95\% CI \\
        \midrule
        ICC1  & 0.1571 & 2.8643 & 197  & 1782 & \(1.93 \times 10^{-30}\) & [0.12, 0.20] \\
        ICC2  & 0.1578 & 2.8923 & 197  & 1773 & \(5.20 \times 10^{-31}\) & [0.12, 0.20] \\
        ICC3  & 0.1591 & 2.8923 & 197  & 1773 & \(5.20 \times 10^{-31}\) & [0.12, 0.21] \\
        ICC1k & 0.6509 & 2.8643 & 197  & 1782 & \(1.93 \times 10^{-30}\) & [0.57, 0.72] \\
        ICC2k & 0.6521 & 2.8923 & 197  & 1773 & \(5.20 \times 10^{-31}\) & [0.58, 0.72] \\
        ICC3k & 0.6543 & 2.8923 & 197  & 1773 & \(5.20 \times 10^{-31}\) & [0.58, 0.72] \\
        \bottomrule
    \end{tabular}
\end{table}
Survey Instructions and main participant demographics are shown in the next pages.

\newpage
\section*{Studie}
Diese Studie zielt darauf ab, Annotationen politischer Diskurse zu sammeln. Der Fokus liegt hierbei auf Evidenz-basiertem und Intuition-basiertem Sprachgebrauch. In dieser Studie werden wir den Teilnehmer*innen politische Texte präsentieren und sie bitten, zu beurteilen, inwieweit diese auf Beweisen beruhen oder nicht.

\section*{Erklärung}
Politischer Diskurs spielt in verschiedenen Aspekten der Gesellschaft eine grundlegende Rolle, dieser kann Regierungsführungen, Politik und die allgemeine Meinungsbildung beeinflussen. Eine Möglichkeit, Diskurs zu charakterisieren, besteht darin, seine Abhängigkeit von Beweisen zu untersuchen. Dabei konzentrieren wir uns auf Intuition-basierte (evidenzfreie) und Evidenz-basierte Diskurse. Diese beiden Konstrukte schließen sich nicht umbedingt gegenseitig aus. Es ist möglich das ein text sowohl evidenz basierte als auch intuitionsbasierte Passagen hat.

\subsection*{Intuitions Basierte Rethorik}
Intuition-basierter Diskurs stützt sich oft auf Intuition, eine Art Bauchgefühl Anekdoten und Meinungen. Er tendiert weniger zur Analyse verfügbarer Informationen und mehr zu persönlichen Überzeugungen oder emotionalen Appellen\\
Beispiel: "Wir müssen uns die Frage stellen, ob wir weiterhin auf kurzfristige Lösungen setzen oder ob wir den Mut haben, einen Weg zu gehen, von dem wir alle tief im Inneren wissen, dass er langfristig das Beste für unsere Gesellschaft ist."

\subsection*{Evidenz Basierte Rhetorik}
Evidenzbasierter Diskurs verwendet überprüfbare Fakten und Analysen. Der Schwerpunkt liegt auf der Anpassung an oder der Suche nach Beweisen, um zu einer fundierten Perspektive oder einem faktenbasiertem Ergebnis zu gelangen.\\
Beispiel: "Die Daten unserer umfangreichen Studien zeigen klar, dass Investitionen in erneuerbare Energien nicht nur die CO$_2$ Emissionen bis 2030 um 40 \% reduzieren können, sondern auch 250.000 neue Arbeitsplätze schaffen werden, wie aus dem Bericht des Bundesumweltamts hervorgeht."

\section*{Anweisung}
Ihre Aufgabe ist es, für jede der unten stehenden Aussagen zu beurteilen, inwieweit es sich um eine Intuition-basierten (evidenzfreie) oder  Evidenz-basierten Diskurs handelt.

\section*{Textbeispiel}
\noindent \textbf{Bewerten Sie diesen Text:}

\vspace{0.5cm}

"Das war der Zweck meiner Bemerkungen, jedoch nicht, um zu behaupten, dass es irgendwie einen Widerspruch zwischen der Geschäftsordnung und der Verfassung gibt. Ich bin der Meinung, dass klar und unmissverständlich gesagt werden muss: Wollen wir noch eine Sitzungsperiode im alten Stil oder nicht?"

\vspace{0.5cm}

\noindent
\begin{adjustbox}{max width=\textwidth}
\begin{tabular}{|l|c|c|c|c|c|}
\hline
\textbf{Dimension} & \textbf{Stimme stark zu} & \textbf{Stimme zu} & \textbf{Neutral} & \textbf{Stimme nicht zu} & \textbf{Stimme überhaupt nicht zu} \\ \hline
Evidenz Basiert & \hspace{1.5cm} & \hspace{1.5cm} & \hspace{1.5cm} & \hspace{1.5cm} & \hspace{1.5cm} \\ \hline
Intuitions Basiert & \hspace{1.5cm} & \hspace{1.5cm} & \hspace{1.5cm} & \hspace{1.5cm} & \hspace{1.5cm} \\ \hline
\end{tabular}
\end{adjustbox}

\newpage
\begin{landscape}  %
\begin{figure}
    \centering
    \includegraphics[width=500pt]{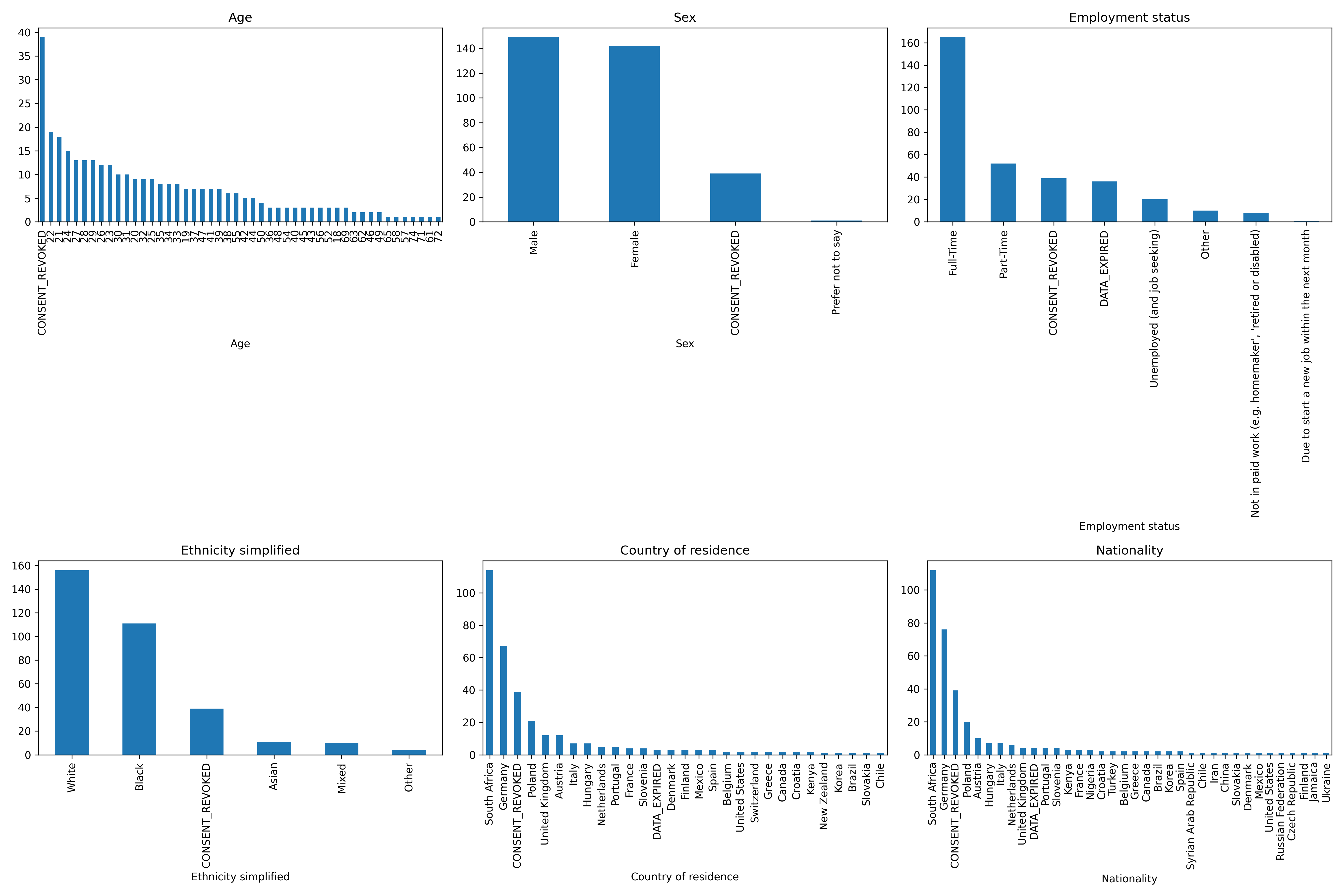}
    \caption{Participant Demographics}
    \label{fig:demo}
\end{figure}
\end{landscape}
\subsubsection{Correlation analysis}

\vspace{0.5cm}
\begin{figure}[h!]
    \centering
    \includegraphics[width=\textwidth]{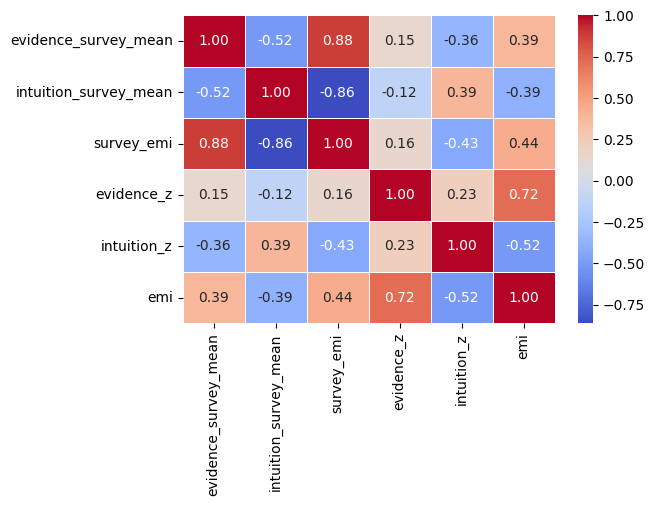}
    \caption{Survey Correlation}
    \label{fig:survey_corr}
\end{figure}

To validate the EMI scores against human annotations, we computed a survey-based equivalent of EMI (\textit{survey\_emi}) using the participants' Likert-scale ratings. The average rating across all annotators was computed for both dimensions as \textit{evidence\_survey\_mean} and \textit{intuition\_survey\_mean}. The survey-based EMI was then derived as the difference between these two mean ratings:

\begin{equation}
\text{survey\_emi} = \text{evidence\_survey\_mean} - \text{intuition\_survey\_mean}
\end{equation}

This metric represents the relative dominance of evidence-based reasoning over intuition-based reasoning in human annotations. The correlation matrix in Figure \ref{fig:survey_corr} provides insights into the relationships between human annotations and computational EMI scores. Moderate positive correlation (\(\rho = 0.15, 95\%CI =  [0.10, 0.20]\)) is observed between \textit{evidence\_survey} and \textit{evidence\_z}, while \textit{intuition\_survey} and \textit{intuition\_z} exhibit a stronger negative correlation (\(\rho = -0.39, 95\%CI = [-0.44, -0.34]\)). This confirms that the computational method effectively captures the contrast between evidence- and intuition-based ratings. A moderate correlation (\(\rho = 0.42, 95\%CI = [0.40, 0.49]\)) exists between \textit{evidence\_survey\_mean} and the computed \textit{emi} score, suggesting a relationship between human annotations and the automated EMI metric.These results suggest that the \textit{survey\_emi} metric effectively captures the balance between evidence-based and intuition-based responses, with a stronger impact from intuition-based reasoning. For the next step we binarized the \textit{survey\_emi} label assigning it 1 for a mean larger than 0.

\newpage
\subsubsection{ROC analysis}
The Receiver Operating Characteristic (ROC) curve is a graphical representation of a classifier’s ability to distinguish between two classes at various threshold levels. It plots the True Positive Rate (TPR) against the False Positive Rate (FPR), where 

\[
\text{TPR} = \frac{\text{True Positives}}{\text{True Positives} + \text{False Negatives}}, \quad
\text{FPR} = \frac{\text{False Positives}}{\text{False Positives} + \text{True Negatives}}
\]

A classifier with strong discriminatory power will have a ROC curve that bends towards the top-left corner. The Area Under the Curve (AUC) quantifies this performance, ranging from 0 to 1. An AUC of 1.0 indicates a perfect classifier, while an AUC of 0.5 corresponds to random guessing. Higher AUC values signify better classification performance.

In the first set of ROC analyses, we assess how well the standardized intuition score (\textit{intuition\_z}) and standardized evidence score (\textit{evidence\_z}) predict human annotations. The ground truth labels for these analyses are the manually assigned binary labels indicating whether a statement was classified as intuition-based (\textit{human\_intuition\_label}) or evidence-based (\textit{human\_evidence\_label}). The ROC curve for \textit{intuition\_z} is computed against \textit{human\_intuition\_label}, while the ROC curve for \textit{evidence\_z} is computed against \textit{human\_evidence\_label}. Additionally, a third ROC curve is computed for the \textit{EMI} score, using the binarized \textit{human\_emi\_label} as the ground truth, providing an overall assessment of how well the EMI score differentiates between evidence- and intuition-based statements. The ROC plot shows three curves: The first corresponds to intuition, yielding an AUC of 0.68, indicating a moderate alignment with human annotations. The second curve represents evidence, with an AUC of 0.59, suggesting weak predictive power which is in line with the correlation analysis before. The third curve evaluates the EMI score against, achieving the highest AUC of 0.72, demonstrating its superior ability to distinguish between evidence- and intuition-based statements. The diagonal dashed gray line represents the random classification baseline.

\begin{figure}[H]
    \centering
    \includegraphics[width=\textwidth]{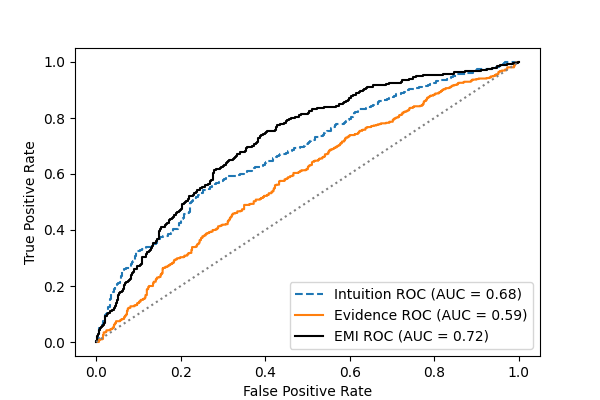}
    \caption{Evidence, Intuition ROC}
    \label{fig:roc1}
\end{figure}

\newpage
In the second set of ROC analyses, we evaluate the performance of the computational EMI score as a predictor of human-assigned evidence and intuition labels. To ensure consistency with the EMI definition, where high values correspond to evidence-based statements and low values to intuition-based statements, we invert the human intuition label by transforming it where 1 indicates a non-intuition-based statement. Two ROC curves are then computed: one for the EMI score predicting \textit{human evidence label} and another for the EMI score predicting \textit{human intuition label inverted}. The resulting ROC plot shows that the EMI score achieves an AUC of 0.69 when predicting the \textit{human evidence label}, as represented by the blue dashed line. Similarly, the EMI score also achieves an AUC of 0.69 when predicting the \textit{human intuition label inverted}, shown as the orange solid line. These identical AUC values indicate that the EMI score is equally effective at distinguishing between both evidence- and intuition-based statements when compared to human annotations. The diagonal dashed gray line represents the baseline of random classification.
\begin{figure}[H]
    \centering
    \includegraphics[width=\textwidth]{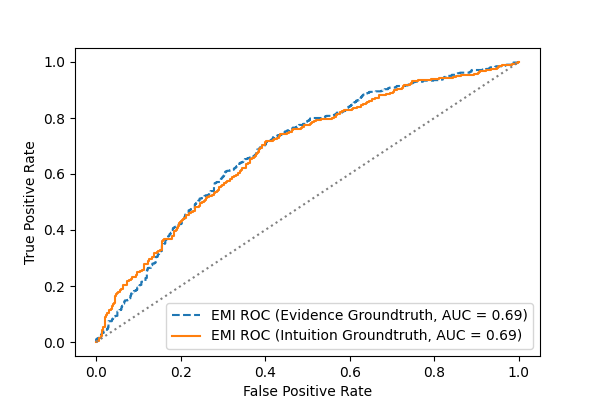}
    \caption{EMI ROC}
    \label{fig:roc2}
\end{figure}

\newpage
\clearpage
\section{Regressions}

\begin{figure}[h!]
\centering\includegraphics[width=\textwidth]{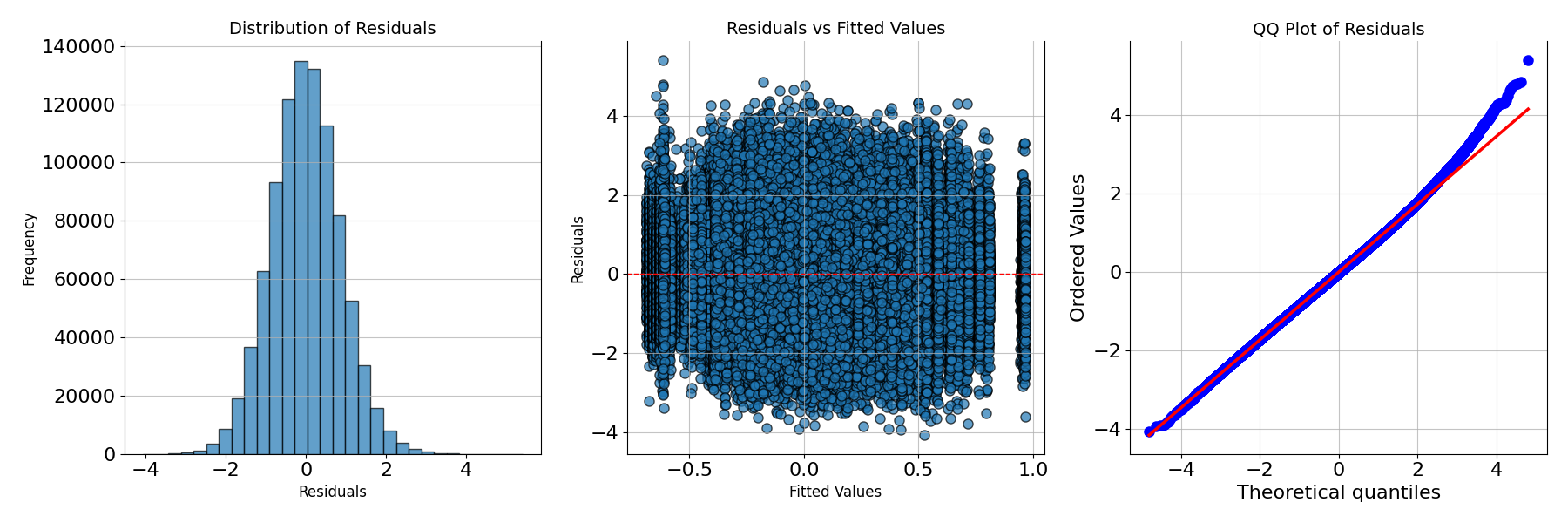}
   \caption{\textbf{Diagnostic plots for the Twitter linear mixed-effects model.} (\textbf{a}) Distribution of standardised residuals, approximately symmetric and centred around zero, indicating no systematic prediction bias. (\textbf{b}) Residuals versus fitted values; the absence of a pronounced non-linear pattern is consistent with the linear specification, while visible clustering reflects the grouped structure induced by actor-level random effects. (\textbf{c}) Quantile--quantile plot of residuals against theoretical normal quantiles, showing close alignment across most of the distribution with moderate deviations in the tails. Taken together, the diagnostics indicate that key model assumptions are reasonably satisfied.}
   \label{fig:diag1}
\end{figure}

\begin{figure}[h!]
\centering\includegraphics[width=\textwidth]{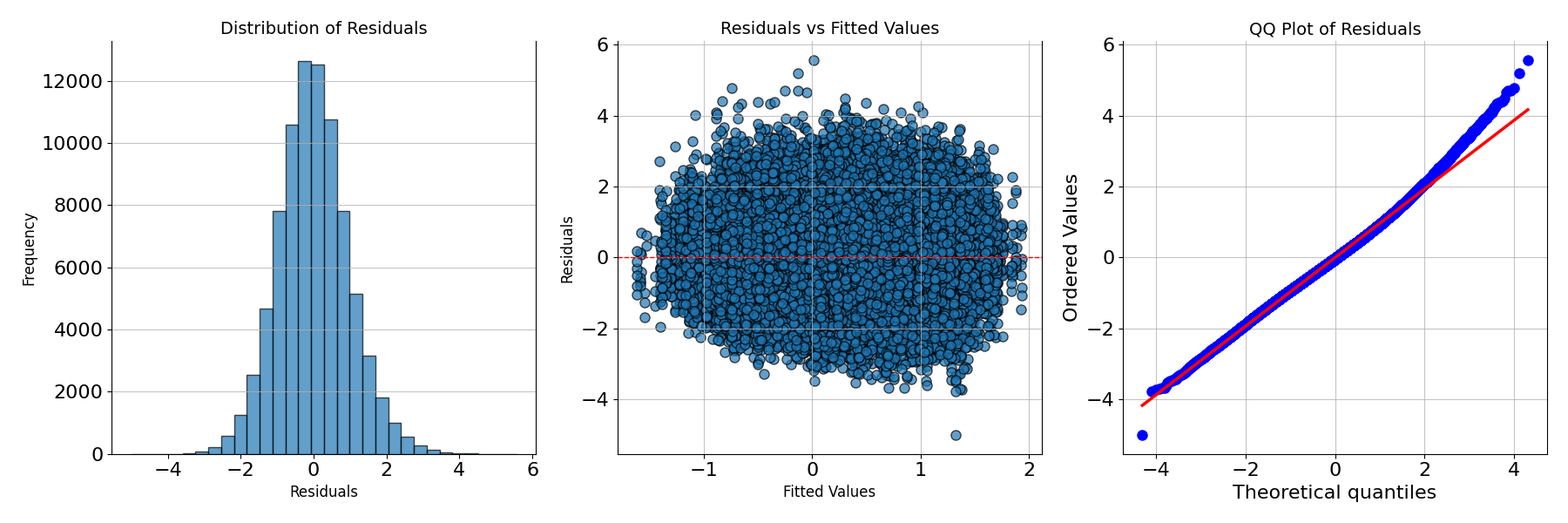}
    \caption{\textbf{Diagnostic plots for the Bundestag breakpoint mixed-effects model.} (\textbf{a}) Distribution of standardised residuals, approximately symmetric and centred around zero. (\textbf{b}) Residuals versus fitted values; no pronounced non-linear pattern, with clustering reflecting the actor-level random-effect structure. (\textbf{c}) Quantile--quantile plot of residuals against theoretical normal quantiles, showing close alignment with moderate tail deviations. Taken together, the diagnostics indicate that key model assumptions are reasonably satisfied.}
     \label{fig:diag2}
\end{figure}

This note reports diagnostic checks for the linear mixed-effects models presented 
in the main text (Methods, Equations~1--3). For each fitted model we inspected 
three standard diagnostics: the distribution of standardised residuals, residuals 
plotted against fitted values, and a quantile--quantile plot of residuals against 
theoretical normal quantiles. Figure~\ref{fig:diag1} shows the diagnostics for 
the Twitter model (Equation~1) and Figure~\ref{fig:diag2} for the Bundestag 
breakpoint model (Equations~2--3). In both cases the residual distributions are 
approximately symmetric and centred around zero, the residual-versus-fitted plots 
show no pronounced non-linear pattern (the visible banding reflects the grouped 
structure induced by the actor-level random effects rather than misspecification), 
and the quantile--quantile plots align closely with the theoretical quantiles 
across most of the distribution, with moderate deviations confined to the tails. 
Given the large sample sizes, such tail deviations have little bearing on the 
fixed-effects estimates. In addition, the lagged EMI term included in all 
specifications adequately addressed serial correlation, as indicated by 
Durbin--Watson statistics on the model residuals close to 2, and model fit was 
supported by likelihood-ratio tests against intercept-only null models 
(see Methods). Taken together, these checks indicate that the key assumptions 
underlying the mixed-effects specifications are reasonably satisfied and that 
the coefficient estimates reported in the main text are not artefacts of model 
misspecification.

\clearpage
\section{Topic Modelling}
To address the agenda-setting confound described in the Discussion (i.e.\ the possibility that observed temporal shifts in EMI reflect changes in \emph{what} is being discussed rather than \emph{how} it is discussed), we estimated a separate BERTopic \cite{grootendorst-2022} model for each corpus and entered the resulting document-level topic assignments as fixed effects in the augmented mixed-effects regressions reported in Methods (Equation~4) and in Extended Data Tables~B8 and~B10.

For each corpus, documents were embedded with the multilingual sentence-transformer paraphrase-multilingual-mpnet-base-v2, which produces 768-dimensional representations and offers strong German-language coverage. Embeddings were then dimensionally reduced with UMAP and clustered with HDBSCAN under the BERTopic defaults; the resulting clusters were reduced to a target of $K = 30$ topics with a minimum cluster size of 50 documents, a setting chosen to balance topical resolution against the requirement that each retained topic carry enough mass to be estimable as a fixed effect in the downstream regression. Topics were labelled by their top class-based TF-IDF (c-TF-IDF) terms; the full term lists per topic are given in Tables~\ref{tab:topics-parliament} and~\ref{tab:topics-twitter} below.

The two corpora differ in size by roughly two orders of magnitude and were fit accordingly. For the contemporary Bundestag corpus the BERTopic model was fit on the full set of 56{,}213 speeches. For the Twitter corpus the model was fit on a stratified subsample of tweets that preserves the joint distribution of ideological leaning and party --- the standard pattern for fitting BERTopic to very large corpora that already span the topic space --- and the fitted model was then used to transform the full corpus to obtain a topic assignment for every tweet. In both corpora, each document received a hard topic assignment $T_d \in \{-1, 0, 1, \ldots, 28\}$, with $T_d = -1$ denoting the BERTopic noise cluster of documents that the model could not confidently assign to any coherent topic. Soft topic probabilities were not retained because the downstream fixed-effects design requires only the identity of each document's assigned topic, not its degree of membership.

In both samples, BERTopic returned 29 named topics plus a noise cluster (rather than the targeted 30), reflecting HDBSCAN merging of clusters that fell below the stable-cluster threshold. The noise cluster accounted for 24{,}130 of 56{,}213 parliamentary speeches (42.9\%) in the topic-modelling sample and 254{,}125 of 499{,}999 tweets (50.8\%) in the Twitter topic-modelling sample. Documents in the noise cluster were excluded from the topic-controlled regressions; the largest remaining topic was selected as the reference category, and the other 28 topics entered the regression as 0/1 dummies (\texttt{topic\_1}--\texttt{topic\_28}). The breakpoint and time coefficients reported in Extended Data Tables~B8 and~B10 are therefore estimated from \emph{within-topic} variation only.

\section*{Face-validity inspection}

The c-TF-IDF term lists for both corpora were inspected for face validity prior to use. In the parliamentary corpus, the recovered topics map cleanly onto well-known German policy domains: the budget and education funding (Topic~1), the COVID-19 pandemic and care system (Topic~2), housing and climate policy (Topic~3), the war in Ukraine (Topic~4), antisemitism and Israel--Iran (Topic~9), Bundeswehr foreign deployments (Topics~11, 12), and several smaller domains spanning agriculture, animal welfare, cannabis policy, China relations, drinking-water protection, religious freedom, AI policy and hydrogen strategy. A small number of topics (e.g.\ Topic~5, dominated by ``sport'' alongside speaker-name terms; Topic~23, dominated by a single MP's name) reflect speaker- or actor-specific clusters rather than policy themes; these were retained because the topic fixed-effects design treats them as additional controls regardless of substantive interpretation.

In the Twitter corpus, the largest topics again map onto core policy domains (climate, Ukraine, COVID-19, Israel and antisemitism, the debt brake and trade, Afghanistan and Iran), but several smaller clusters reflect platform-specific or off-policy content typical of Twitter rather than parliamentary speech: daily-life and casual conversation (Topic~8: \emph{schlafen}, \emph{besserung}, \emph{bett}; Topic~12: food and beverages; Topic~21: weather; Topic~26: running and stress), platform meta-discussion (Topic~17: Elon Musk and the Twitter--X transition; Topic~23: fake accounts and follower listing; Topic~28: Telegram and encrypted-messenger discussion), and short English-language fragments (Topic~15: biographical snippets; Topic~27: short sport/casual phrases). For the agenda-setting confound the relevant property of these clusters is not their thematic coherence but the consistency with which BERTopic assigns related documents to them: any time-varying mix of off-policy or platform-meta content is absorbed by the corresponding dummy and cannot drive the time-trend or breakpoint coefficients in the augmented regression.

\begin{table}[htbp]
\centering
\caption{All 29 named topics recovered by BERTopic in the contemporary Bundestag corpus (2015--2025), in decreasing order of topic size. Topic IDs match the \texttt{topic\_<id>} dummy variables in Extended Data Table~B10. The largest topic (ID~0) is used as the reference category and therefore does not appear as a dummy. The BERTopic noise cluster ($T_d = -1$, $n = 24{,}130$ speeches) is omitted from the table and excluded from the topic-controlled regression.}
\label{tab:topics-parliament}
\small
\begin{tabular}{r r p{10cm}}
\toprule
Topic & $n$ & Top c-TF-IDF terms \\
\midrule
0 & 5{,}688 & gesetzentwurf, cducsu, afd, geht, gibt, kollegen, damen, mehr \\
1 & 4{,}091 & euro, bildung, milliarden, milliarden euro, haushalt, prozent, unternehmen, forschung \\
2 & 3{,}751 & pflege, patienten, pandemie, versorgung, impfpflicht, krankenhuser, menschen, patientinnen \\
3 & 3{,}252 & klimaschutz, wohnungen, mieter, wohnraum, bauen, mietpreisbremse, energien, erneuerbaren \\
4 & 3{,}113 & ukraine, russland, bundeswehr, nato, putin, soldaten, krieg, russischen \\
5 & 2{,}262 & sport, spd, norbert, liebe, kollegen, menschen, kolleginnen, kultur \\
6 & 2{,}074 & kinder, familien, eltern, kindern, kind, ehe, gewalt, menschenrechte \\
7 & 1{,}153 & daten, digitalisierung, digitale, digitalen, datenschutz, meinungsfreiheit, internet, verwaltung \\
8 & 1{,}090 & griechenland, europa, europischen, ceta, europische, griechischen, abkommen, europischen union \\
9 & 1{,}068 & israel, antisemitismus, juden, iran, israels, hamas, jdinnen, jdinnen juden \\
10 & 774 & landwirtschaft, landwirte, ernhrung, bauern, lebensmittel, landwirten, betriebe, agrarpolitik \\
11 & 707 & mali, mission, soldaten, sdsudan, mandat, einsatz, libyen, operation \\
12 & 642 & afghanistan, libanon, taliban, syrien, irak, einsatz, soldaten, mandat \\
13 & 488 & verbraucher, verbraucherinnen, banken, verbraucherschutz, verbraucherinnen verbraucher, bafin, glubiger, richtlinie \\
14 & 393 & kosovo, trkei, erdogan, serbien, westbalkan, trkischen, region, trkische \\
15 & 303 & tierschutz, tiere, wolf, tieren, wlfe, tierhaltung, tierwohl, tier \\
16 & 200 & cannabis, rauchen, legalisierung, substanzen, schwarzmarkt, konsum, drogen, konsumenten \\
17 & 127 & china, hongkong, chinastrategie, chinesischen, chinesische, volksrepublik, chinas, peking \\
18 & 127 & afrika, entwicklungszusammenarbeit, afrikanischen, friedhoff, entwicklungshilfe, kontinent, friedhoff afd, dietmar friedhoff \\
19 & 124 & inseln, griechenland, griechischen inseln, griechischen, europischen mitgliedstaaten, aufnahme, moria, geflchteten \\
20 & 99 & recycling, verpackungen, plastik, abfall, umwelt, kreislaufwirtschaft, mll, thews \\
21 & 95 & fracking, trinkwasser, bereits vorhandene, expertenkommission, verantwortbar, umwelt, serises, mensch umwelt \\
22 & 89 & religionsfreiheit, christen, religion, religionen, bericht, muslime, religise, menschenrecht \\
23 & 75 & hacker, thomas hacker, thomas, welle, deutsche welle, freiheit, rundfunk, kultur \\
24 & 67 & wasser, trinkwasser, gewsser, sauberes, damerow, wasserversorgung, sauberem, astrid damerow \\
25 & 60 & intelligenz, knstliche intelligenz, knstliche, knstlicher intelligenz, knstlicher, enquetekommission, knstlichen intelligenz, technologie \\
26 & 59 & drohnen, drohne, bewaffnete, bewaffneten, bewaffnung, faber, soldaten, beschaffung \\
27 & 56 & grenzregionen, benutzung, parlamentarischen beratungen, spd deutsche, ursprnglichen gesetzentwurf, einfhrung, spdbundestagsfraktion, bedenken \\
28 & 56 & wasserstoff, wasserstoffstrategie, nestle, rimkus, grnen wasserstoff, andreas rimkus, ingrid nestle, energiewende \\
\bottomrule
\end{tabular}
\end{table}

\begin{table}[htbp]
\centering
\caption{All 29 named topics recovered by BERTopic in the German Twitter corpus (2015--2025), in decreasing order of topic size. Topic IDs match the \texttt{topic\_<id>} dummy variables in Extended Data Table~B8. The largest topic (ID~0) is used as the reference category and therefore does not appear as a dummy. The BERTopic noise cluster ($T_d = -1$, $n = 254{,}125$ tweets in the topic-modelling sample) is omitted from the table and excluded from the topic-controlled regression.}
\label{tab:topics-twitter}
\small
\begin{tabular}{r r p{10cm}}
\toprule
Topic & $n$ & Top c-TF-IDF terms \\
\midrule
0 & 104{,}902 & merkel, twitter, tweet, spd, glckwunsch, digitalen, interview, cdu \\
1 & 54{,}017 & polizei, migration, kinder, brexit, frauen, flchtlinge, schulen, eltern \\
2 & 15{,}400 & klimaschutz, klimakrise, klimawandel, energiewende, klima, wind, energie, strom \\
3 & 12{,}810 & ukraine, putin, russland, erdogan, russische, krieg, nato, polen \\
4 & 11{,}723 & corona, impfung, coronakrise, pandemie, covid, virus, pflege, gesundheit \\
5 & 10{,}461 & israel, juden, hanau, gewalt, juedischeonline, opfer, frauen, terror \\
6 & 8{,}194 & afghanistan, iran, syrien, islam, regime, terror, trkei, abschiebungen \\
7 & 7{,}202 & schuldenbremse, ttip, inflation, rente, schulden, steuern, ceta, trade \\
8 & 5{,}397 & elonmusk, schlafen, besserung, bett, niemamovassat, tatort, eher, lesen \\
9 & 4{,}239 & bauern, wissenschaft, bmel, lebensmittel, fleisch, forschung, wissenschaftler, juliakloeckner \\
10 & 2{,}360 & macron, carlomasala, frankreich, paris, france, italien, fabiodemasi, esc \\
11 & 1{,}864 & china, taiwan, sdkorea, menschenrechte, usa, the, with, abhngigkeit \\
12 & 1{,}516 & trinken, wein, essen, tee, bier, frhstck, kse, wasser \\
13 & 1{,}163 & trump, biden, donald, donald trump, joe biden, joe, realdonaldtrump, trumps \\
14 & 831 & lillyblaudszun, bahajam, said, wnsche, feiern, ali, gedenken, olsun \\
15 & 628 & born, she, woman, from, june, french, february, july \\
16 & 613 & homeoffice, office, handy, home, app, bro, apple, mobile \\
17 & 445 & musk, elon, elon musk, elonmusk, twitter, weidel, brandenburg, space \\
18 & 388 & london, court, freilassung, usa, sevimdagdelen, freiheit, freedom, free \\
19 & 379 & spenden, blut, retten, rettet, selbstbestimmung, forschung, warten, gespendet \\
20 & 327 & katastrophe, gedanken, betroffenen, severe, opfer, opfern, betroffene, helfen \\
21 & 231 & regen, wetter, sturm, schirm, maybe, spiel, mus, gelaufen \\
22 & 193 & supermarkt, google, beschftigten, gewinn, verdi, bestellt, ausbeutung, arbeitsbedingungen \\
23 & 144 & people, account, person, one, fake, follower, dozens, arnddiringer \\
24 & 123 & frauen, pay, mnnern, verdienen, gleiche, bezahlung, mnner, gender \\
25 & 116 & haare, maske, masken, tragen, schneiden, gesicht, trage, supermarkt \\
26 & 90 & laufen, stress, homeoffice, check, gelaufen, sport, katjakeul, entspannung \\
27 & 63 & win, what the, idea, his, best, doing, team, game \\
28 & 55 & telegram, per, geteilt, group, signal, jagen, helft, erreichbar \\
\bottomrule
\end{tabular}
\end{table}
\clearpage
To rule out that the temporal and breakpoint effects reported in the main text 
reflect a changing topic mix rather than a shift in epistemic style, we re-estimated 
the baseline regression specifications with document-level topic fixed effects, 
using the BERTopic assignments described in Supplementary Note~3.1 (the largest 
topic serving as the reference category). Because the fixed effects absorb all 
between-topic variation, the time, leaning and breakpoint coefficients in the 
augmented models are identified from within-topic variation only. 
Tables~\ref{tab:topic_comparison_parliament} and~\ref{tab:topic_comparison_twitter} 
compare the baseline and topic-controlled estimates for the Bundestag and Twitter 
corpora, respectively. Across both corpora the substantive coefficients are 
essentially unchanged: the negative baseline EMI of right-leaning actors, the 
overall time trends and all election-year level shifts retain their sign, 
magnitude and significance, with the largest attenuation observed for the 2021 
level shift in Parliament ($-0.246$ to $-0.194$). We therefore conclude that the 
rhetorical shifts documented in the main text are not driven by changes in the 
topical composition of political debate.

\begin{table}
\caption{Parliament: baseline vs.\ topic-controlled model}
\label{tab:topic_comparison_parliament}%
\begin{center}
\begin{tabular}{@{}lrrrr@{}}
\toprule
 & \multicolumn{2}{c}{Baseline} & \multicolumn{2}{c}{Topic-controlled} \\
 & Coef. & $p$ & Coef. & $p$ \\
\midrule
Intercept                       &   0.603 &  $<$0.001 &   0.560 &  $<$0.001  \\
leaning[T.left]                 &  -0.139 &     0.045 &  -0.148 &     0.013  \\
leaning[T.right]                &  -0.404 &  $<$0.001 &  -0.405 &  $<$0.001  \\
time\_centered                  &   0.003 &     0.076 &   0.003 &     0.066  \\
time\_centered:leaning[T.left]  &   0.002 &     0.005 &   0.002 &     0.008  \\
time\_centered:leaning[T.right] &   0.001 &     0.225 &   0.001 &     0.375  \\
slope\_elec2017                 &   0.000 &     0.999 &  -0.002 &     0.303  \\
level\_elec2017                 &  -0.419 &  $<$0.001 &  -0.392 &  $<$0.001  \\
slope\_elec2021                 &  -0.009 &  $<$0.001 &  -0.006 &  $<$0.001  \\
level\_elec2021                 &  -0.246 &  $<$0.001 &  -0.194 &  $<$0.001  \\
slope\_elec2025                 &   0.340 &  $<$0.001 &   0.345 &  $<$0.001  \\
level\_elec2025                 &  -0.528 &  $<$0.001 &  -0.525 &  $<$0.001  \\
emi\_lag1                       &   0.130 &  $<$0.001 &   0.122 &  $<$0.001  \\
Group Var                       &   0.392 &  $<$0.001 &   0.302 &  $<$0.001  \\
\botrule
\end{tabular}
\end{center}
\end{table}

\begin{table}
\caption{Twitter: baseline vs.\ topic-controlled model}
\label{tab:topic_comparison_twitter}%
\begin{center}
\begin{tabular}{@{}lrrrr@{}}
\toprule
 & \multicolumn{2}{c}{Baseline} & \multicolumn{2}{c}{Topic-controlled} \\
 & Coef. & $p$ & Coef. & $p$ \\
\midrule
Intercept                       &   0.050 &  $<$0.001 &   0.021 &     0.003  \\
leaning[T.left]                 &   0.001 &     0.945 &   0.005 &     0.813  \\
leaning[T.right]                &  -0.111 &  $<$0.001 &  -0.110 &  $<$0.001  \\
time\_centered                  &  -0.001 &  $<$0.001 &  -0.001 &  $<$0.001  \\
time\_centered:leaning[T.left]  &  -0.000 &     0.494 &  -0.000 &     0.460  \\
time\_centered:leaning[T.right] &   0.001 &  $<$0.001 &   0.001 &  $<$0.001  \\
log\_like\_count                &  -0.040 &  $<$0.001 &  -0.043 &  $<$0.001  \\
emi\_lag1                       &   0.495 &  $<$0.001 &   0.463 &  $<$0.001  \\
Group Var                       &   0.075 &           &   0.075 &            \\
\botrule
\end{tabular}
\end{center}
\end{table}

\newpage
\bibliographystyle{IEEEtran}
\bibliography{supplementary.bib}